%% file: example_paper.tex
\documentclass{article}

\usepackage{microtype}
\usepackage{graphicx}
\usepackage{subcaption}
\usepackage{booktabs} 

\usepackage{hyperref}

\usepackage[accepted]{icml2026}

\usepackage{amsmath}
\usepackage{amssymb}
\usepackage{mathtools}
\usepackage{amsthm}
\usepackage{tcolorbox}
\usepackage{url}

\usepackage{booktabs}
\usepackage{multirow}
\usepackage{adjustbox}
\usepackage[table]{xcolor}
\definecolor{lightblue}{RGB}{230,240,255}
\usepackage{fontawesome}
\usepackage{graphicx}
\usepackage{subcaption}
\usepackage[justification=centering]{subcaption}

\usepackage[capitalize,noabbrev]{cleveref}

\theoremstyle{plain}
\newtheorem{theorem}{Theorem}[section]

\theoremstyle{definition}

\theoremstyle{remark}

\usepackage[textsize=tiny]{todonotes}

\icmltitlerunning{LEC: Linear Expectation Constraints for Selection-Conditioned Risk Control in Selective Prediction and Routing Systems}

\begin{document}

\twocolumn[
  \icmltitle{LEC: Linear Expectation Constraints for Selection-Conditioned Risk Control in Selective Prediction and Routing Systems}



  \icmlsetsymbol{equal}{*}

  \begin{icmlauthorlist}
    \icmlauthor{Zhiyuan Wang}{1}
    \icmlauthor{Aniri}{2,7}
    \icmlauthor{Tianlong Chen}{3}
    \icmlauthor{Yue Zhang}{4}
    \icmlauthor{Heng Tao Shen}{5}
    \icmlauthor{Xiaoshuang Shi}{1}
    \icmlauthor{Kaidi Xu}{6}
  \end{icmlauthorlist}

  \icmlaffiliation{1}{University of Electronic Science and Technology of China}
  \icmlaffiliation{2}{LMU Munich}
  \icmlaffiliation{7}{Munich Center for Machine Learning}
  \icmlaffiliation{3}{University of North Carolina at Chapel Hill}
  \icmlaffiliation{4}{Shandong University}
  \icmlaffiliation{5}{Tongji University}
  \icmlaffiliation{6}{City University of Hong Kong}

  \icmlcorrespondingauthor{Xiaoshuang Shi}{xsshi2013@gmail.com}
  \icmlcorrespondingauthor{Kaidi Xu}{kaidixu@cityu.edu.hk}

  \icmlkeywords{Machine Learning, ICML}

  \vskip 0.3in
]



\printAffiliationsAndNotice{}  

\input{sections/abstract}
\input{sections/introduction}
\input{sections/relatedWork}
\input{sections/method}

\input{sections/experiments}
\input{sections/conclusion}

\section*{Acknowledgements}
The paper was supported by Noncommunicable Chronic Diseases-National
Science and Technology Major Project (2025ZD0551300, 2025ZD0551302).

\input{sections/impactStatement}


\bibliography{ref}
\bibliographystyle{icml2026}

\newpage
\appendix
\onecolumn
\input{sections/proofs}
\input{sections/additionalExperimentalSettings}
\input{sections/additionResults}

\end{document}

%% file: sections/abstract.tex
\begin{abstract}
Foundation models often generate unreliable answers, while heuristic uncertainty estimators fail to fully distinguish correct from incorrect outputs, causing users to accept erroneous answers without any statistical guarantee. 
We address this problem through selection-conditioned risk control, aiming to ensure that an accepted prediction has an error probability no larger than a user-specified risk level. 
To this end, we propose \texttt{LEC}, a principled framework that reframes selective prediction as a decision problem governed by a \emph{linear expectation constraint} over selection and error indicators. 
This formulation directly controls the ratio between the expected number of accepted errors and the expected number of accepted predictions, which corresponds to the marginal error probability conditioned on selection. 
Under exchangeability, we derive a \emph{finite-sample sufficient condition} that relies only on a held-out calibration set, enabling the computation of a risk-constrained, retention-maximizing threshold. 
Furthermore, we extend \texttt{LEC} to two-model routing systems: if the primary model's uncertainty exceeds its calibrated threshold, the input is delegated to a subsequent model, while maintaining system-level selection-conditioned error control. 
Experiments on both closed-ended and open-ended question answering (QA) and vision question answering (VQA) demonstrate that \texttt{LEC} maintains the prescribed risk level in accepted predictions and substantially improves sample retention compared to baselines. 
\end{abstract}

%% file: sections/introduction.tex
\section{Introduction}
\label{sec: Introduction}
Foundation models, like large language models (LLMs) and large vision-language models (LVLMs), are increasingly being integrated into real-world decision-making pipelines~\citep{xiaolan2025evaluating,brady2025dual,singhal2025toward}, where it is crucial to evaluate the reliability of their outputs and determine whether to trust them. 
Uncertainty quantification (UQ) is a promising approach to estimate the uncertainty of model predictions, with the uncertainty score serving as an indicator of whether the model's output is likely to be incorrect~\citep{zhang2024vl,wang2025word,duan2024shifting,duan2025uprop}. 
In practice, when the model shows high uncertainty, its predictions should be clarified or abstained from to prevent the propagation of incorrect information. 

\begin{figure}[!t]
    \centering
    \includegraphics[width=\linewidth]{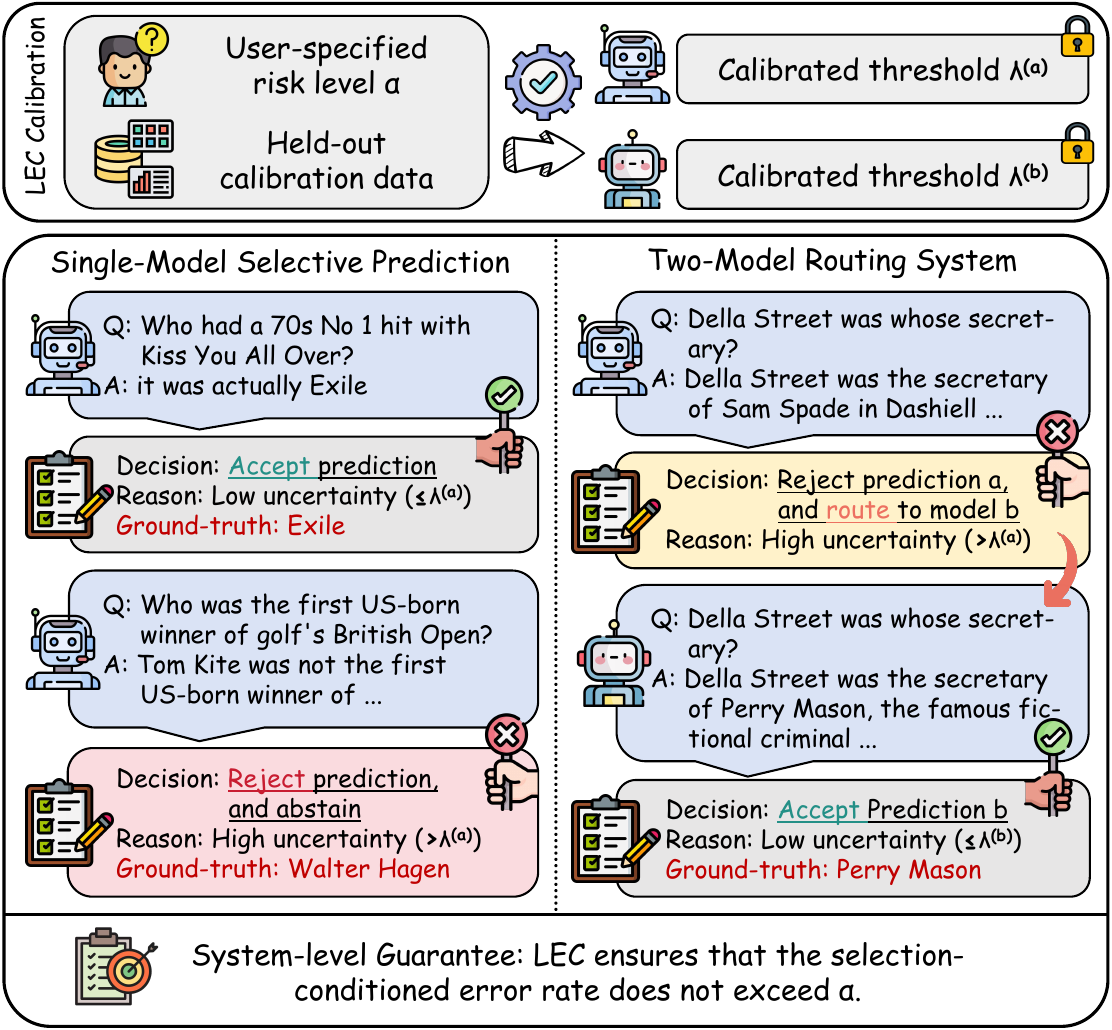}
    \caption{Illustration of selective prediction in single-model and two-model routing systems. By calibrating when to accept, escalate, or abstain, \texttt{LEC} provides system-level selection-conditioned error control. Code is available \href{https://github.com/Zhiyuan-GG/LinearExpectationConstraint}{here}.}
    \label{fig: overview}
\end{figure}

However, when the model generates hallucinations or exhibits overconfidence in its erroneous predictions~\citep{shorinwa2025survey,atf2025challenge}, uncertainty scores derived from model logits or self-consistency measures may remain low, leading users to accept incorrect answers without task-specific risk guarantees~\citep{angelopoulos2024conformal}. 
Split conformal prediction (SCP) can transform heuristic uncertainty notions to statistically calibrated decision rules~\citep{angelopoulos2021gentle,campos2024conformal,tan2025conformal}. 
Assuming data exchangeability, SCP produces prediction sets that include ground-truth answers with at least a user-defined probability~\citep{li2026set}. 
Nonetheless, set-valued predictions often contain unreliable candidates, leading to biased decision-making in downstream tasks~\citep{wang2025safer,cresswell2025conformal}. 
In this paper, we investigate point prediction with provable finite-sample guarantees on the error rate among accepted predictions.

Although uncertainty scores cannot perfectly separate correct from incorrect predictions, selective prediction allows us to enforce a prespecified risk level (e.g., $\alpha$): a prediction is accepted if and only if its associated uncertainty score falls below a calibrated threshold, ensuring that the selection-conditioned error rate does not exceed $\alpha$. 
To achieve this in a principled way, we introduce \texttt{LEC}, which reframes selective prediction not as an uncertainty-ranking problem, but as a decision problem governed by a statistical constraint. 
The central idea is to express selection-conditioned error control as a constraint on the expectation of a linear functional involving two binary indicators: one capturing whether a prediction is selected and the other indicating whether it is incorrect. 
This formulation enables us to establish a finite-sample sufficient condition utilizing calibration uncertainty scores and error labels that, if satisfied, guarantees selection-conditioned error control for unseen test samples. 
Since this condition depends only on the empirical quantities from the calibration set, it yields a calibrated threshold that maximizes retention subject to the prescribed risk constraint. 

We further extend \texttt{LEC} to a two-model routing framework. 
For each input, the system accepts the current model’s prediction if its uncertainty falls below a calibrated threshold; otherwise, the input is routed to the subsequent model. 
If neither model satisfies its acceptance criterion, the system abstains. 
To preserve the statistical guarantee, we impose a linear expectation constraint on the system-level selection and error indicators, which enables joint calibration of model-specific thresholds with unified system-level selection-conditioned error control. 
Figure~\ref{fig: overview} illustrates examples of selective prediction in single-model prediction and two-model routing systems, where uncertainty serves as the decision signal for \emph{accepting}, \emph{routing}, or \emph{abstaining}. 

We evaluate \texttt{LEC} on four benchmarks across closed-ended and open-ended generation scenarios. 
In selective prediction of both single-model and two-model routing systems, \texttt{LEC} keeps the empirical accepted error rate below the prescribed risk level across feasible risk levels, consistent with its finite-sample selection-conditioned error guarantee. 
Compared to confidence interval-based methods~\citep{wang2025coin,jung2025trust}, \texttt{LEC} establishes tighter risk control while accepting more admissible samples (e.g., $+9\%$ on TriviaQA). 
Furthermore, across different UQ methods, admission functions, calibration-test split ratios, and sampling sizes under black-box scenarios, \texttt{LEC} maintains statistical rigor while consistently achieving higher power than the best baseline. 
These results highlight the practical effectiveness and generality of \texttt{LEC}, motivating its potential integration into real-world uncertainty-aware agentic systems.

\noindent \textbf{Conflict of Interest Disclosure.} 
The authors declare that they have no financial conflicts of interest related to this paper. No author is employed by, consults for, or holds equity in any company that could benefit from the results
presented in this study. 
All funding sources are acknowledged in the Acknowledgments section.

%% file: sections/relatedWork.tex
\section{Related Work}
\label{sec: Related Work}
\noindent \textbf{SCP in LLMs.} 
SCP provides statistical guarantees of coverage for correct answers~\citep{campos-etal-2024-conformal}.
It evaluates the nonconformity (or residual) between model prediction and ground-truth on a calibration set, and then computes a rigorously calibrated threshold, which is applied to construct prediction/conformal sets at test time. 
Under exchangeability~\citep{angelopoulos2023conformal}, these sets contain admissible answers with at least a user-specified probability.
However, previous research predominantly focuses on \emph{set-valued predictions}~\citep{quach2024conformal,kaur2024addressing,wang2024conu,wang2025sample,wang2025safer,li2026set}, which are not inherently actionable due to unreliable candidates, and can cause disparate impact~\citep{cresswell2024conformal,cresswell2025conformal}. 
Our work targets selection-conditioned risk control over accepted \emph{point predictions}, rather than conformal coverage. 

\noindent \textbf{Risk Control in Selective Prediction.}
Several frameworks grounded in significance testing~\citep{jin2023selection,jin2025model} and confidence intervals~\citep{bates2021distribution} have been introduced to provide statistical error control for selective prediction~\citep{jia2026balancerag}. 
For instance, conformal alignment~\citep{gui2024conformal} and labeling~\citep{huang2025selective} calculate conformal p-values and control false discoveries under multiple-testing formulations. 
To retain more admissible answers and accelerate test-time inference, COIN~\citep{wang2025coin} constructs an upper confidence bound (UCB) for the system risk on calibration examples and computes a rigorous threshold for test-time selection, achieving PAC-style risk control~\citep{park2020pac}. 
Furthermore, Trust of Escalate~\citep{jung2025trust} guarantees human agreement of cascaded LLM judges through Clopper-Pearson-style UCB (\texttt{UCB-CLP}) computation~\citep{clopper1934use}. 
While these methods provide valid risk control through high-probability upper confidence bounds, they are often overly conservative because they enforce worst-case tail control over the empirical risk estimate. 
In contrast, \texttt{LEC} directly constrains the expectation of a linear functional of selection and error indicators, yielding tighter yet still statistically valid selection-conditioned risk control.

%% file: sections/method.tex
\section{Methodology}
\label{sec: Methodology}

\subsection{Notations and Problem Formulation}
\label{sec: Notations and Problem Formulation}

\noindent\textbf{\emph{1) Single-Model Selective Prediction with Selection-Conditioned Error Control.}} Let $\mathcal{G}^{(a)}: \mathcal{X} \rightarrow \mathcal{Y}$ denote a pretrained model that maps an input prompt to a textual output. 
For a given prompt $x \in \mathcal{X}$ with an unknown ground-truth answer $y^* \in \mathcal{Y}$, the model produces a prediction $\hat{y}^{(a)} = \mathcal{G}^{(a)}(x) \in \mathcal{Y}$. 
We quantify the model's uncertainty for $x$ as $u^{(a)} = \mathcal{U}(x; \mathcal{G}^{(a)})$, where $\mathcal{U}(\cdot)$ denotes a scalar uncertainty function. 
Intuitively, small $u^{(a)}$ indicates high trustworthiness in $\hat{y}^{(a)}$. 
For a specified threshold $\lambda^{(a)}$, the prediction $\hat{y}^{(a)}$ is deemed admissible and accepted if $u^{(a)} \le \lambda^{(a)}$. 
Let the admission function be
\[
A(y^*, y) =
\begin{cases}
1, & \text{if $y\in \mathcal{Y}$ is aligned with $y^*$},\\[4pt]
0, & \text{otherwise.}
\end{cases}
\]
However, prior uncertainty methods are inherently imperfect and cannot fully separate correct from incorrect outputs~\citep{liu2025uncertainty}. 
Thus, applying a fixed $\lambda^{(a)}$ at test time may admit some erroneous predictions. 
To mitigate this issue, our goal is to derive a statistically rigorous threshold $\hat{\lambda}^{(a)}$ that ensures the conditional probability that an accepted prediction is incorrect does not exceed a target risk level $\alpha$. 

Formally, we define the selection indicator as 
$S^{(a)}\left( \lambda^{(a)} \right) = \mathbf{1}\left\{ u^{(a)} \leq \lambda^{(a)} \right\}$, 
and the corresponding error indicator as 
$err^{(a)} = \mathbf{1}\left\{ A(y^*, \hat{y}^{(a)}) = 0 \right\}$. 
Our objective is to calibrate a statistically valid threshold $\hat{\lambda}^{(a)}$ such that
\begin{equation}
    \Pr \left( err^{(a)} = 1 \mid S^{(a)}(\hat{\lambda}^{(a)}) = 1 \right) \le \alpha, \quad \alpha \in (0,1).
    \label{eq:single-model goal}
\end{equation}
We refer to the left-hand side of Eq.~\eqref{eq:single-model goal} as the \emph{selection-conditioned error rate}: the marginal error probability of a prediction after it has been selected by the calibrated rule.

\noindent\textbf{\emph{2) Two-Model Routing with System-Level Selection-Conditioned Error Control.}} Under a specific uncertainty function $\mathcal{U}(\cdot)$, the uncertainty scores of model $\mathcal{G}^{(a)}$ on test examples may cluster too tightly in a low range, making it impossible to achieve small target risk levels. 
Moreover, when $\mathcal{G}^{(a)}$ has limited predictive ability, many challenging or critical prompts may be abstained from, leading to reduced system efficiency. 
To alleviate these issues, we develop a collaborative routing mechanism that dynamically delegates uncertain samples to another model with stronger accuracy or a more discriminative uncertainty profile, while controlling the system-level selection-conditioned error rate. 

Formally, we define the alternative model as $\mathcal{G}^{(b)}:\mathcal{X}\rightarrow\mathcal{Y}$. 
For a given prompt $x$, when the estimated uncertainty $u^{(a)}$ exceeds $\lambda^{(a)}$, we route the prompt to $\mathcal{G}^{(b)}$. 
We denote the prediction of $\mathcal{G}^{(b)}$ as $\hat{y}^{(b)} \in \mathcal{Y}$, along with the corresponding uncertainty $u^{(b)} = \mathcal{U}(x; \mathcal{G}^{(b)})$. 
Similarly, if $u^{(b)}$ does not exceed the threshold $\lambda^{(b)}$ of model $\mathcal{G}^{(b)}$, we trust $\hat{y}^{(b)}$; otherwise, the two-model routing system abstains from the prompt $x$. 
We define the selection indicator of model $\mathcal{G}^{(b)}$ as 
\[
S^{(b)}\left( \lambda^{(a)}, \lambda^{(b)} \right) 
= \mathbf{1}\left\{ u^{(a)} > \lambda^{(a)} \land  u^{(b)} \leq \lambda^{(b)} \right\},
\]
and the error indicator as 
$err^{(b)} = \mathbf{1}\left\{ A(y^*, \hat{y}^{(b)}) = 0 \right\}$. 

The two-model routing system $\mathcal{G}$ integrates $\mathcal{G}^{(a)}$ and $\mathcal{G}^{(b)}$, with the system-level selection indicator
\[
S\left( \lambda^{(a)}, \lambda^{(b)} \right) 
= S^{(a)}\left( \lambda^{(a)} \right) + S^{(b)}\left( \lambda^{(a)}, \lambda^{(b)} \right)\in \{0,1\}.
\]
The system-level accepted-error indicator is
\[
\begin{split}
    err = & ~ \mathbf{1}\{S^{(a)}(\lambda^{(a)})=1 \land 
 err^{(a)}=1\}\\
 &+ \mathbf{1}\{S^{(b)}(\lambda^{(a)},\lambda^{(b)})=1 \land 
 err^{(b)}=1\}.
\end{split}
\]
When $S(\lambda^{(a)}, \lambda^{(b)}) = 1$, the prediction from either $\mathcal{G}^{(a)}$ or $\mathcal{G}^{(b)}$ is accepted. 
We aim to jointly calibrate $(\lambda^{(a)}, \lambda^{(b)})$ and obtain statistically rigorous thresholds $(\hat{\lambda}^{(a)}, \hat{\lambda}^{(b)})$ such that 
\begin{equation}
    \Pr \left( err = 1 \mid S\left( \hat{\lambda}^{(a)}, \hat{\lambda}^{(b)} \right) = 1 \right) \leq \alpha, \quad \alpha \in (0, 1).
    \label{eq:two-model goal} 
\end{equation}
This guarantees that the overall two-model routing system performs selective prediction with system-level selection-conditioned error control.

\subsection{Threshold Calibration for Single-Model Settings}
\label{sec: Threshold Calibration for Single-Model Settings}
We begin by describing how to calibrate a statistically valid threshold $\hat{\lambda}^{(a)}$ for $\mathcal{G}^{(a)}$. 
Following the standard split calibration protocol~\citep{split2002conformal}, the dataset is partitioned into a calibration set and a test set.
The threshold is learned solely from the calibration data for a user-specified risk level $\alpha$, and is then fixed during test-time evaluation. 

\noindent \textbf{From selection-conditioned error control to linear expectation constraint.} 
For a fixed threshold $\lambda^{(a)}$, recall the selection and error indicators $S^{(a)}(\lambda^{(a)})$ and $err^{(a)}$.
We further define the joint indicator as 
$Z^{(a)}(\lambda^{(a)}) = S^{(a)}(\lambda^{(a)}) \cdot err^{(a)}$, 
which equals $1$ if and only if we accept the prediction and the model errs. 
The selection-conditioned error rate can then be written as
\begin{equation}
\begin{split}
    &\quad \ \mathrm{SCER}^{(a)}(\lambda^{(a)}) 
    = \Pr \big(err^{(a)}=1 \mid S^{(a)}(\lambda^{(a)})=1\big)\\
    &= \frac{\Pr \big(err^{(a)}=1 \land S^{(a)}(\lambda^{(a)})=1\big)}{\Pr \big( S^{(a)}(\lambda^{(a)})=1\big)} 
    = \frac{\mathbb{E}[Z^{(a)}(\lambda^{(a)})]}{\mathbb{E}[S^{(a)}(\lambda^{(a)})]}.
\end{split}
\label{eq:single-scer}
\end{equation}
As long as $\mathbb{E}[S^{(a)}(\lambda^{(a)})] > 0$, 
$\mathrm{SCER}^{(a)}(\lambda^{(a)}) \le \alpha$ is equivalent to a constraint on the expectation of a linear functional of the selection and error indicators:
\begin{equation}
\mathbb{E}\big[Z^{(a)}(\lambda^{(a)}) - \alpha S^{(a)}(\lambda^{(a)})\big] \le 0.
\label{eq:single-linear}
\end{equation}
Intuitively, the random variable $Z^{(a)} - \alpha S^{(a)}$ measures \emph{accepted error count minus $\alpha$ times selection count} on a single example; if its expectation is non-positive, then the marginal error probability conditioned on selection does not exceed~$\alpha$. 

\noindent \textbf{Finite-sample sufficient condition.} 
To enforce the population constraint in Eq.~\eqref{eq:single-linear} using only the calibration data, we derive a finite-sample sufficient condition. 
Let the calibration set be $\mathcal{D}_{\mathrm{cal}} = \{(u_i^{(a)}, err_i^{(a)})\}_{i=1}^n$, with $\{S^{(a)}_i\}_{i=1}^n$, and let $u_{(1)}^{(a)} \le \cdots \le u_{(n)}^{(a)}$ denote the calibration uncertainty scores sorted in ascending order, with corresponding error indicators $err_{(j)}^{(a)}$.  
For a candidate threshold $\lambda^{(a)}$, we define
\[
k^{(a)}(\lambda^{(a)}) = \#\{i : S^{(a)}_i (\lambda^{(a)}) =1 \} = \#\{i : u_i^{(a)} \le \lambda^{(a)}\}
\]
as the number of calibration data points that would be accepted at threshold $\lambda^{(a)}$. 
Motivated by the standard leave-one-out correction in distribution-free calibration, we use the following finite-sample sufficient condition, whose validity under exchangeability is established in Appendix~\ref{sec: proof of theorem 1}: 
\begin{equation}
    \textstyle\sum^{k^{(a)}(\lambda^{(a)})}_{j=1}
    \big( err_{(j)}^{(a)} - \alpha \big) \le -1.
    \label{eq:single-crc-emp}
\end{equation}
We then define the feasible set of thresholds at level $\alpha$ as
\begin{equation}
    \Lambda^{(a)}_\alpha 
    = \Big\{\lambda^{(a)} : \sum_{j=1}^{k^{(a)}(\lambda^{(a)})} \big( err_{(j)}^{(a)} - \alpha \big) \le -1 \Big\}.
\end{equation}

\noindent \textbf{Calibrated retention-maximizing threshold.}
Among all thresholds in $\Lambda_\alpha^{(a)}$, we choose the largest feasible one to maximize the acceptance rate: 
\begin{equation}
    \begin{split}
        \hat{\lambda}^{(a)} 
    &= \sup \Lambda^{(a)}_\alpha\\
    &= \sup\left\{\lambda^{(a)} : \sum_{j=1}^{k^{(a)}(\lambda^{(a)})} \big( err_{(j)}^{(a)} - \alpha \big) \le -1 \right\}.
    \end{split}
    \label{eq:max-crc-single}
\end{equation}
If $\Lambda^{(a)}_\alpha$ is empty, we declare the target risk level $\alpha$ infeasible for $\mathcal{G}^{(a)}$ and abstain from all samples at this level.

\begin{theorem}[Single-model selection-conditioned error control]
\label{thm:single-fdr}
Assume that calibration and test examples are exchangeable~\citep{angelopoulos2023conformal}.  
Let $\hat{\lambda}^{(a)}$ be defined by Eq.~\eqref{eq:max-crc-single} using $\mathcal{D}_{\mathrm{cal}}$. 
Then, for a new test sample $(x_{n+1},y_{n+1}^*)$ with $(u_{n+1}^{(a)}, err_{n+1}^{(a)})$, 
\[
\Pr\big( err_{n+1}^{(a)} = 1 \mid u_{n+1}^{(a)}\leq \hat{\lambda}^{(a)} \big) \le \alpha,
\]
where the probability is taken over the joint randomness of the calibration set and the test sample (marginal guarantee). 
If $\Pr(u_{n+1}^{(a)}\le \hat{\lambda}^{(a)})=0$, the guarantee is vacuous.
\end{theorem}

A complete proof of Theorem~\ref{thm:single-fdr} is given in Appendix~\ref{sec: proof of theorem 1}. 
At test time, for a new instruction $x_{n+1}$, we obtain the model prediction $\hat{y}_{n+1}^{(a)}$ with uncertainty $u_{n+1}^{(a)}$. 
We accept $\hat{y}_{n+1}^{(a)}$ if and only if $u_{n+1}^{(a)}\leq \hat{\lambda}^{(a)}$; otherwise, we abstain.

\subsection{Threshold Calibration for Two-Model Routing}
We now extend the above calibration procedure to the two-model routing system $\mathcal{G}$.
For each example $i$, we observe uncertainties $(u_i^{(a)}, u_i^{(b)})$ and error indicators $(err_i^{(a)}, err_i^{(b)})$. 
Given thresholds $(\lambda^{(a)},\lambda^{(b)})$, routing is defined by the selection indicators $S_i^{(a)}(\lambda^{(a)})$ and $S_i^{(b)}(\lambda^{(a)},\lambda^{(b)})$. 
The system-level selection indicator is 
\[
S_i(\lambda^{(a)},\lambda^{(b)}) 
= S_i^{(a)}(\lambda^{(a)}) + S_i^{(b)}(\lambda^{(a)},\lambda^{(b)}) \in \{0,1\}.
\]
The accepted-error indicator is
\[
\begin{split}
err_i = &\mathbf{1}\{S_i^{(a)}(\lambda^{(a)})=1\land\,err_i^{(a)}=1\} 
+\\& \mathbf{1}\{S_i^{(b)}(\lambda^{(a)},\lambda^{(b)})=1\land err_i^{(b)}=1\},
\end{split}
\]
which remains binary because routing selects at most one prediction. 
We also define the system-level joint indicator as
\[
Z_i(\lambda^{(a)},\lambda^{(b)})
= S_i^{(a)}(\lambda^{(a)}) \cdot err_i^{(a)}
+ S_i^{(b)}(\lambda^{(a)},\lambda^{(b)}) \cdot err_i^{(b)}.
\]

\noindent \textbf{From system-level selection-conditioned error control to expectation constraint.}
The system-level selection-conditioned error rate at thresholds $(\lambda^{(a)},\lambda^{(b)})$ is 
\[
\mathrm{SCER}(\lambda^{(a)},\lambda^{(b)})
= \frac{\mathbb{E}[Z(\lambda^{(a)},\lambda^{(b)})]}
       {\mathbb{E}[S(\lambda^{(a)},\lambda^{(b)})]}.
\]
Whenever $\mathbb{E}[S(\lambda^{(a)},\lambda^{(b)})]>0$, 
$\mathrm{SCER}(\lambda^{(a)},\lambda^{(b)}) \le \alpha$ is equivalent to a linear expectation inequality
\begin{equation}
\mathbb{E}\big[ Z(\lambda^{(a)},\lambda^{(b)}) - \alpha S(\lambda^{(a)},\lambda^{(b)}) \big] \le 0.
\label{eq:routing-linear}
\end{equation}
This condition generalizes the single-model constraint to the routing system and captures the difference between the system-level accepted-error count and the $\alpha$-fraction of accepted samples. 

\noindent\textbf{Finite-sample sufficient condition.} 
To enforce Eq.~\eqref{eq:routing-linear} from calibration points, we construct an empirical sufficient condition. 
Let $\mathcal{D}_{\mathrm{cal}}^{\mathrm{sys}} = \{(u_i^{(a)},u_i^{(b)},err_i^{(a)},err_i^{(b)})\}_{i=1}^n$ denote the calibration set for the two-model routing system. 
Using the same leave-one-out correction for the system-level pair $(Z_i,S_i)$, we obtain the following finite-sample sufficient condition, with the validity argument deferred to Appendix~\ref{sec: proof of theorem 2}:
\begin{equation}
    \sum_{i=1}^n 
    \Big(
      Z_i(\lambda^{(a)},\lambda^{(b)})
      - \alpha S_i(\lambda^{(a)},\lambda^{(b)})
    \Big)
    \le -1.
    \label{eq:routing-crc-emp}
\end{equation}
We then obtain the feasible set of two-model threshold pairs
\begin{equation}
\begin{split}
    \Lambda^{(a,b)}_\alpha
    =  \Big\{
       & (\lambda^{(a)},\lambda^{(b)}): \sum_{i=1}^n 
    \Big(
      Z_i(\lambda^{(a)},\lambda^{(b)})
      \\& - \alpha S_i(\lambda^{(a)},\lambda^{(b)})
    \Big)
    \le -1
      \Big\}.
\end{split}
\label{eq:routing feasible set}
\end{equation}

\noindent \textbf{Calibrated retention-maximizing thresholds.}
Among all pairs $(\lambda^{(a)},\lambda^{(b)}) \in \Lambda^{(a,b)}_\alpha$, we choose those that maximize the empirical acceptance rate of the routing system:
\begin{equation}
    (\hat{\lambda}^{(a)},\hat{\lambda}^{(b)}) 
    = \operatorname*{argmax}_{(\lambda^{(a)},\lambda^{(b)}) \in \Lambda^{(a,b)}_\alpha}
    \frac{1}{n}\sum_{i=1}^n S_i(\lambda^{(a)},\lambda^{(b)}).
    \label{eq:max-routing}
\end{equation}
If $\Lambda^{(a,b)}_\alpha$ is empty, the risk level $\alpha$ is infeasible for the two-model routing system, and the system abstains on all inputs. 

\begin{theorem}[Selection-conditioned error control for the two-model routing system]
\label{thm:routing-fdr}
Assume calibration and test examples are exchangeable. 
Let $(\hat{\lambda}^{(a)},\hat{\lambda}^{(b)})$ be any solution of Eq.~\eqref{eq:max-routing}.
Then the two-model routing system satisfies
\[
\Pr\big( err_{n+1} = 1 \mid S_{n+1}(\hat{\lambda}^{(a)},\hat{\lambda}^{(b)}) = 1 \big) \le \alpha,
\]
where the probability is taken over the joint randomness of calibration and test samples (marginal guarantee). 
If $\Pr(S_{n+1}(\hat{\lambda}^{(a)},\hat{\lambda}^{(b)})=1)=0$, the guarantee is vacuous.
\end{theorem}

See a proof of Theorem~\ref{thm:routing-fdr} in Appendix~\ref{sec: proof of theorem 2}. 
At test time, each user instruction $x_{n+1}$ is processed as follows:
we accept $\hat{y}^{(a)}_{n+1}$ via $\mathcal{G}^{(a)}$ if $u_{n+1}^{(a)}\le\hat{\lambda}^{(a)}$; otherwise we route the prompt to $\mathcal{G}^{(b)}$ and accept $\hat{y}^{(b)}_{n+1}$ if $u_{n+1}^{(b)}\le\hat{\lambda}^{(b)}$.
If neither condition is satisfied, the system abstains. 
Our analysis highlights that system-level selection-conditioned error control is preserved as long as the routing policy is deterministic and each example is routed to at most one model. 
The statistical guarantees arise from the linear decomposition, rather than any model-specific assumptions.

The above two-model calibration can readily be extended to routing systems with more than two models. 
In Appendix~\ref{sec: Extension to General Multi-Model Routing Systems}, we outline how \texttt{LEC} extends to general multi-model routing systems, offering a principled mechanism for unified selection-conditioned error control across routing policies of arbitrary depth.

%% file: sections/experiments.tex
\section{Experiments}
\label{sec: Experiments}

\subsection{Experimental Settings}
\label{sec: Experimental Settings}
\noindent \textbf{Benchmarks and Models.} (1) QA: We evaluate \texttt{LEC} on the CommonsenseQA (closed-ended)~\citep{talmor-etal-2019-commonsenseqa} and TriviaQA (open-ended)~\citep{joshi-etal-2017-triviaqa} datasets using eight LLMs, including LLaMA~\citep{touvron2023llama}, Qwen~\citep{bai2023qwen}, Vicuna~\citep{zheng2023judging}, and OpenChat~\citep{wang2024openchat} families. 
(2) VQA: We also consider the ScienceQA (closed-ended)~\citep{lu2022learn} and MM-Vet v2 (open-ended)~\citep{yu2024mm} benchmarks, using four LVLMs, including LLaVA1.5~\citep{liu2023visual}, LLaVA-NeXT~\citep{liu2024improved}, and InternVL2~\citep{chen2024internvl} groups. 
We omit suffixes such as ``hf'' and ``Instruct''. 

\begin{figure*}[!t]
  \centering
  \begin{subfigure}[c]{0.24\textwidth}
    \centering
    \includegraphics[width=\textwidth]{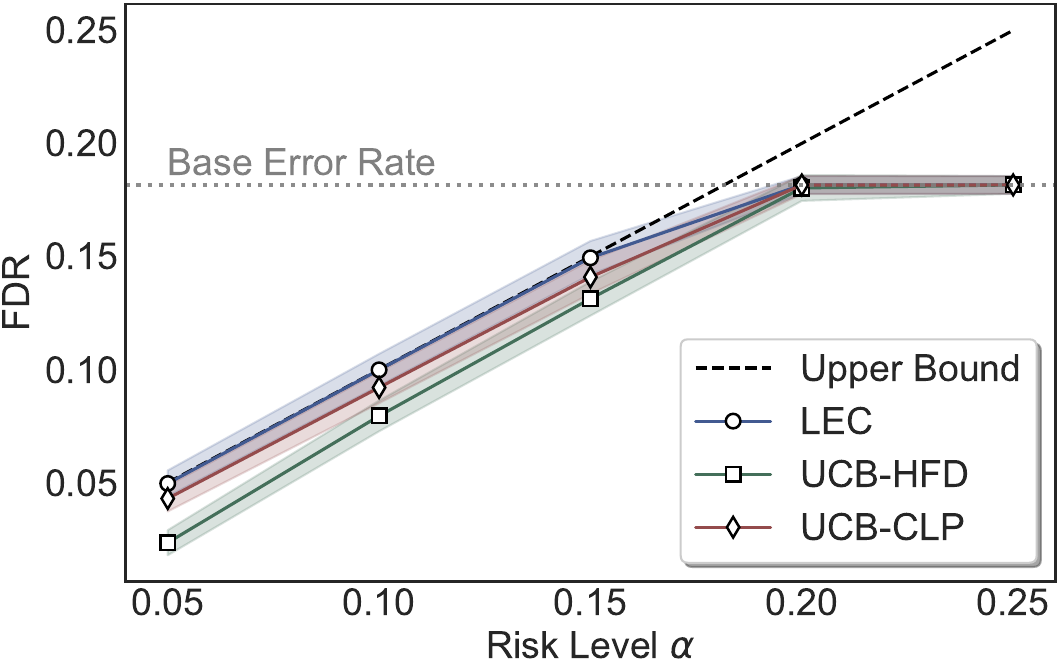}
    \caption{OpenChat-3.5.}
  \end{subfigure}
  \hfill
  \begin{subfigure}[c]{0.24\textwidth}
    \centering
    \includegraphics[width=\textwidth]{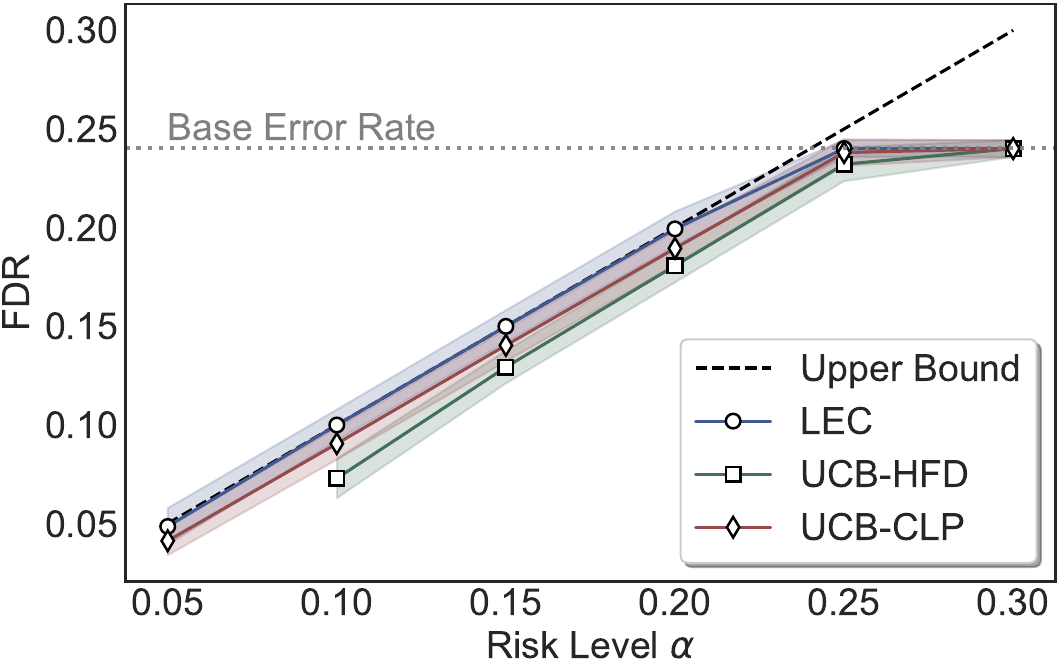}
    \caption{Qwen2.5-3B.}
  \end{subfigure}
  \hfill
  \begin{subfigure}[c]{0.24\textwidth}
    \centering
    \includegraphics[width=\textwidth]{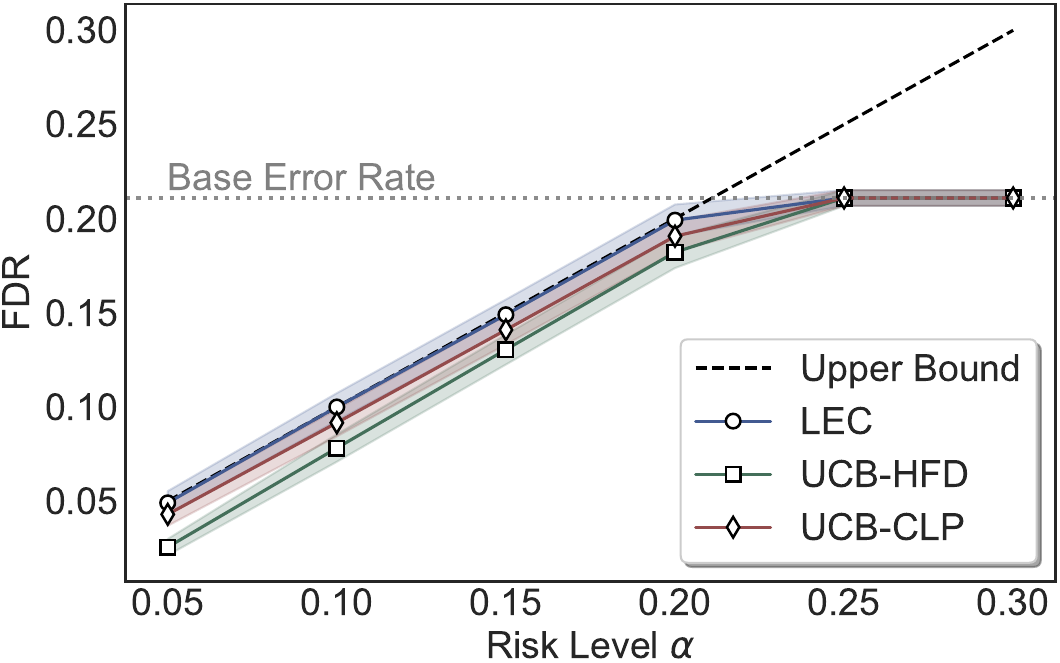}
    \caption{Qwen2.5-7B}
  \end{subfigure}
  \hfill
  \begin{subfigure}[c]{0.24\textwidth}
    \centering
    \includegraphics[width=\textwidth]{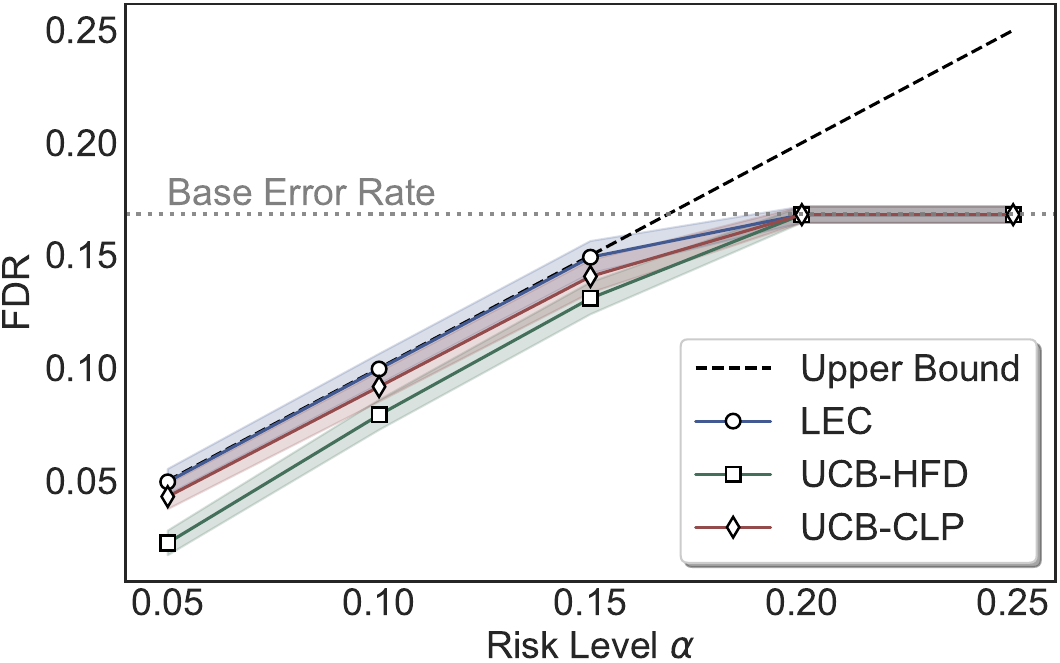}
    \caption{Qwen2.5-14B.}
  \end{subfigure}
    
  \begin{subfigure}[c]{0.24\textwidth}
    \centering
    \includegraphics[width=\textwidth]{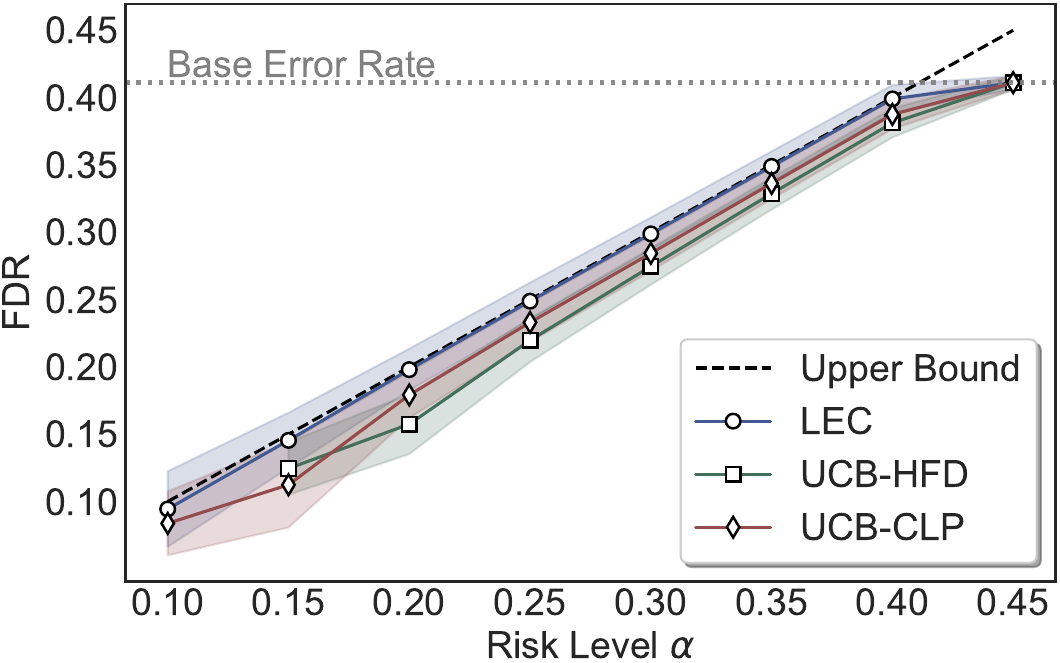}
    \caption{Vicuna-7B-V1.5.}
  \end{subfigure}
  \hfill
  \begin{subfigure}[c]{0.24\textwidth}
    \centering
    \includegraphics[width=\textwidth]{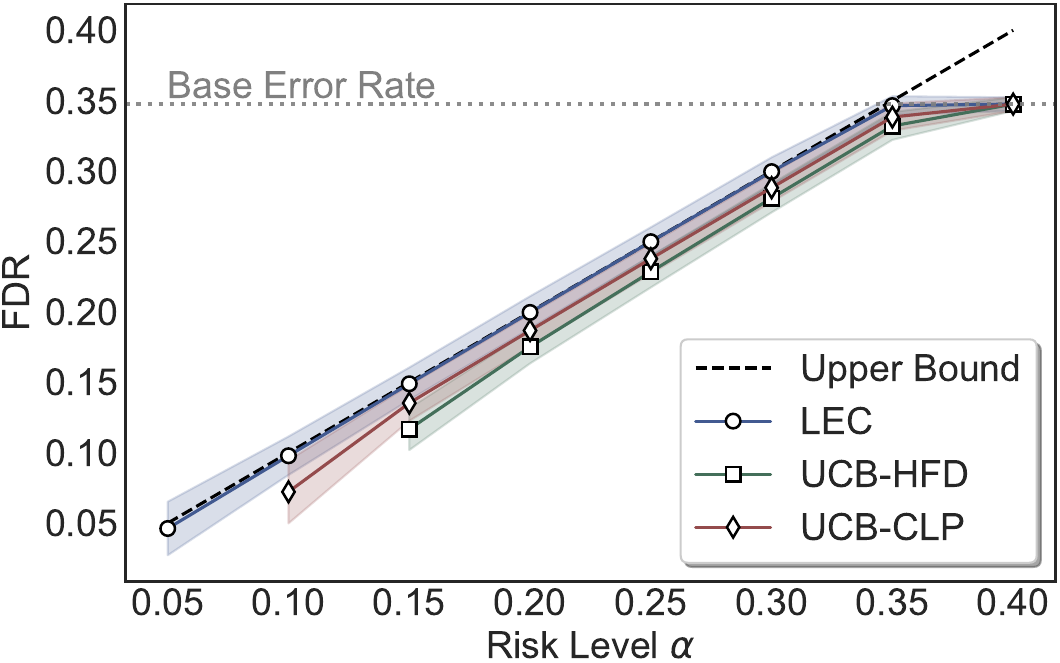}
    \caption{Vicuna-13B-V1.5.}
  \end{subfigure}
  \hfill
  \begin{subfigure}[c]{0.24\textwidth}
    \centering
    \includegraphics[width=\textwidth]{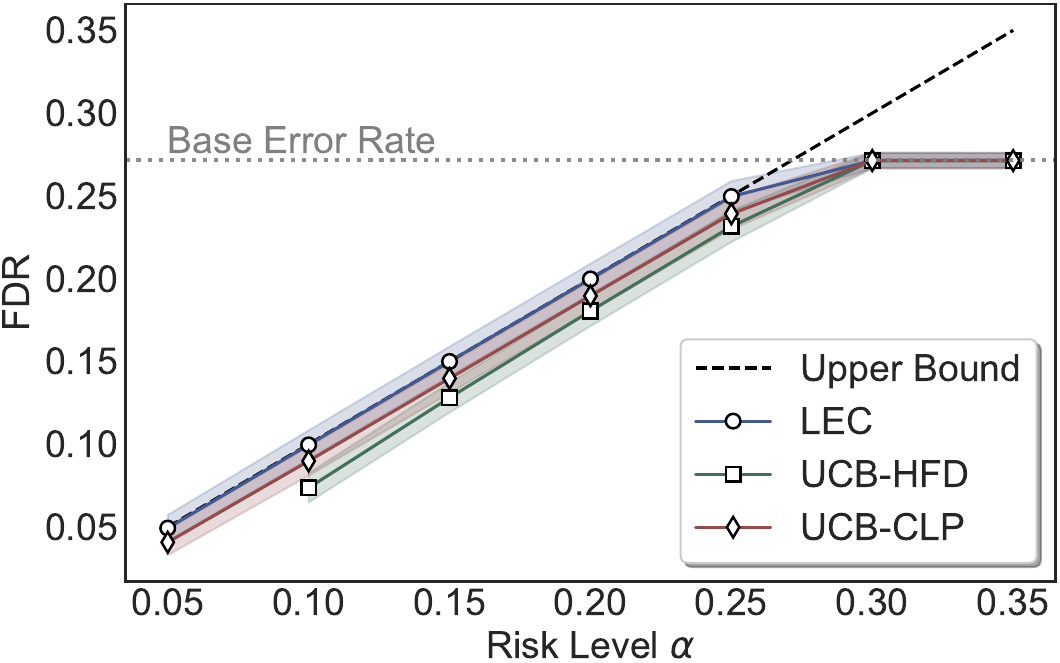}
    \caption{LLaMA-3.1-8B.}
  \end{subfigure}
  \hfill
  \begin{subfigure}[c]{0.24\textwidth}
    \centering
    \includegraphics[width=\textwidth]{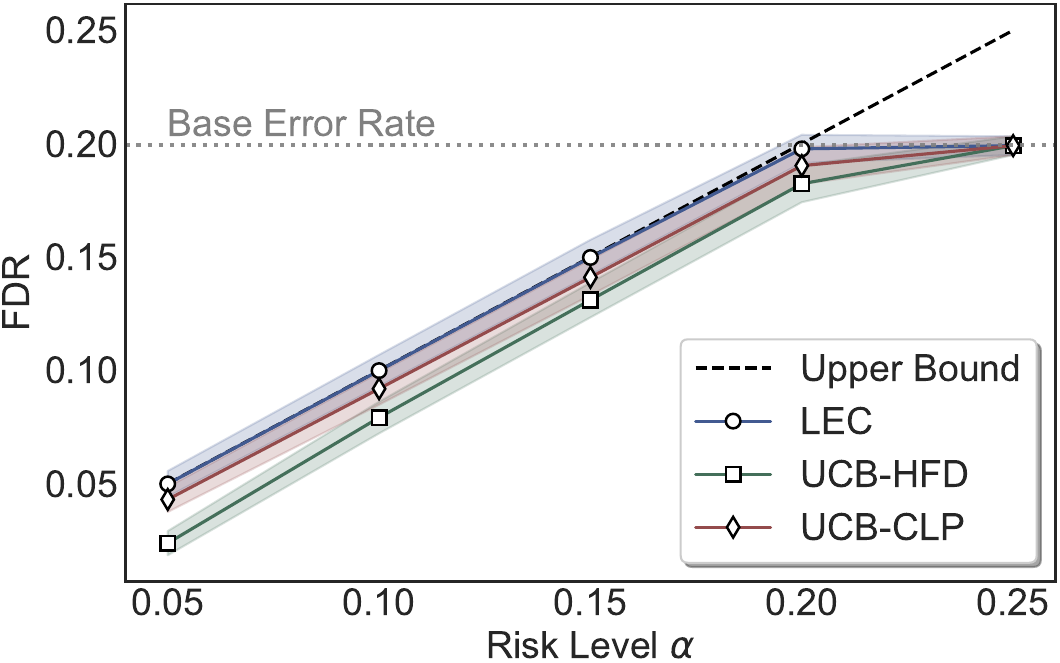}
    \caption{LLaMA-3.1-70B.}
  \end{subfigure}
  \caption{Test-time empirical selection-conditioned error rate on the CommonsenseQA dataset (mean$\pm$std). The y-axis label ``FDR'' denotes the observed fraction of erroneous predictions among accepted predictions. \texttt{LEC} provides tighter risk control while maintaining the prescribed risk level.}
  \label{fig: Test-time FDR on the CommonsenseQA dataset (mean±std).}
\end{figure*}

\begin{figure*}[!t]
  \centering
  \begin{subfigure}[c]{0.24\textwidth}
    \centering
    \includegraphics[width=\textwidth]{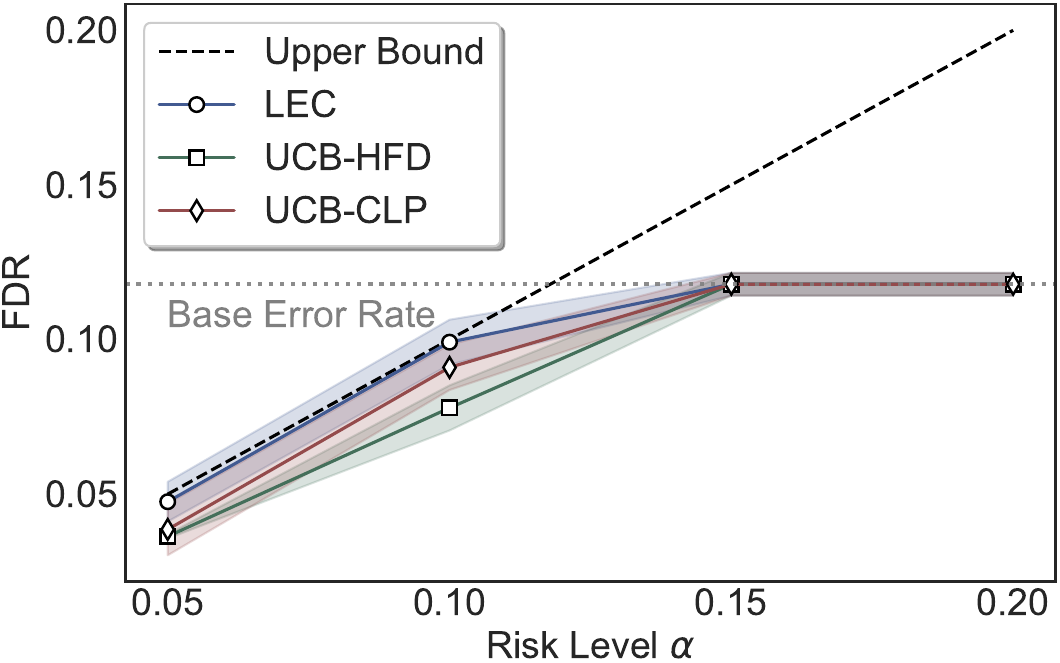}
    \caption{OpenChat-3.5.}
  \end{subfigure}
  \hfill
  \begin{subfigure}[c]{0.24\textwidth}
    \centering
    \includegraphics[width=\textwidth]{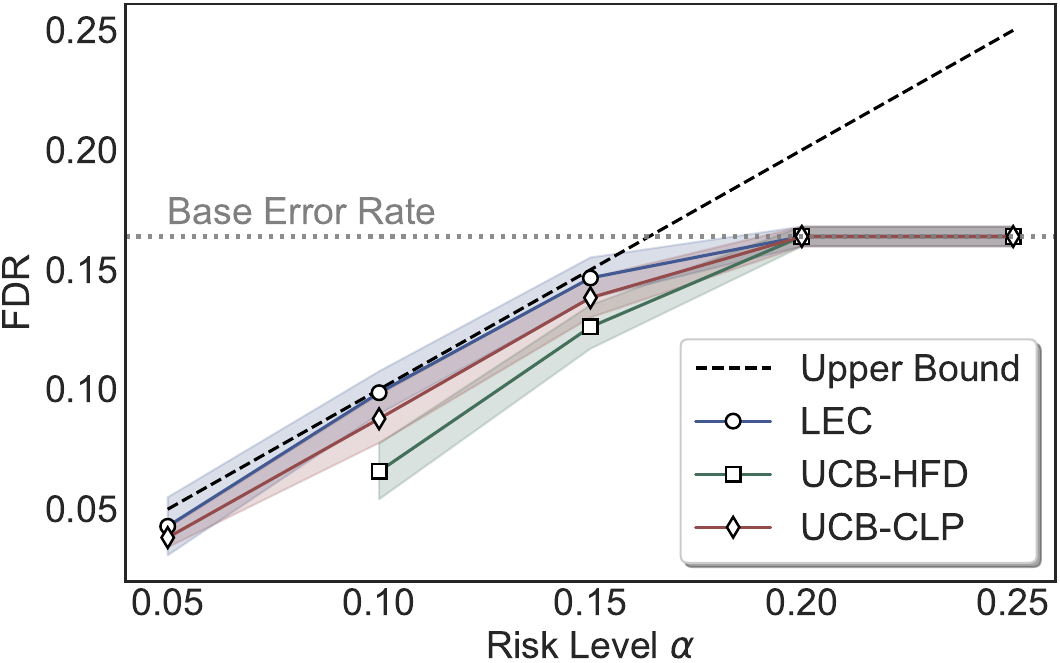}
    \caption{Qwen2.5-3B.}
  \end{subfigure}
  \hfill
  \begin{subfigure}[c]{0.24\textwidth}
    \centering
    \includegraphics[width=\textwidth]{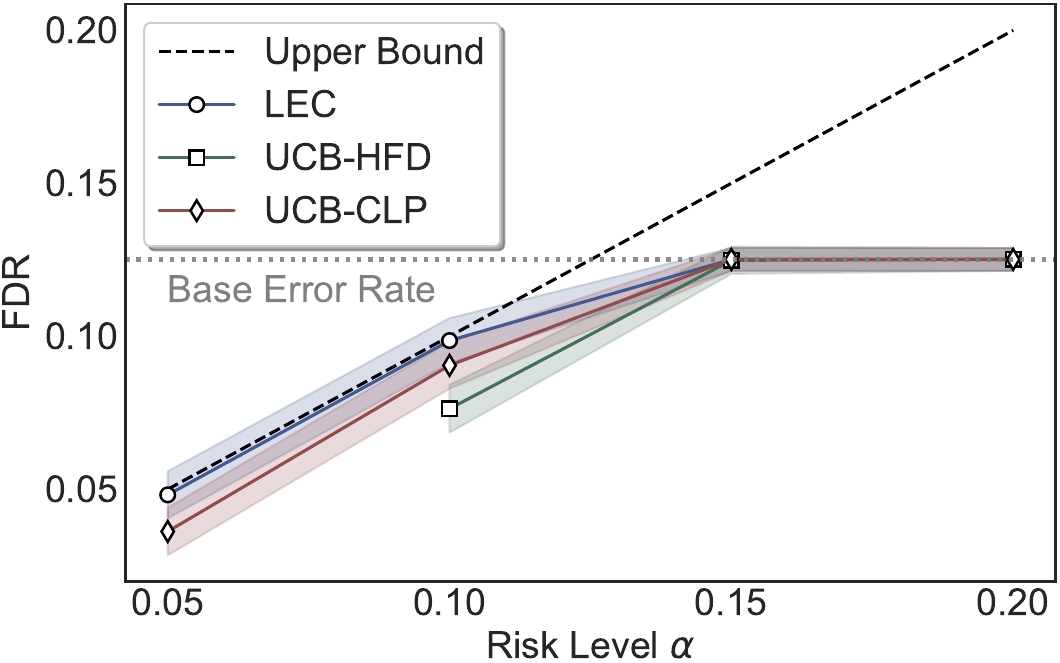}
    \caption{Qwen2.5-7B}
  \end{subfigure}
  \hfill
  \begin{subfigure}[c]{0.24\textwidth}
    \centering
    \includegraphics[width=\textwidth]{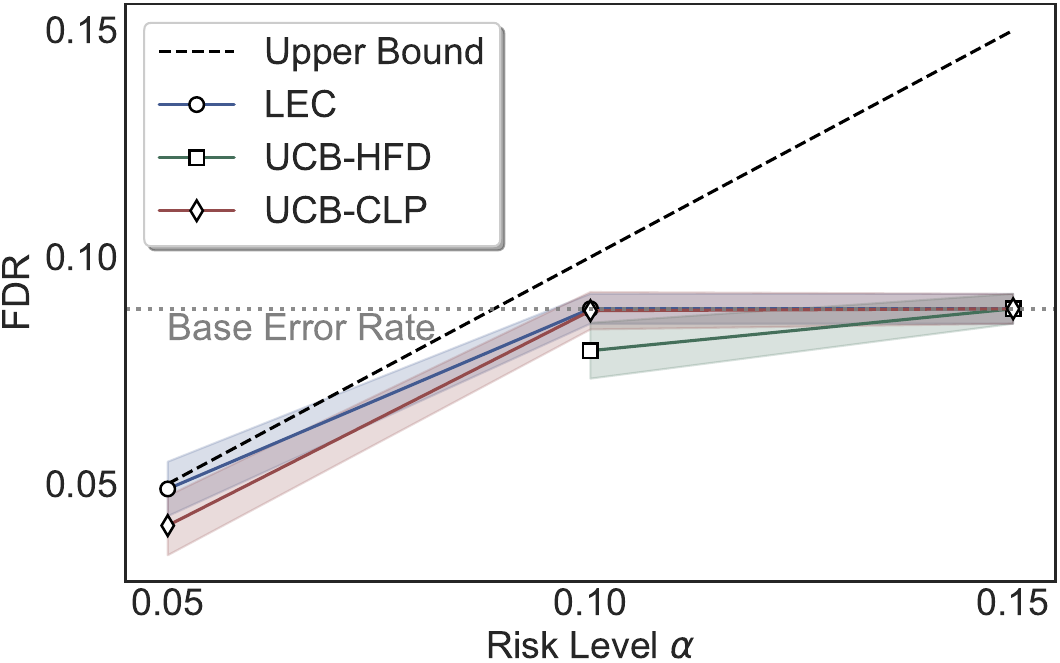}
    \caption{Qwen2.5-14B.}
  \end{subfigure}

  \begin{subfigure}[c]{0.24\textwidth}
    \centering
    \includegraphics[width=\textwidth]{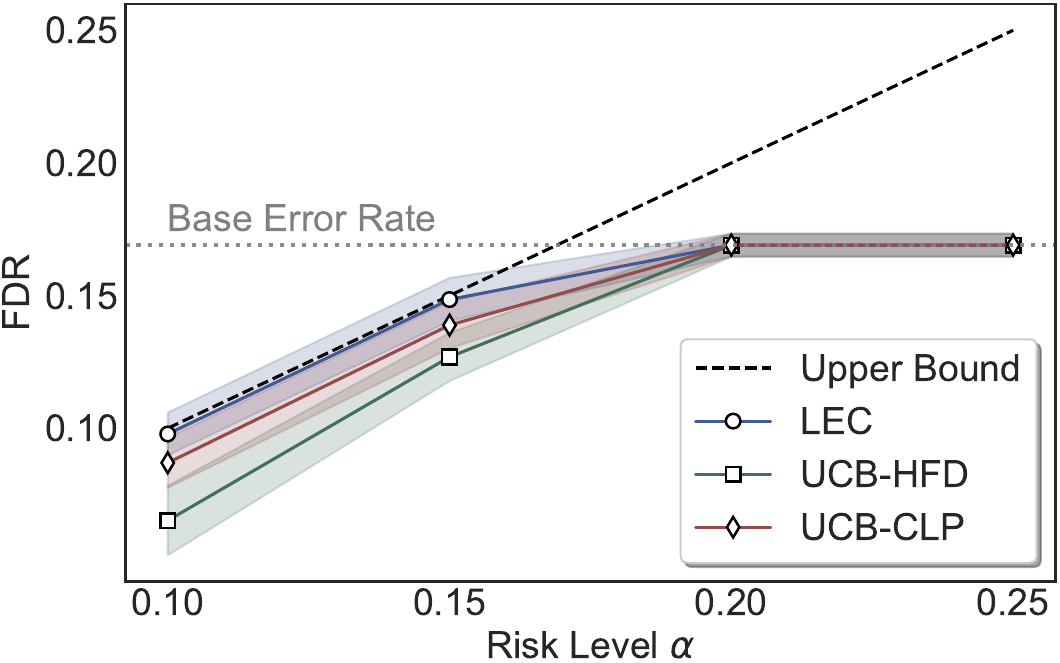}
    \caption{Vicuna-7B-V1.5.}
  \end{subfigure}
  \hfill
  \begin{subfigure}[c]{0.24\textwidth}
    \centering
    \includegraphics[width=\textwidth]{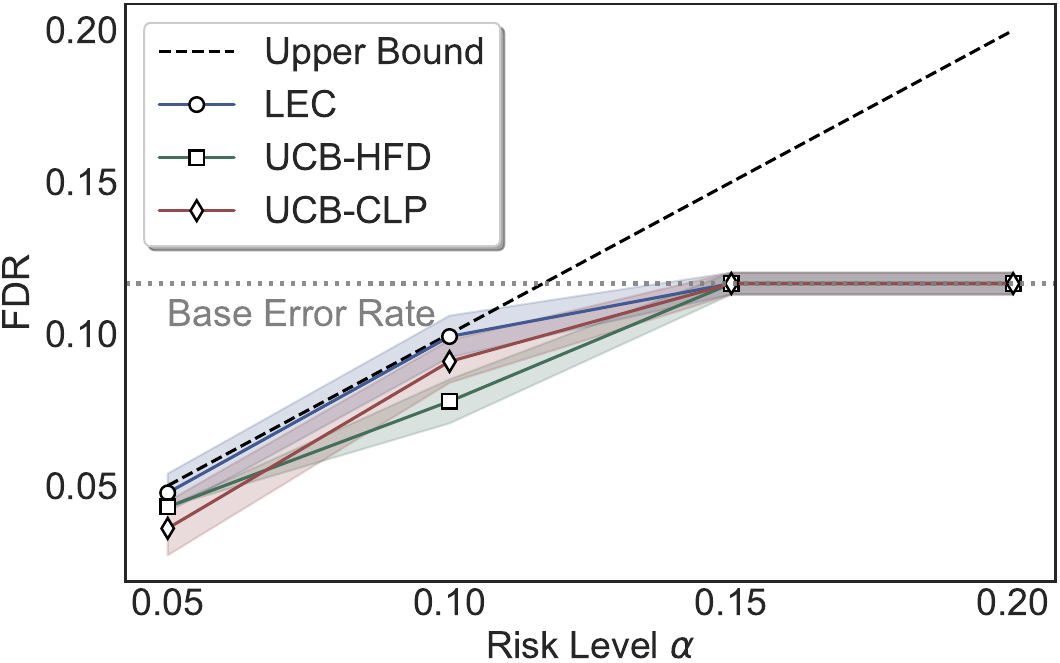}
    \caption{Vicuna-13B-V1.5.}
  \end{subfigure}
  \hfill
  \begin{subfigure}[c]{0.24\textwidth}
    \centering
    \includegraphics[width=\textwidth]{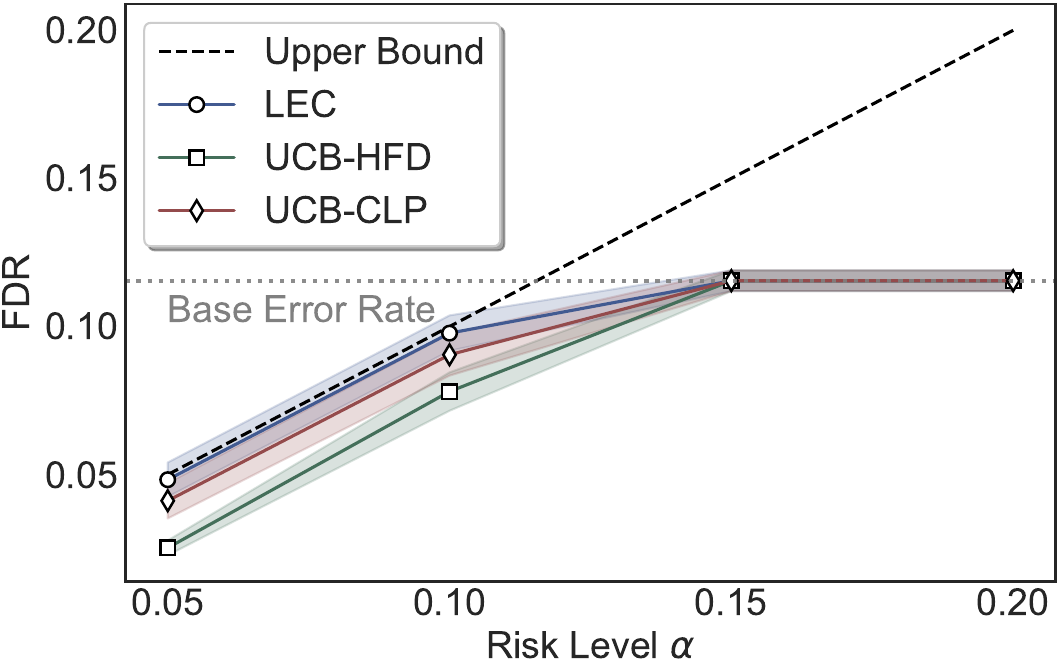}
    \caption{LLaMA-3.1-8B.}
  \end{subfigure}
  \hfill
  \begin{subfigure}[c]{0.24\textwidth}
    \centering
    \includegraphics[width=\textwidth]{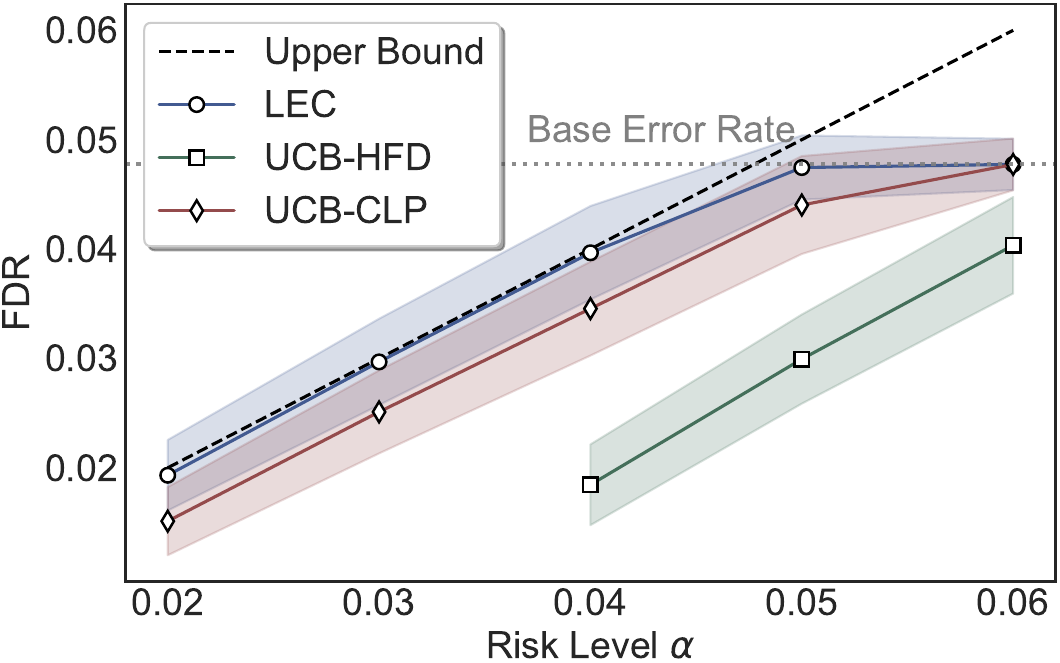}
    \caption{LLaMA-3.1-70B.}
  \end{subfigure}
  \caption{Test-time empirical selection-conditioned error rate on the TriviaQA dataset (mean$\pm$std). \texttt{LEC} provides tighter risk control while maintaining the prescribed risk level.}
  \label{fig: Test-time FDR on the TriviaQA dataset (mean±std).}
\end{figure*}

\noindent \textbf{Evaluation Metrics.} 
Following previous evaluation protocols~\citep{jung2025trust,wang2025coin}, we evaluate the statistical validity of \texttt{LEC} by verifying that the empirical error rate among accepted predictions does not exceed the target risk level. 
Our theoretical target is the selection-conditioned error rate,
\[
\mathrm{SCER}(\lambda)
=
\Pr(err=1\mid S(\lambda)=1)
=
\frac{\mathbb{E}[S(\lambda)\cdot err]}{\mathbb{E}[S(\lambda)]}.
\]
In the experimental figures, the y-axis label ``FDR'' is retained as a compact plotting shorthand for the observed false-discovery proportion among accepted predictions:
\[
\widehat{\mathrm{FDP}}_{\mathrm{acc}}
=
\frac{\sum_{i\in\mathcal{D}_{\mathrm{test}}} S_i(\hat{\lambda})\,err_i}
{\left(\sum_{i\in\mathcal{D}_{\mathrm{test}}} S_i(\hat{\lambda})\right)\vee 1}.
\]
This empirical quantity is utilized only to summarize test-time performance and serves as a plug-in estimate of the selection-conditioned error rate controlled by our theory. 
Throughout the paper, all theoretical guarantees are stated in terms of selection-conditioned error control. 
We further assess power, defined as the proportion of aligned test predictions accepted by the method among all aligned predictions. 
In two-model routing, we additionally report the allocation ratio of accepted samples across the two models.

\noindent \textbf{Baselines.}
\textit{1) Single-model:} 
We consider UCB-based methods that control the accepted error rate by computing UCBs on the system risk from calibration data. 
Specifically, we implement two variants:
\texttt{UCB-HFD}, which derives the UCB using Hoeffding's inequality~\citep{hoeffding1963probability},
and \texttt{UCB-CLP}, which adopts the exact Clopper--Pearson interval. 
These two variants abstract the core confidence-bound-based risk control mechanism used in prior single-model methods such as COIN~\citep{wang2025coin}. 
\textit{2) Two-model routing:} We extend the UCB-based approach to the routing setting by applying the same confidence-bound-based risk control to the system-level
selection and error indicators. 
We consider \texttt{UCB-CLP-Routing}, corresponding to the cascaded judge in \citet{jung2025trust}, as well as \texttt{UCB-HFD-Routing}, which replaces the Clopper-Pearson bound with Hoeffding's inequality for a distribution-free variant. 
We do not consider routing with more than two models, as it only increases the number of threshold parameters and leads to nested threshold searches during calibration, without altering the formulation or its statistical guarantees.


\noindent \textbf{Alignment Criteria.} 
We use sentence similarity~\citep{reimers2019sentence} with a 0.6 threshold to decide whether the model's answer is aligned with the ground truth in the admission function $A$ by default. 
We also use bi-entailment~\citep{kuhn2023semantic} and LLM-as-a-Judge~\citep{zhang2024vl}.

\noindent \textbf{Uncertainty Estimator $\mathcal{U}$.} 
In closed-ended QA and VQA, we estimate uncertainty scores by computing the predictive entropy (PE)~\citep{kadavath2022language}. 
We use the softmax output of model logits by default. 
We also generate multiple answers per input and employ sampling frequency as the generative probability~\citep{wang-etal-2025-sconu}. 
In open-ended QA and VQA, we compute the black-box semantic entropy (SE)~\citep{farquhar2024detecting} by default. 
Moreover, we use the sum of eigenvalues of the graph laplacian (EigV), degree matrix (Deg), and eccentricity (Ecc)~\citep{lin2024generating}. 
We also consider the length-normalized PE~\citep{malinin2021uncertainty} of the model's output itself (SELF).

\noindent \textbf{Hyperparameters.} 
Following previous work~\citep{wang2025coin}, we employ beam search (\texttt{num\_beams=5}) to obtain the most likely generation as the model output. 
By default, for open-domain QA, we sample 10 answers per input for UQ. 
In addition, we fix the calibration-test split ratio to 0.5. 

We provide the details of additional experimental settings in Appendix~\ref{sec: Additional Experimental Settings}. 
Following prior research~\citep{quach2024conformal}, we randomly split the calibration and test samples 500 times and report the mean and standard deviation (mean±std). 
We annotate this information alongside the subsequent results.

\begin{figure*}[!t]
  \centering
  \begin{subfigure}[b]{0.19\textwidth}
    \centering
    \includegraphics[width=\textwidth]{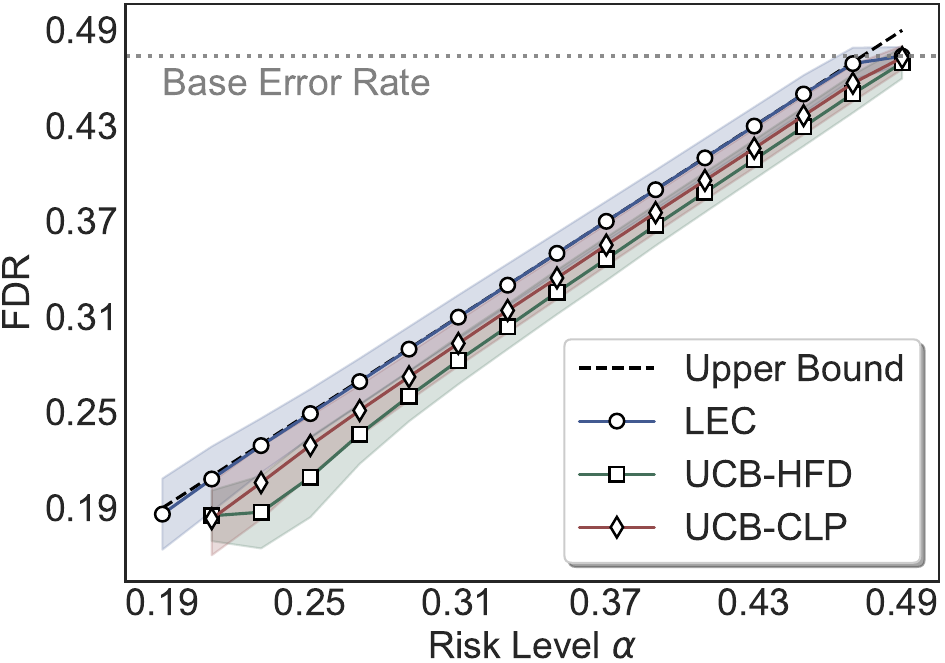}
    \caption{Qwen2.5-3B.}
  \end{subfigure}
  \hfill
  \begin{subfigure}[b]{0.19\textwidth}
    \centering
    \includegraphics[width=\textwidth]{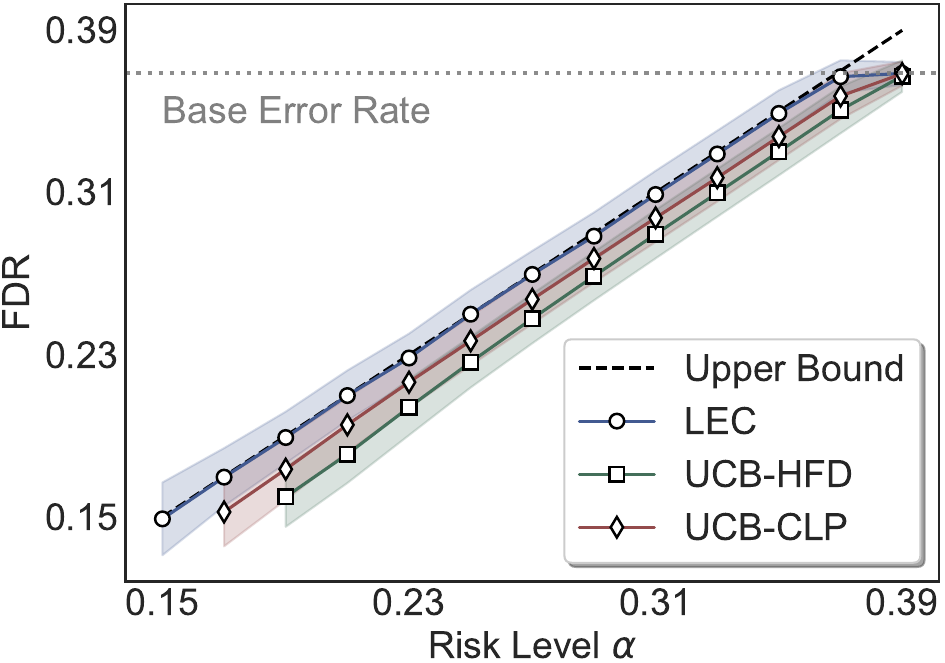}
    \caption{Qwen2.5-7B.}
  \end{subfigure}
  \hfill
  \begin{subfigure}[b]{0.19\textwidth}
    \centering
    \includegraphics[width=\textwidth]{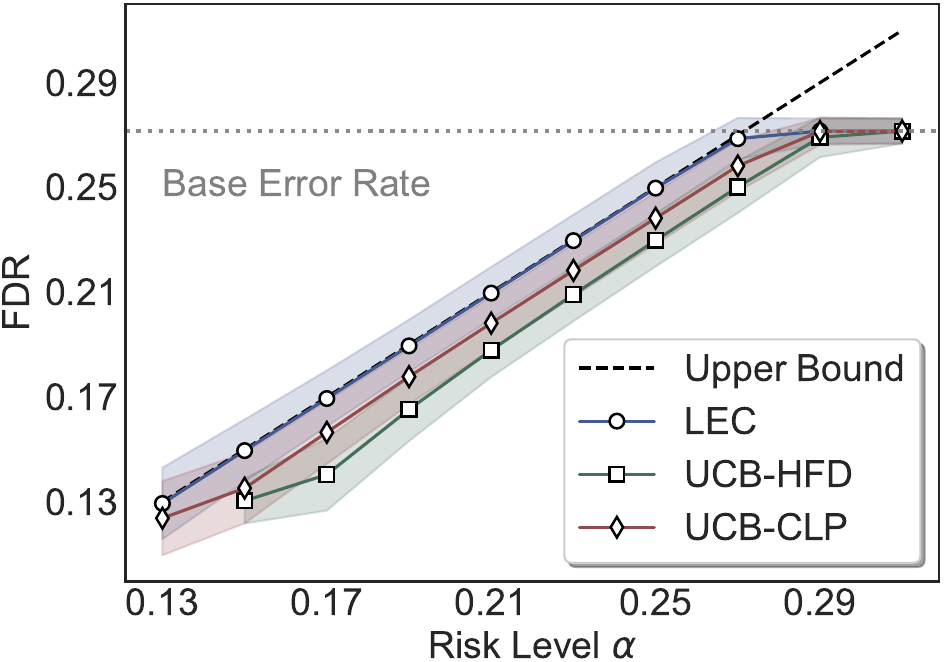}
    \caption{Qwen2.5-14B.}
  \end{subfigure}
  \hfill
  \begin{subfigure}[b]{0.19\textwidth}
    \centering
    \includegraphics[width=\textwidth]{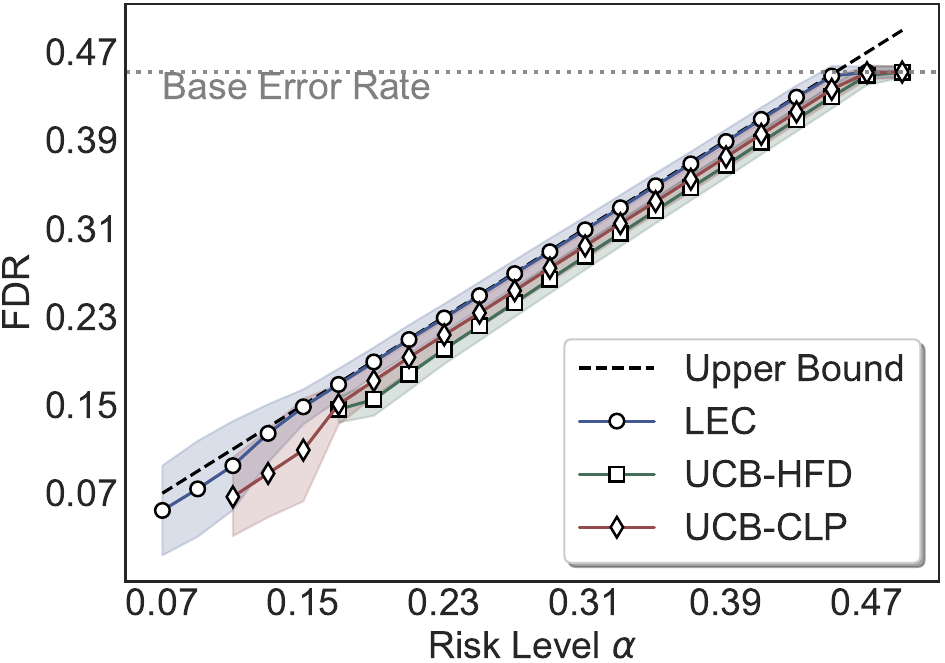}
    \caption{Vicuna-7B-V1.5.}
  \end{subfigure}
  \hfill
  \begin{subfigure}[b]{0.19\textwidth}
    \centering
    \includegraphics[width=\textwidth]{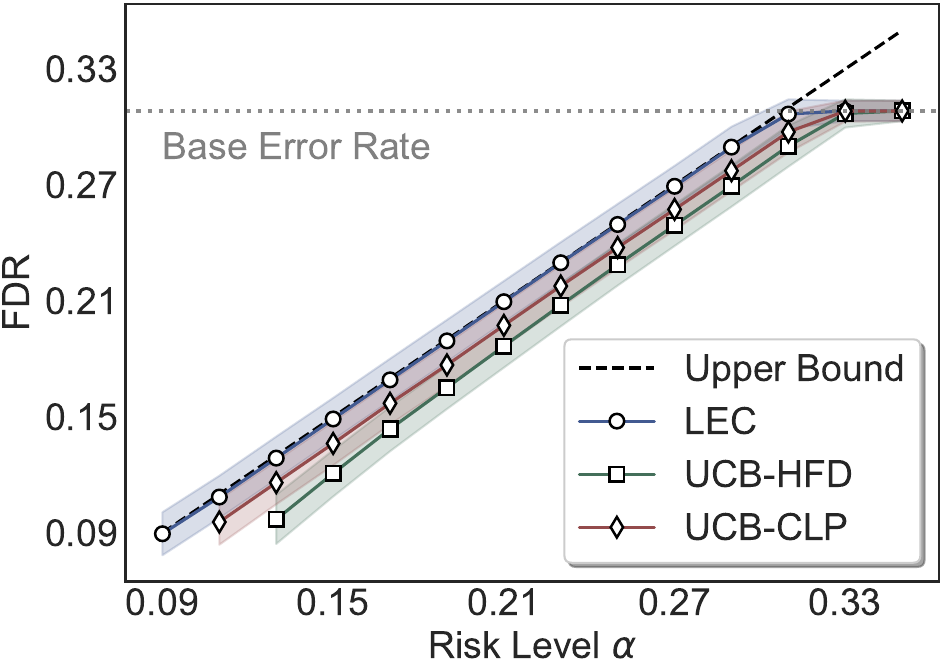}
    \caption{Vicuna-13B-V1.5.}
  \end{subfigure}
      \caption{Test-time empirical selection-conditioned error rate on TriviaQA with entailment for correctness evaluation (mean$\pm$std). }
  \label{fig: Test-time FDR (mean±std) on the TriviaQA dataset with entailment for correctness evaluation.}
\end{figure*}

\begin{figure*}[!t]
  \centering
 \begin{subfigure}[b]{0.19\textwidth}
    \centering
    \includegraphics[width=\textwidth]{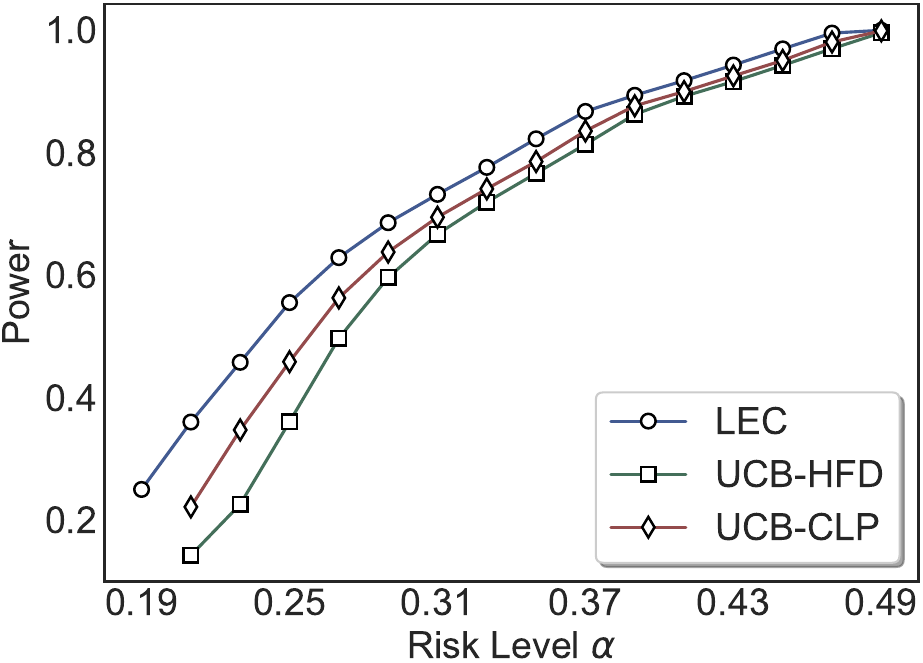}
    \caption{Qwen2.5-3B.}
  \end{subfigure}
  \hfill
  \begin{subfigure}[b]{0.19\textwidth}
    \centering
    \includegraphics[width=\textwidth]{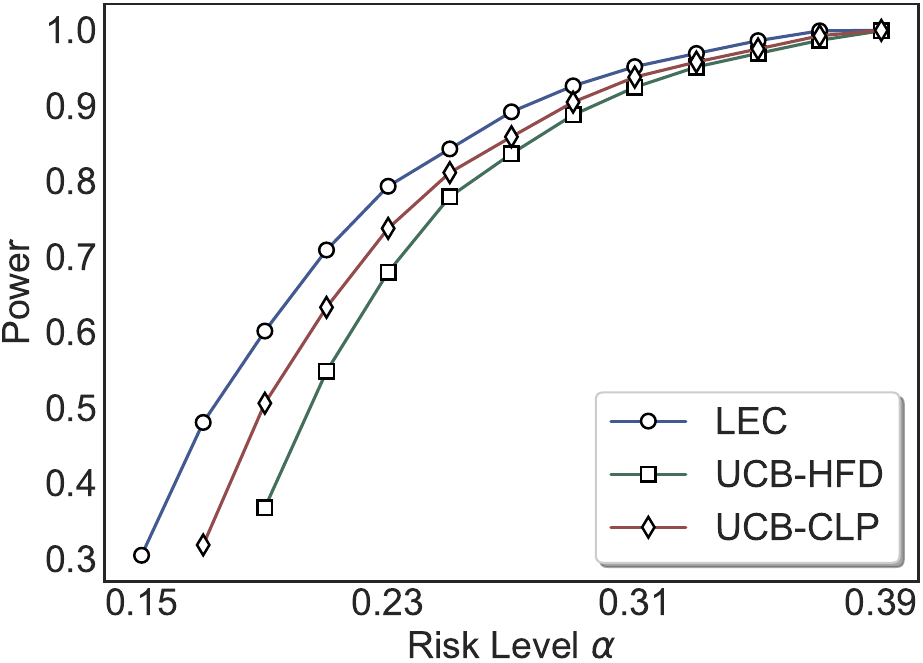}
    \caption{Qwen2.5-7B.}
  \end{subfigure}
  \hfill
  \begin{subfigure}[b]{0.19\textwidth}
    \centering
    \includegraphics[width=\textwidth]{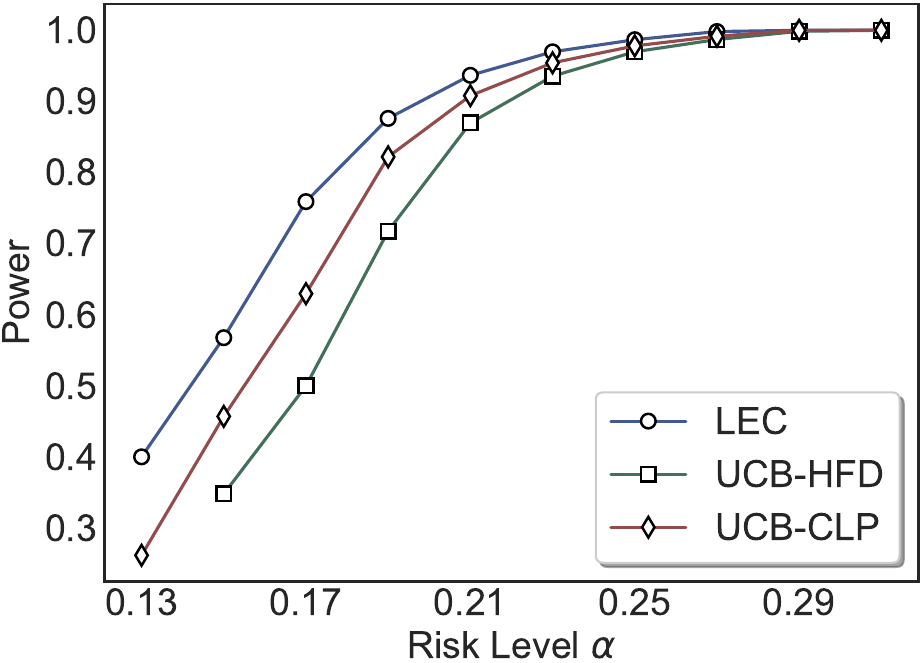}
    \caption{Qwen2.5-14B.}
  \end{subfigure}
  \hfill
  \begin{subfigure}[b]{0.19\textwidth}
    \centering
    \includegraphics[width=\textwidth]{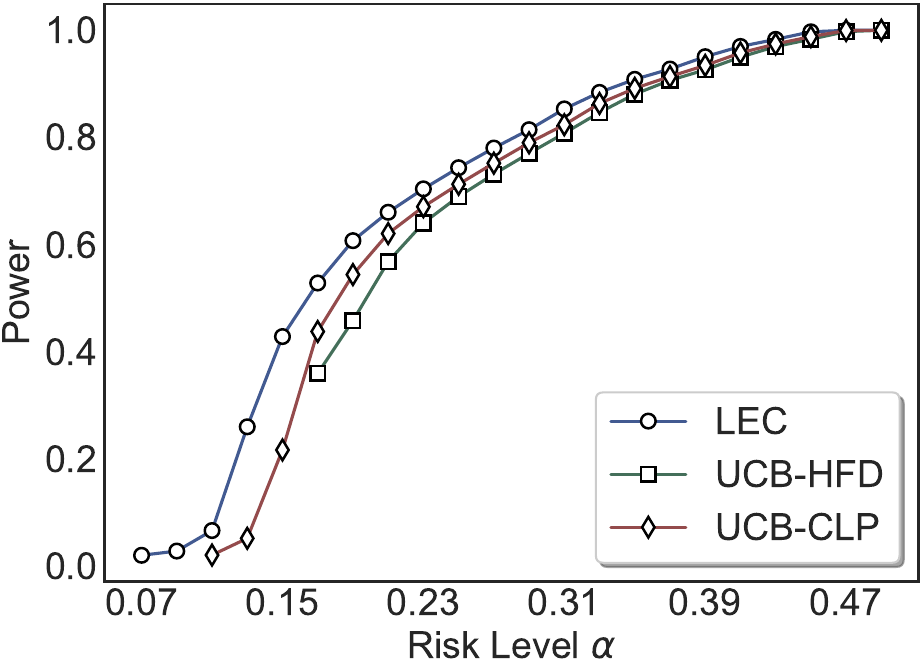}
    \caption{Vicuna-7B-V1.5.}
  \end{subfigure}
  \hfill
  \begin{subfigure}[b]{0.19\textwidth}
    \centering
    \includegraphics[width=\textwidth]{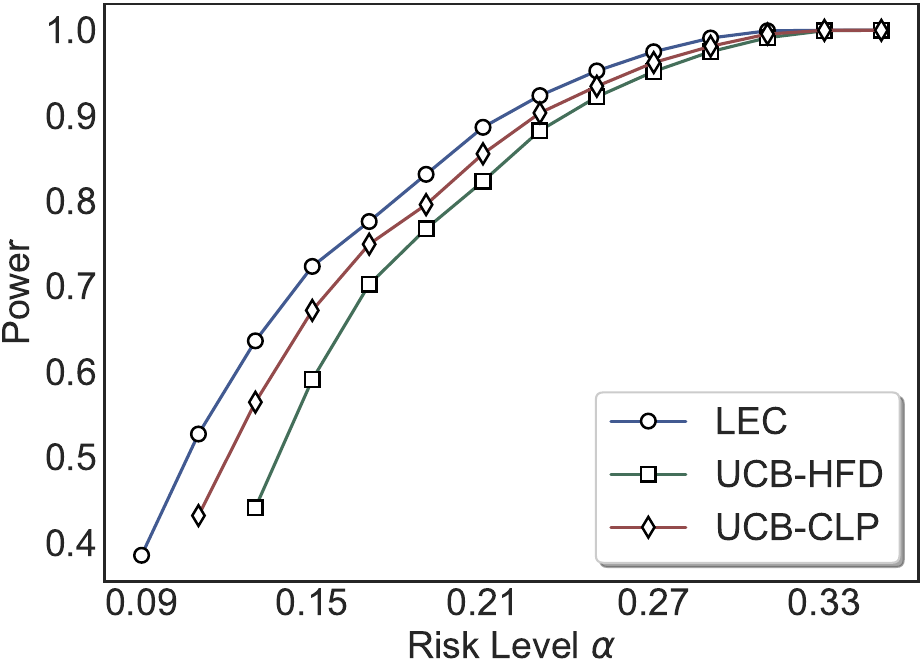}
    \caption{Vicuna-13B-V1.5.}
  \end{subfigure}
      \caption{Test-time Power on the TriviaQA dataset with entailment for correctness evaluation (mean).}
  \label{fig: Test-time Power (mean) on the TriviaQA dataset with entailment for correctness evaluation.}
\end{figure*}

\input{tables/powerTriviaQASimilarity0.6Entailment-Seb}

\subsection{Evaluations in Single-Model Selective Prediction}
\label{sec: Evaluations in Single-Model Selective Prediction}

\noindent \textbf{Statistical Validity.} 
We first evaluate \texttt{LEC} in single-model selective prediction settings. 
As demonstrated in Figures~\ref{fig: Test-time FDR on the CommonsenseQA dataset (mean±std).} and~\ref{fig: Test-time FDR on the TriviaQA dataset (mean±std).}, across both CommonsenseQA and TriviaQA datasets and eight LLMs, \texttt{LEC} consistently matches the prescribed selection-conditioned risk target: the empirical accepted error rate, averaged over 500 random splits, remains below the target risk level. 
For example, on CommonsenseQA with a risk level of 0.05, \texttt{LEC} achieves an average empirical accepted error rate of 0.0497 when applied to OpenChat-3.5.

\noindent \textbf{Tighter Risk Control.} 
Beyond statistical validity, we examine how tightly different methods control the system-level risk under the same target risk constraint. 
As presented in Figure~\ref{fig: Test-time FDR on the CommonsenseQA dataset (mean±std).} and Figure~\ref{fig: Test-time FDR on the TriviaQA dataset (mean±std).}, across both datasets and all LLMs, \texttt{LEC} consistently operates near the
target risk level, while UCB-based baselines, including those using exact Clopper-Pearson-style UCB, remain well below it, indicating more conservative behavior. 
For instance, on TriviaQA utilizing the Qwen2.5-3B model, \texttt{LEC} achieves an empirical accepted error rate of 0.0987, whereas \texttt{UCB-CLP} attains 0.0878, and \texttt{UCB-HFD}
fails to identify feasible thresholds due to overly conservative UCB. 

This tighter control allows \texttt{LEC} to retain substantially more samples
without compromising the user-specified risk constraint. 
As illustrated in Table~\ref{tab: power comparison (TriviaQA) similarity-0.6 seb-entailment}, this difference in tightness is further reflected in the power of each method. 
Across all evaluated LLMs and risk levels, \texttt{LEC} consistently achieves higher power than UCB-based baselines, indicating that it admits more valid predictions under the same statistical constraint. 
In contrast, the conservative nature of UCB-based methods, particularly those based on Hoeffding's inequality, often leads to substantially reduced power, or even the absence of feasible thresholds at low risk levels. 
For example, at a risk level of 0.05 on TriviaQA with the Qwen2.5-14B model, \texttt{LEC} achieves a power of 0.7193, retaining $9.5\%$ more admissible samples than \texttt{UCB-CLP}, while \texttt{UCB-HFD} fails to yield any feasible threshold at this level.

\begin{figure*}[!t]
  \centering
  \begin{subfigure}[b]{0.45\textwidth}
    \centering
    \includegraphics[width=\textwidth]{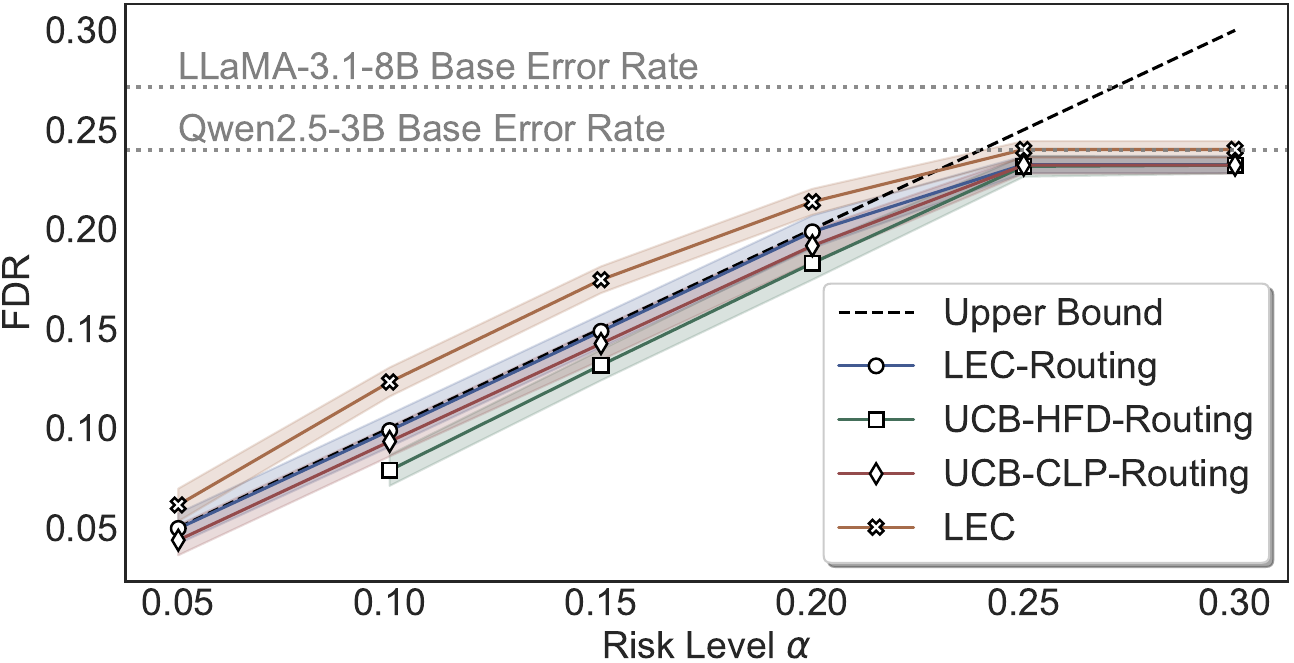}
    \caption{Qwen2.5-3B $\&$ LLaMA-3.1-8B.}
  \end{subfigure}
  \hspace{4mm}
  \begin{subfigure}[b]{0.45\textwidth}
    \centering
    \includegraphics[width=\textwidth]{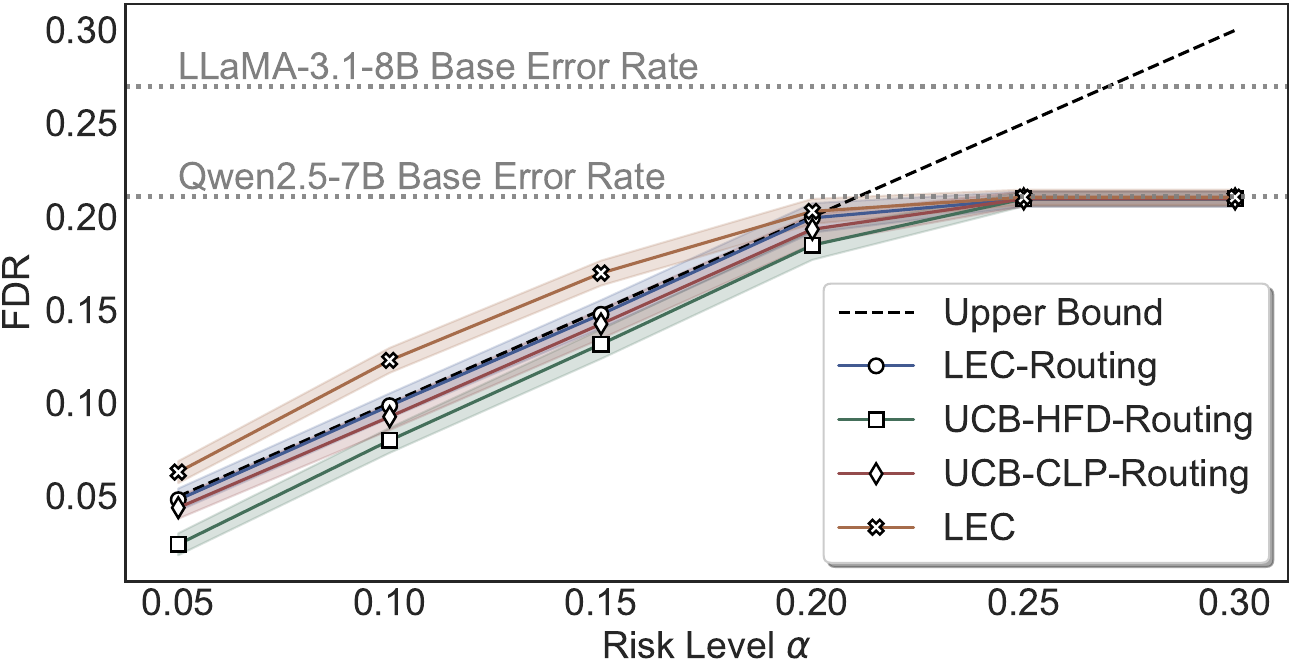}
    \caption{Qwen2.5-7B $\&$ LLaMA-3.1-8B.}
  \end{subfigure}
      \caption{Test-time empirical system-level selection-conditioned error rate of two-model routing on CommonsenseQA (mean$\pm$std). }
  \label{fig: routing FDR control commonsenseQA.}
\end{figure*}

\begin{figure*}[!t]
  \centering
  \begin{subfigure}[b]{0.3\textwidth}
    \centering
    \includegraphics[width=\textwidth]{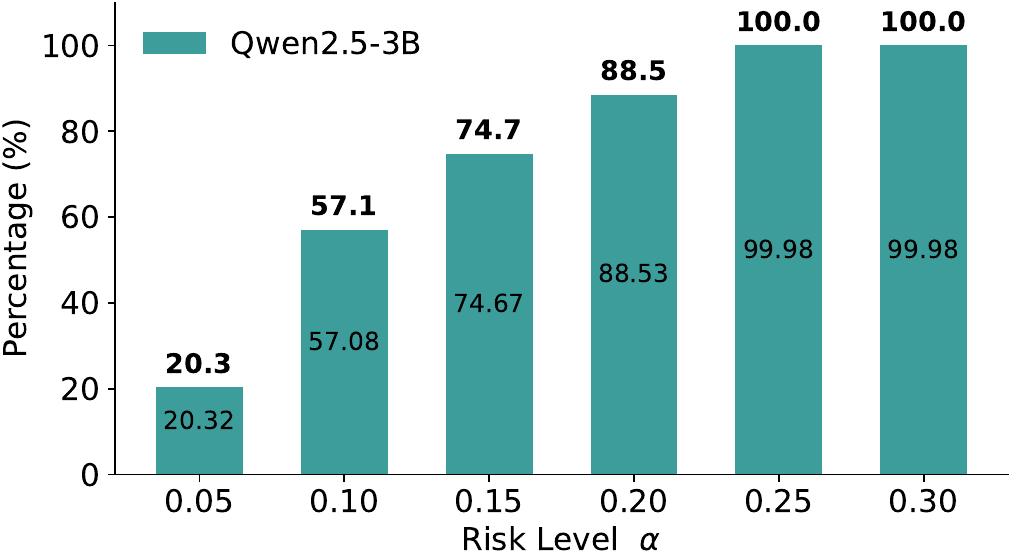}
    \caption{Qwen2.5-3B.}
  \end{subfigure}
  \hspace{1.5mm}
  \begin{subfigure}[b]{0.3\textwidth}
    \centering
    \includegraphics[width=\textwidth]{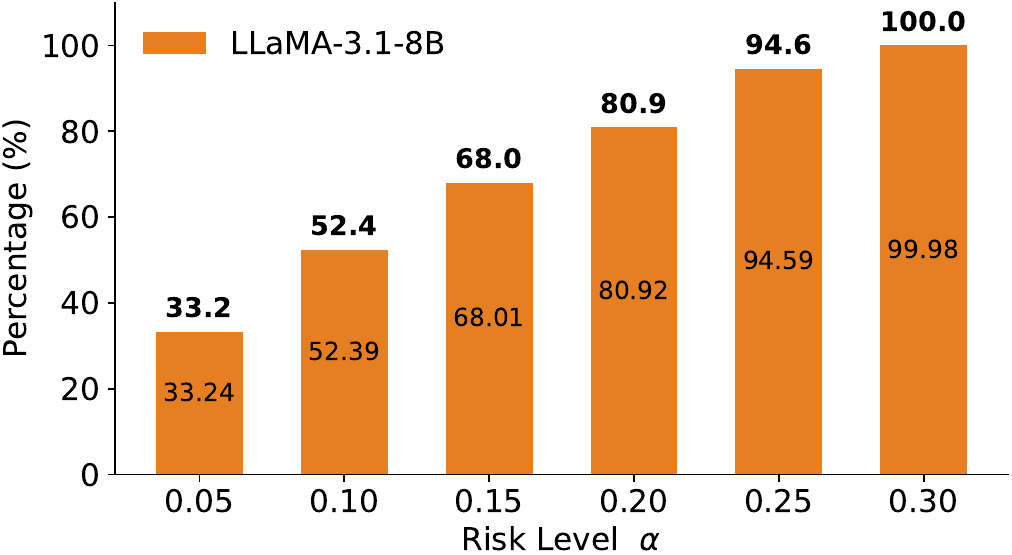}
    \caption{LLaMA-3.1-8B.}
  \end{subfigure}
  \hspace{1.5mm}
  \begin{subfigure}[b]{0.3\textwidth}
    \centering
    \includegraphics[width=\textwidth]{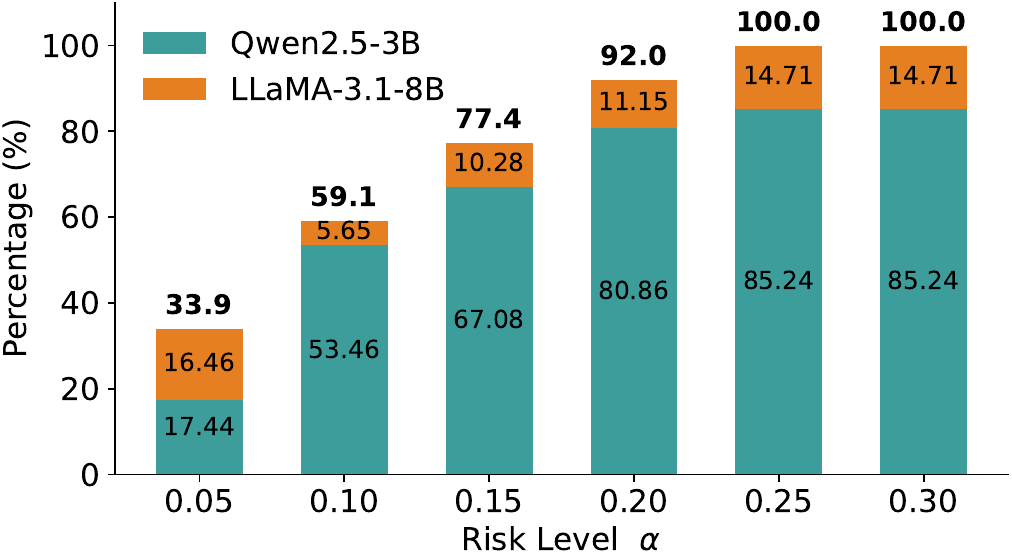}
    \caption{Qwen2.5-3B $\&$ LLaMA-3.1-8B.}
  \end{subfigure}

  \begin{subfigure}[b]{0.3\textwidth}
    \centering
    \includegraphics[width=\textwidth]{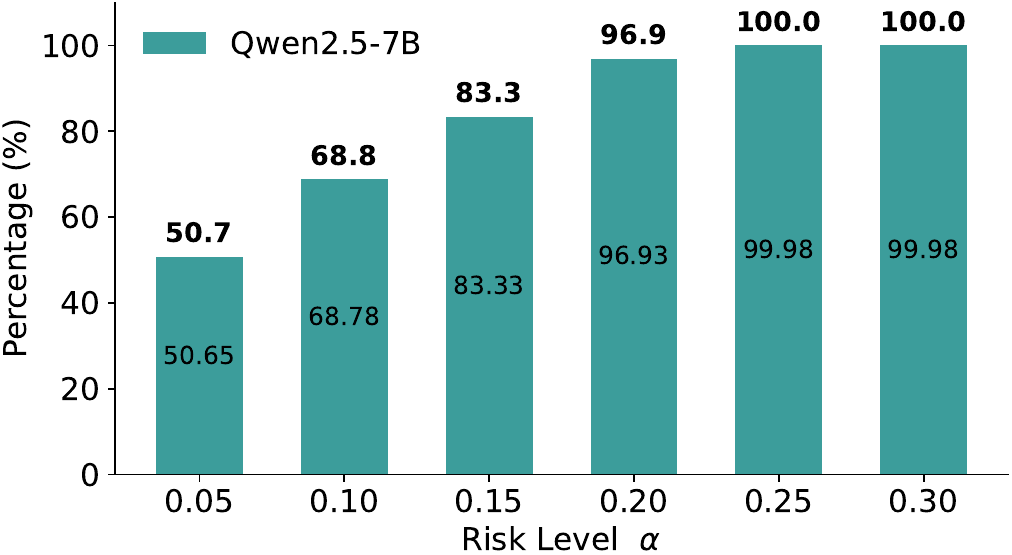}
    \caption{Qwen2.5-7B.}
  \end{subfigure}
  \hspace{1.5mm}
  \begin{subfigure}[b]{0.3\textwidth}
    \centering
    \includegraphics[width=\textwidth]{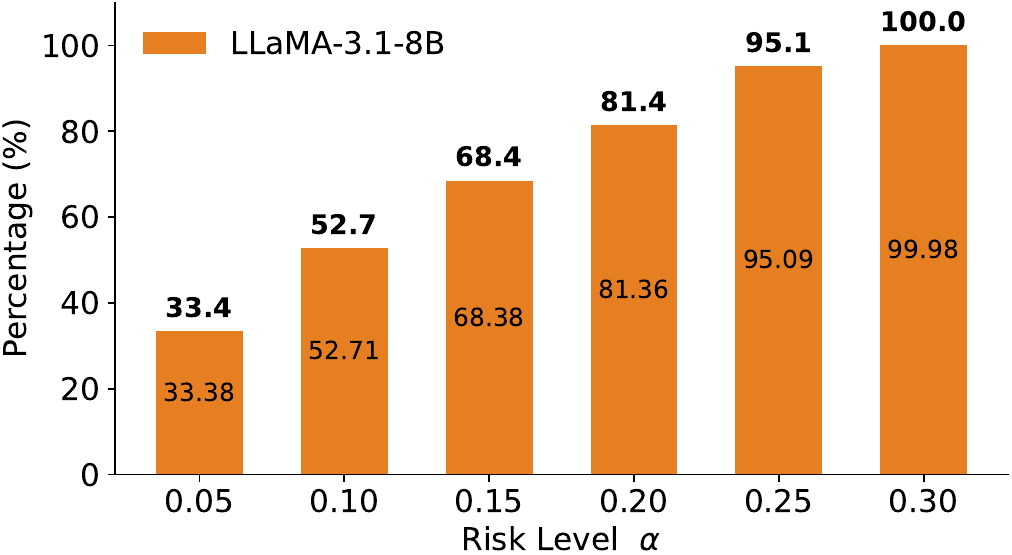}
    \caption{LLaMA-3.1-8B.}
  \end{subfigure}
  \hspace{1.5mm}
  \begin{subfigure}[b]{0.3\textwidth}
    \centering
    \includegraphics[width=\textwidth]{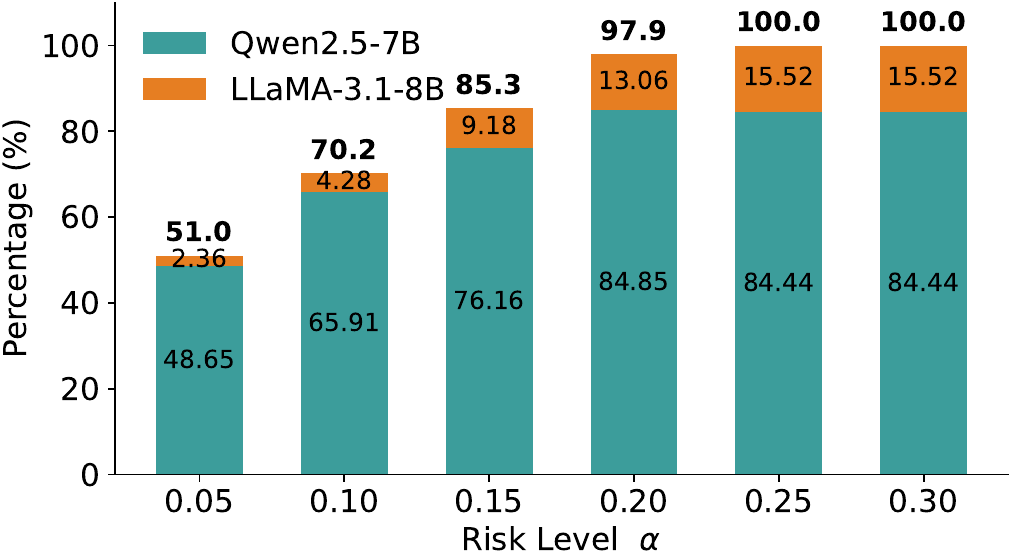}
    \caption{Qwen2.5-7B $\&$ LLaMA-3.1-8B.}
  \end{subfigure}
      \caption{Allocation ratio of accepted test samples in two-model routing systems on the CommonsenseQA dataset (mean).}
  \label{fig: routing proportion commonsenseQA.}
\end{figure*}

\noindent \textbf{Robustness Across Evaluation Settings.}
We further examine whether the observed advantages of \texttt{LEC} are sensitive to specific evaluation settings. 
As shown in Figures~\ref{fig: Test-time FDR (mean±std) on the TriviaQA dataset with entailment for correctness evaluation.} and~\ref{fig: Test-time Power (mean) on the TriviaQA dataset with entailment for correctness evaluation.}, across five LLMs and all tested risk levels, \texttt{LEC} consistently maintains tight selection-conditioned error control and achieves higher power than UCB-based baselines under the same statistical constraints, with bi-entailment as the alignment criterion in function $A$.

\subsection{Evaluations in Two-Model Routing Systems}
\label{sec: Evaluations in Two-Model Routing Systems}
We denote by \texttt{LEC-Routing} the routing strategy obtained by applying the proposed linear expectation constraint to the two-model routing setting, where joint thresholds are calibrated over the system-level selection and error indicators to ensure system-level selection-conditioned error control across the entire routing pipeline. 

\noindent \textbf{System-Level Selection-Conditioned Risk Control.}
We compare \texttt{LEC-Routing} with UCB-based routing baselines~\citep{jung2025trust}. 
In addition, we consider a naive routing variant that calibrates thresholds for each model independently using \texttt{LEC} at the same target, without joint threshold calibration. 
Figure~\ref{fig: routing FDR control commonsenseQA.} shows that \texttt{LEC-Routing} consistently maintains valid and tight system-level selection-conditioned error control on CommonsenseQA by employing both Qwen2.5-3B and Qwen2.5-7B as primary models and selectively delegating inputs to the LLaMA-3.1-8B model, while \texttt{UCB-HFD-Routing} and \texttt{UCB-CLP-Routing} exhibit more conservative behavior. 
Notably, applying \texttt{LEC} without joint threshold calibration does not achieve valid system-level guarantees, which highlights the necessity of joint threshold calibration for achieving reliable system-level risk control in routing systems.

\noindent \textbf{Routing Allocation.}
We further examine the allocation of accepted test samples under two-model routing.
As shown in Figure~\ref{fig: routing proportion commonsenseQA.}, at different risk levels and model pairs, the distribution of accepted samples adapts to the risk budget and the uncertainty profiles of the models: the primary model handles a substantial portion of the accepted samples when its predictions are sufficiently reliable, while uncertain cases are selectively delegated to the secondary model. 
This adaptive allocation leads to both higher system-level coverage and improved cost-efficiency under the prescribed risk constraint.
For example, at $\alpha = 0.05$, using Qwen2.5-3B alone accepts only $20.3\%$ of the test samples. 
In contrast, under \texttt{LEC-Routing} with Qwen2.5-3B as the primary model and LLaMA-3.1-8B as the secondary model, the system accepts $33.9\%$ of the samples in total, with $17.44\%$ handled by Qwen2.5-3B and an additional $16.46\%$ selectively routed to LLaMA-3.1-8B—representing a $13.6\%$ absolute increase in accepted samples over using the primary model alone.

\input{tables/twoModelRoutingCorrectnessCommonsenseQA}

\begin{figure}[!t]
  \centering
  \begin{subfigure}[b]{0.23\textwidth}
    \centering
    \includegraphics[width=\textwidth]{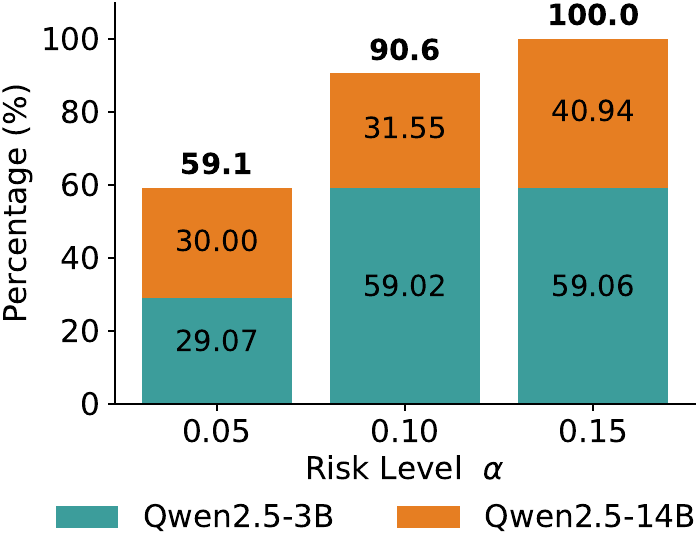}
    \caption{\texttt{LEC-Routing}.}
  \end{subfigure}
  \hfill
  \begin{subfigure}[b]{0.23\textwidth}
    \centering
    \includegraphics[width=\textwidth]{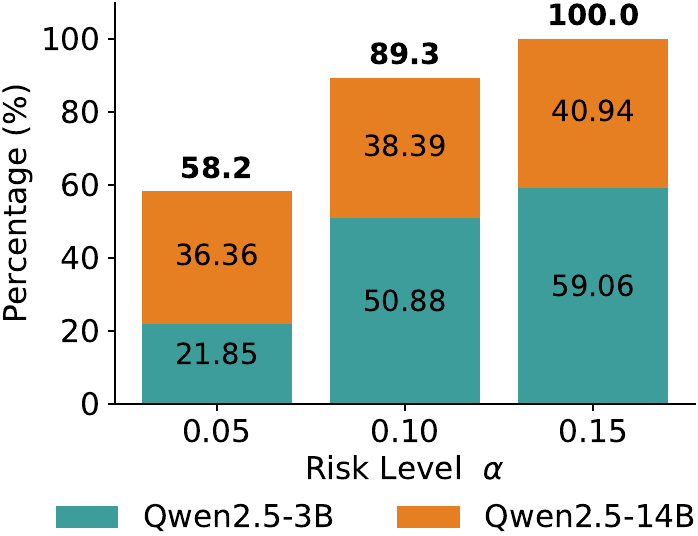}
    \caption{\texttt{UCB-CLP-Routing}.}
  \end{subfigure}
      \caption{Comparison in the allocation ratio of accepted samples in two-model routing systems on the TriviaQA dataset (mean).}
  \label{fig: routing proportion triviaqa comparison lec and ucb-clp.}
\end{figure}

Importantly, routing does not trivially favor the secondary model.
At the same risk level, Qwen2.5-7B alone accepts $50.7\%$ of the samples, while LLaMA-3.1-8B alone accepts only $33.4\%$. 
In this case, \texttt{LEC-Routing} still achieves an acceptance rate of $51.0\%$, with the majority of samples ($48.65\%$) processed by the more efficient primary model Qwen2.5-7B.
This demonstrates that \texttt{LEC-Routing} dynamically balances model usage based on reliability and risk, rather than indiscriminately escalating samples. 
Overall, these results highlight a principled trade-off between efficiency and cost: depending on the risk level and model characteristics, \texttt{LEC-Routing} maximizes system-level utility by invoking the secondary models only when necessary, while preserving rigorous system-level selection-conditioned error guarantees. 

Also, we show that \texttt{LEC-Routing} has the potential to increase the effective set of correct predictions under system-level selection-conditioned error control. 
Table~\ref{tab: routing correctness}
shows that two-model routing under \texttt{LEC-Routing} consistently retains more
correct samples than using either model alone across all risk levels. 
For instance, at a risk level of $\alpha = 0.05$, routing Qwen2.5-3B with
LLaMA-3.1-8B admits 1610 correct samples, compared to 965 and 1579 correct samples
when using Qwen2.5-3B or LLaMA-3.1-8B alone. 


Finally, we illustrate a comparison of the routing behavior between
\texttt{LEC-Routing} and \texttt{UCB-CLP-Routing}. 
As demonstrated in Figure~\ref{fig: routing proportion triviaqa comparison lec and ucb-clp.}, under the same target risk levels, \texttt{LEC-Routing} tends to prioritize the primary model more effectively, while retaining a larger set of accepted samples overall. 
By contrast, \texttt{UCB-CLP-Routing} exhibits a more conservative allocation pattern, with a higher reliance on the secondary model. 
These results further suggest that the tighter system-level risk control of our \texttt{LEC} framework can translate into more efficient routing decisions, although the extent of this advantage may vary across settings. 

Additional experimental results for both single-model and two-model routing systems are reported in Appendix~\ref{sec: Additional Experimental Results}.

%% file: tables/powerTriviaQASimilarity0.6Entailment-Seb.tex
\begin{table}[!t]
\centering
\caption{Power comparison on the TriviaQA dataset (mean).} 
\label{tab: power comparison (TriviaQA) similarity-0.6 seb-entailment}
\adjustbox{max width=\linewidth}{
    \begin{tabular}{ccccccccccc}
        \toprule
        
        \textbf{LLMs} & \textbf{Methods /} $\boldsymbol{\alpha}$  & \textbf{0.05} & \textbf{0.1} & \textbf{0.15} & \textbf{0.2} & \textbf{0.25}\\

        \midrule

        \multirow{3}{*}{OpenChat-3.5} & \texttt{UCB-CLP} & 0.6684 & 0.9294 & 1.0 & 1.0 & 1.0\\
        {} & \texttt{UCB-HFD} & 0.6091 & 0.8884 & 1.0 & 1.0 & 1.0\\
        {} &   \cellcolor{gray!20}  \texttt{LEC} &   \textbf{0.7230} &   \textbf{0.9521} &   \textbf{1.0} &   \textbf{1.0} &   \textbf{1.0}\\
        \midrule
        \multirow{3}{*}{Qwen2.5-3B} & \texttt{UCB-CLP} & 0.2376 & 0.5219 & 0.8554 & 1.0 & 1.0\\
        {} & \texttt{UCB-HFD} & - & 0.3882 & 0.7772 & 1.0 & 1.0\\
        {} &   \cellcolor{gray!20}  \texttt{LEC} &   \textbf{0.2706} &   \textbf{0.5998} &   \textbf{0.9081} &   \textbf{1.0} &   \textbf{1.0}\\
        \midrule
        \multirow{3}{*}{Qwen2.5-7B} & \texttt{UCB-CLP} & 0.3905 & 0.8331 & 1.0 & 1.0 & 1.0\\
        {} & \texttt{UCB-HFD} & - & 0.7396 & 0.9990 & 1.0 & 1.0\\
        {} &   \cellcolor{gray!20}  \texttt{LEC} &   \textbf{0.4889} &   \textbf{0.8850} &   \textbf{1.0} &   \textbf{1.0} &   \textbf{1.0}\\
        \midrule
       \multirow{3}{*}{Qwen2.5-14B} & \texttt{UCB-CLP} & 0.6240 & 0.9987 & 1.0 & 1.0 & 1.0\\
        {} & \texttt{UCB-HFD} & - & 0.9718 & 1.0 & 1.0 & 1.0\\
        {} &   \cellcolor{gray!20}  \texttt{LEC} &   \textbf{0.7193} &   \textbf{1.0} &   \textbf{1.0} &   \textbf{1.0} &   \textbf{1.0}\\
        \midrule
        \multirow{3}{*}{Vicuna-7B-V1.5} & \texttt{UCB-CLP} & - & 0.5686 & 0.8630 & 1.0 & 1.0\\
        {} & \texttt{UCB-HFD} & - & 0.4228 & 0.8068 & 0.9999$_{(5)}$ & 1.0\\
        {} &    \cellcolor{gray!20}  \texttt{LEC} &   - &   \textbf{0.6508} &   \textbf{0.9208} &   \textbf{1.0} &   \textbf{1.0}\\
        \midrule
       \multirow{3}{*}{Vicuna-13B-V1.5} & \texttt{UCB-CLP} & 0.5602 & 0.8944 & 1.0 & 1.0 & 1.0\\
        {} & \texttt{UCB-HFD} & 0.5241 & 0.8364 & 1.0 & 1.0 & 1.0\\
        {} &   \cellcolor{gray!20}  \texttt{LEC} &   \textbf{0.6545} &   \textbf{0.9342} &   \textbf{1.0} &   \textbf{1.0} &   \textbf{1.0}\\
        \midrule
        \multirow{3}{*}{LLaMA-3.1-8B} & \texttt{UCB-CLP} & 0.7143 & 0.9396 & 1.0 & 1.0 & 1.0\\
        {} & \texttt{UCB-HFD} & 0.5339 & 0.9039 & 1.0 & 1.0 & 1.0\\
        {} &   \cellcolor{gray!20}  \texttt{LEC} &   \textbf{0.7538} &   \textbf{0.9612} &   \textbf{1.0} &   \textbf{1.0} &   \textbf{1.0}\\
        \midrule
        \multirow{3}{*}{LLaMA-3.1-70B} & \texttt{UCB-CLP} & 0.9935 & 1.0 & 1.0 & 1.0 & 1.0\\
        {} & \texttt{UCB-HFD} & 0.9503 & 1.0 & 1.0 & 1.0 & 1.0\\
        {} &  \cellcolor{gray!20}  \texttt{LEC} &   \textbf{0.9996} &   \textbf{1.0} &   \textbf{1.0} &   \textbf{1.0} &   \textbf{1.0}\\

        \bottomrule
        
    \end{tabular}
}
\end{table}

%% file: tables/twoModelRoutingCorrectnessCommonsenseQA.tex
\begin{table}[!t]
\centering
\caption{Comparison of the number of accepted correct samples at test time on the CommonsenseQA dataset (mean).}
\label{tab: routing correctness}
\adjustbox{max width=\linewidth}{
    \begin{tabular}{ccccccc}
        \toprule
        \textbf{LLMs /} $\boldsymbol{\alpha}$ & \textbf{0.05} & \textbf{0.1} & \textbf{0.15} & \textbf{0.2} & \textbf{0.25} & \textbf{0.3}\\
        \midrule    
        Qwen2.5-3B & 965 & 2569 & 3174 & 3540 & 3797 & 3797\\
        LLaMA-3.1-8B & 1579 & 2357 & 2890 & 3238 & 3549 & 3643 \\
        \rowcolor{gray!20} Qwen2.5-3B $\&$ LLaMA-3.1-8B & \textbf{1610} & \textbf{2663} & \textbf{3293} & \textbf{3686} & \textbf{3836} & \textbf{3836}\\
        \midrule
        Qwen2.5-7B & 2392 & 3078 & 3523 & 3858 & 3924 & 3924\\
        LLaMA-3.1-8B & 1577 & 2360 & 2892 & 3237 & 3546 & 3629\\
        \rowcolor{gray!20} Qwen2.5-7B $\&$ LLaMA-3.1-8B & \textbf{2413} & \textbf{3144} & \textbf{3615} & \textbf{3896} & \textbf{3928} & \textbf{3928}\\
        \bottomrule
    \end{tabular}
}
\end{table}

%% file: sections/conclusion.tex
\section{Conclusion}
\label{sec: Conclusion}
In this paper, we introduce \texttt{LEC}, a principled formulation that frames selective prediction as a decision problem governed by a linear expectation constraint over selection and error indicators. 
By directly constraining the expected system-level risk, \texttt{LEC} departs from conventional UCB-based approaches that rely on worst-case tail bounds, and instead characterizes a tighter feasible region for admissible decisions. 
We demonstrate that our framework applies naturally to single-model selective prediction and two-model routing systems, where joint calibration over system-level indicators is essential for reliable risk control. 
Empirically, \texttt{LEC} consistently keeps the observed accepted-error rate below the prescribed risk level across various closed-ended and open-ended QA and VQA datasets. 
In routing settings, risk-aware calibration enables adaptive allocation of samples across models and can, in favorable regimes, increase the number of accepted correct predictions relative to single-model deployment, while supporting principled trade-offs between coverage, accuracy, and cost. 
Overall, \texttt{LEC} provides a general foundation for system-level risk control in selective prediction and routing. 
Future work may explore tighter characterizations of when routing yields maximal benefits, extend the framework to broader task-specific risk measures, and integrate \texttt{LEC} with more expressive routing architectures to support reliable decision-making in complex, multi-agent systems.

%% file: sections/impactStatement.tex
\section*{Impact Statement}
\texttt{LEC} offers a foundation for integrating foundation models into high-stakes scenarios that require transparent and auditable reliability guarantees. 
Relying solely on held-out calibration data and exchangeability assumptions, the framework remains applicable in black-box settings and across heterogeneous data sources. We believe this work opens avenues for future research on composable risk control, adaptive model coordination, and uncertainty-aware decision-making in increasingly complex agent systems.

%% file: sections/proofs.tex
\section{Proofs}
\label{sec: Proofs}

\subsection{Proof of Theorem~\ref{thm:single-fdr}}
\label{sec: proof of theorem 1}

Let $\hat{\lambda}^{(a)}$ denote the calibrated threshold obtained from the calibration set by Eq.~\eqref{eq:max-crc-single}. 
For the test sample $(x_{n+1},y^*_{n+1})$, let
\[
err_{n+1}^{(a)}
=
\mathbf{1}\!\left\{
A\!\left(y^*_{n+1},\hat{y}_{n+1}^{(a)}\right)=0
\right\},
\]
\[
S_{n+1}^{(a)}(\hat{\lambda}^{(a)})
=
\mathbf{1}\!\left\{
u_{n+1}^{(a)}\le \hat{\lambda}^{(a)}
\right\},
\]
and
\[
Z_{n+1}^{(a)}(\hat{\lambda}^{(a)})
=
S_{n+1}^{(a)}(\hat{\lambda}^{(a)})
\cdot
err_{n+1}^{(a)}.
\]
If $\mathbb{E}\!\left[
S_{n+1}^{(a)}(\hat{\lambda}^{(a)})
\right]=0$, then the calibrated rule accepts a test example with probability zero, and the guarantee is vacuous. 
We therefore consider the case
\[
\mathbb{E}\!\left[
S_{n+1}^{(a)}(\hat{\lambda}^{(a)})
\right]>0.
\]
The selection-conditioned error rate of the calibrated rule can be written as
\begin{equation}
\begin{split}
&\Pr\!\left(
err_{n+1}^{(a)}=1
\mid
u_{n+1}^{(a)}\le \hat{\lambda}^{(a)}
\right)\\
&\quad =
\Pr\!\left(
err_{n+1}^{(a)}=1
\mid
S_{n+1}^{(a)}(\hat{\lambda}^{(a)})=1
\right)\\
&\quad =
\frac{
\Pr\!\left(
err_{n+1}^{(a)}=1,\,
S_{n+1}^{(a)}(\hat{\lambda}^{(a)})=1
\right)
}{
\Pr\!\left(
S_{n+1}^{(a)}(\hat{\lambda}^{(a)})=1
\right)
}\\
&\quad =
\frac{
\mathbb{E}\!\left[
Z_{n+1}^{(a)}(\hat{\lambda}^{(a)})
\right]
}{
\mathbb{E}\!\left[
S_{n+1}^{(a)}(\hat{\lambda}^{(a)})
\right]
}.
\end{split}
\label{eq:single-scer-ratio-proof}
\end{equation}
Thus, it suffices to prove the linear expectation constraint
\begin{equation}
\mathbb{E}\!\left[
Z_{n+1}^{(a)}(\hat{\lambda}^{(a)})
-
\alpha S_{n+1}^{(a)}(\hat{\lambda}^{(a)})
\right]\le 0.
\label{eq:single-linear-proof-goal}
\end{equation}

We now connect the calibration condition to this population-level linear constraint. 
For any candidate threshold $\lambda^{(a)}$, recall that
\[
S_i^{(a)}(\lambda^{(a)})
=
\mathbf{1}\!\left\{
u_i^{(a)}\le \lambda^{(a)}
\right\},
\quad
Z_i^{(a)}(\lambda^{(a)})
=
S_i^{(a)}(\lambda^{(a)})\,err_i^{(a)}.
\]
Therefore,
\[
Z_i^{(a)}(\lambda^{(a)})
-
\alpha S_i^{(a)}(\lambda^{(a)})
=
\begin{cases}
err_i^{(a)}-\alpha,
& \text{if } u_i^{(a)}\le \lambda^{(a)},\\[3pt]
0,
& \text{if } u_i^{(a)}>\lambda^{(a)}.
\end{cases}
\]
Let $u_{(1)}^{(a)}\le \cdots \le u_{(n)}^{(a)}$ be the sorted calibration uncertainty scores and let $err_{(j)}^{(a)}$ denote the corresponding error indicator. 
For any $\lambda^{(a)}$, define
\[
k^{(a)}(\lambda^{(a)})
=
\#\{i:u_i^{(a)}\le \lambda^{(a)}\}.
\]
Then the calibration sum can be rewritten as
\begin{equation}
\sum_{i=1}^{n}
\left(
Z_i^{(a)}(\lambda^{(a)})
-
\alpha S_i^{(a)}(\lambda^{(a)})
\right)
=
\sum_{j=1}^{k^{(a)}(\lambda^{(a)})}
\left(
err_{(j)}^{(a)}-\alpha
\right).
\label{eq:single-calibration-sum-rewrite}
\end{equation}
By the definition of $\hat{\lambda}^{(a)}$ in Eq.~\eqref{eq:max-crc-single}, the selected threshold satisfies
\begin{equation}
\sum_{j=1}^{k^{(a)}(\hat{\lambda}^{(a)})}
\left(
err_{(j)}^{(a)}-\alpha
\right)
\le -1.
\label{eq:single-calibration-feasible-proof}
\end{equation}
Equivalently,
\begin{equation}
\sum_{i=1}^{n}
\left(
Z_i^{(a)}(\hat{\lambda}^{(a)})
-
\alpha S_i^{(a)}(\hat{\lambda}^{(a)})
\right)
\le -1.
\label{eq:single-linear-calibration-feasible-proof}
\end{equation}

Following the exchangeability-based leave-one-out calibration argument used in conformal risk control~\citep{angelopoulos2024conformal}, the calibration examples and the test example can be treated symmetrically after applying the finite-sample correction. 
Since the calibration and test examples are exchangeable, Eq.~\eqref{eq:single-linear-calibration-feasible-proof}, together with the ``$+1$'' correction, implies
\begin{equation}
\begin{split}
&\mathbb{E}\!\left[
Z_{n+1}^{(a)}(\hat{\lambda}^{(a)})
-
\alpha S_{n+1}^{(a)}(\hat{\lambda}^{(a)})
\right]\\
&\quad =
\frac{1}{n+1}
\mathbb{E}\!\left[
\sum_{i=1}^{n}
\left(
Z_i^{(a)}(\hat{\lambda}^{(a)})
-
\alpha S_i^{(a)}(\hat{\lambda}^{(a)})
\right)
+
\left(
Z_{n+1}^{(a)}(\hat{\lambda}^{(a)})
-
\alpha S_{n+1}^{(a)}(\hat{\lambda}^{(a)})
\right)
\right].
\end{split}
\label{eq:single-exchangeability-proof}
\end{equation}
Substituting Eq.~\eqref{eq:single-linear-calibration-feasible-proof} into Eq.~\eqref{eq:single-exchangeability-proof} gives
\begin{equation}
\begin{split}
&\mathbb{E}\!\left[
Z_{n+1}^{(a)}(\hat{\lambda}^{(a)})
-
\alpha S_{n+1}^{(a)}(\hat{\lambda}^{(a)})
\right]\\
&\quad \le
\frac{1}{n+1}
\mathbb{E}\!\left[
-1+
Z_{n+1}^{(a)}(\hat{\lambda}^{(a)})
-
\alpha S_{n+1}^{(a)}(\hat{\lambda}^{(a)})
\right].
\end{split}
\end{equation}
Since
\[
Z_{n+1}^{(a)}(\hat{\lambda}^{(a)})
\le
S_{n+1}^{(a)}(\hat{\lambda}^{(a)})
\]
and $S_{n+1}^{(a)}(\hat{\lambda}^{(a)})\in\{0,1\}$, we have
\[
Z_{n+1}^{(a)}(\hat{\lambda}^{(a)})
-
\alpha S_{n+1}^{(a)}(\hat{\lambda}^{(a)})
\le 1.
\]
Therefore,
\begin{equation}
\mathbb{E}\!\left[
Z_{n+1}^{(a)}(\hat{\lambda}^{(a)})
-
\alpha S_{n+1}^{(a)}(\hat{\lambda}^{(a)})
\right]
\le 0.
\end{equation}
This proves Eq.~\eqref{eq:single-linear-proof-goal}. 
Rearranging gives
\[
\mathbb{E}\!\left[
Z_{n+1}^{(a)}(\hat{\lambda}^{(a)})
\right]
\le
\alpha\,
\mathbb{E}\!\left[
S_{n+1}^{(a)}(\hat{\lambda}^{(a)})
\right].
\]
Combining this inequality with Eq.~\eqref{eq:single-scer-ratio-proof}, we obtain
\[
\Pr\!\left(
err_{n+1}^{(a)}=1
\mid
u_{n+1}^{(a)}\le \hat{\lambda}^{(a)}
\right)
\le \alpha.
\]
This establishes the claimed single-model selection-conditioned error control.
\qed

\subsection{Proof of Theorem~\ref{thm:routing-fdr}}
\label{sec: proof of theorem 2}

Let $\hat{\boldsymbol{\lambda}}
=
(\hat{\lambda}^{(a)},\hat{\lambda}^{(b)})$ denote the calibrated threshold pair obtained from the calibration set by solving Eq.~\eqref{eq:max-routing}. 
For the test sample $(x_{n+1},y^*_{n+1})$, recall the system-level selection indicator
\[
S_{n+1}(\hat{\boldsymbol{\lambda}})\in\{0,1\},
\]
and the system-level accepted-error indicator
\[
err_{n+1}\in\{0,1\}.
\]
We define the corresponding joint indicator as
\[
Z_{n+1}(\hat{\boldsymbol{\lambda}})
=
S_{n+1}(\hat{\boldsymbol{\lambda}})\cdot err_{n+1}.
\]
If $\mathbb{E}\!\left[
S_{n+1}(\hat{\boldsymbol{\lambda}})
\right]=0$, then the calibrated routing rule accepts a test example with probability zero, and the guarantee is vacuous. 
We therefore consider the case
\[
\mathbb{E}\!\left[
S_{n+1}(\hat{\boldsymbol{\lambda}})
\right]>0.
\]
The system-level selection-conditioned error rate can be written as
\begin{equation}
\begin{split}
&\Pr\!\left(
err_{n+1}=1
\mid
S_{n+1}(\hat{\boldsymbol{\lambda}})=1
\right)\\
&\quad =
\frac{
\Pr\!\left(
err_{n+1}=1,\,
S_{n+1}(\hat{\boldsymbol{\lambda}})=1
\right)
}{
\Pr\!\left(
S_{n+1}(\hat{\boldsymbol{\lambda}})=1
\right)
}\\
&\quad =
\frac{
\mathbb{E}\!\left[
Z_{n+1}(\hat{\boldsymbol{\lambda}})
\right]
}{
\mathbb{E}\!\left[
S_{n+1}(\hat{\boldsymbol{\lambda}})
\right]
}.
\end{split}
\label{eq:routing-scer-ratio}
\end{equation}
Thus, it suffices to show
\begin{equation}
\mathbb{E}\!\left[
Z_{n+1}(\hat{\boldsymbol{\lambda}})
-
\alpha S_{n+1}(\hat{\boldsymbol{\lambda}})
\right]\le 0.
\label{eq:routing-linear-proof-goal}
\end{equation}

By exchangeability among the calibration and test examples at the level of joint model outputs, $(u_i^{(a)},u_i^{(b)},err_i^{(a)},err_i^{(b)})$, and because the routing policy is deterministic and selects at most one model per input, the induced system-level pairs $\left(S_i(\hat{\boldsymbol{\lambda}}),Z_i(\hat{\boldsymbol{\lambda}})\right)$ are exchangeable across examples. 
Following the same exchangeability-based leave-one-out calibration argument used in conformal risk control~\citep{angelopoulos2024conformal}, we have
\begin{equation}
\begin{split}
&\mathbb{E}\!\left[
Z_{n+1}(\hat{\boldsymbol{\lambda}})
-
\alpha S_{n+1}(\hat{\boldsymbol{\lambda}})
\right]\\
&\quad =
\frac{1}{n+1}
\mathbb{E}\!\left[
\sum_{i=1}^{n}
\left(
Z_i(\hat{\boldsymbol{\lambda}})
-
\alpha S_i(\hat{\boldsymbol{\lambda}})
\right)
+
\left(
Z_{n+1}(\hat{\boldsymbol{\lambda}})
-
\alpha S_{n+1}(\hat{\boldsymbol{\lambda}})
\right)
\right].
\end{split}
\label{eq:routing-exchangeability-proof}
\end{equation}
By the definition of the feasible region $\Lambda^{(a,b)}_\alpha$ in Eq.~\eqref{eq:routing feasible set}, the calibrated pair $\hat{\boldsymbol{\lambda}}$ satisfies
\begin{equation}
\sum_{i=1}^{n}
\left(
Z_i(\hat{\boldsymbol{\lambda}})
-
\alpha S_i(\hat{\boldsymbol{\lambda}})
\right)
\le -1.
\label{eq:routing-calibration-feasible-proof}
\end{equation}
Substituting Eq.~\eqref{eq:routing-calibration-feasible-proof} into Eq.~\eqref{eq:routing-exchangeability-proof} gives
\begin{equation}
\begin{split}
&\mathbb{E}\!\left[
Z_{n+1}(\hat{\boldsymbol{\lambda}})
-
\alpha S_{n+1}(\hat{\boldsymbol{\lambda}})
\right]\\
&\quad \le
\frac{1}{n+1}
\mathbb{E}\!\left[
-1+
Z_{n+1}(\hat{\boldsymbol{\lambda}})
-
\alpha S_{n+1}(\hat{\boldsymbol{\lambda}})
\right].
\end{split}
\end{equation}
Since routing selects at most one prediction, 
\[
Z_{n+1}(\hat{\boldsymbol{\lambda}})
\le
S_{n+1}(\hat{\boldsymbol{\lambda}}),
\]
and hence
\[
Z_{n+1}(\hat{\boldsymbol{\lambda}})
-
\alpha S_{n+1}(\hat{\boldsymbol{\lambda}})
\le 1.
\]
Therefore,
\begin{equation}
\mathbb{E}\!\left[
Z_{n+1}(\hat{\boldsymbol{\lambda}})
-
\alpha S_{n+1}(\hat{\boldsymbol{\lambda}})
\right]
\le 0.
\end{equation}
This proves Eq.~\eqref{eq:routing-linear-proof-goal}. 
Rearranging gives
\[
\mathbb{E}\!\left[
Z_{n+1}(\hat{\boldsymbol{\lambda}})
\right]
\le
\alpha\,
\mathbb{E}\!\left[
S_{n+1}(\hat{\boldsymbol{\lambda}})
\right].
\]
Combining this inequality with Eq.~\eqref{eq:routing-scer-ratio}, we obtain
\[
\Pr\!\left(
err_{n+1}=1
\mid
S_{n+1}(\hat{\boldsymbol{\lambda}})=1
\right)
\le \alpha.
\]
Therefore, the two-model routing system satisfies system-level selection-conditioned error control at level $\alpha$.
\qed

\section{Extension to General Multi-Model Routing Systems}
\label{sec: Extension to General Multi-Model Routing Systems}

Suppose we have a collection of $M$ foundation models $\{\mathcal{G}^{(1)},\dots,\mathcal{G}^{(M)}\}$, where each model $\mathcal{G}^{(m)}$ is equipped with an uncertainty score $u^{(m)}$ and a threshold $\lambda^{(m)}$. 
Let $\boldsymbol{\lambda}
=
(\lambda^{(1)},\dots,\lambda^{(M)})$ denote the threshold vector. 
A deterministic routing policy maps the uncertainty scores and thresholds to either a unique accepted model index $r_{\boldsymbol{\lambda}}(x)\in\{1,\dots,M\}$ or abstention. 
For example, in a cascaded system, the policy may select the first model whose uncertainty score does not exceed its threshold; if no model satisfies its threshold, the system abstains.

For any fixed threshold vector $\boldsymbol{\lambda}$, the induced system-level selection indicator is
\[
S_i(\boldsymbol{\lambda})
=
\mathbf{1}\!\left\{
\text{sample } i \text{ is accepted by one model under } \boldsymbol{\lambda}
\right\}.
\]
If $S_i(\boldsymbol{\lambda})=1$, the accepted prediction is $\hat{y}_i
=
\mathcal{G}^{(r_{\boldsymbol{\lambda}}(x_i))}(x_i)$. 
The corresponding accepted-error indicator is
\[
err_i(\boldsymbol{\lambda})
=
\mathbf{1}\!\left\{
S_i(\boldsymbol{\lambda})=1
\ \land\
A(y_i^*,\hat{y}_i)=0
\right\}.
\]
We then define
\[
Z_i(\boldsymbol{\lambda})
=
S_i(\boldsymbol{\lambda})\cdot err_i(\boldsymbol{\lambda})
=
err_i(\boldsymbol{\lambda}),
\]
where the last equality holds because $err_i(\boldsymbol{\lambda})$ is already defined as an accepted-error indicator.

The system-level selection-conditioned error rate is
\[
\mathrm{SCER}(\boldsymbol{\lambda})
=
\frac{
\mathbb{E}[Z(\boldsymbol{\lambda})]
}{
\mathbb{E}[S(\boldsymbol{\lambda})]
},
\]
whenever $\mathbb{E}[S(\boldsymbol{\lambda})]>0$. 
Thus, controlling $\mathrm{SCER}(\boldsymbol{\lambda})\le \alpha$ is equivalent to the linear expectation constraint
\[
\mathbb{E}\!\left[
Z(\boldsymbol{\lambda})
-
\alpha S(\boldsymbol{\lambda})
\right]\le 0.
\]
Following the same finite-sample calibration argument as in the single-model and two-model cases, a sufficient empirical condition is
\begin{equation}
\sum_{i=1}^{n}
\left(
Z_i(\boldsymbol{\lambda})
-
\alpha S_i(\boldsymbol{\lambda})
\right)
\le -1.
\label{eq:multi-model-empirical-condition}
\end{equation}
Equivalently, since $Z_i(\boldsymbol{\lambda})=S_i(\boldsymbol{\lambda})\cdot err_i(\boldsymbol{\lambda})$, Eq.~\eqref{eq:multi-model-empirical-condition} can be written as
\[
\sum_{i=1}^{n}
\left(
S_i(\boldsymbol{\lambda})\,err_i(\boldsymbol{\lambda})
-
\alpha S_i(\boldsymbol{\lambda})
\right)
\le -1.
\]
We define the feasible threshold region as
\begin{equation}
\Lambda^{(1:M)}_\alpha
=
\left\{
\boldsymbol{\lambda}:
\sum_{i=1}^{n}
\left(
Z_i(\boldsymbol{\lambda})
-
\alpha S_i(\boldsymbol{\lambda})
\right)
\le -1
\right\}.
\label{eq:multi-model-feasible-region}
\end{equation}
Among all feasible threshold vectors, we select the retention-maximizing solution
\begin{equation}
\hat{\boldsymbol{\lambda}}
=
\operatorname*{argmax}_{\boldsymbol{\lambda}\in\Lambda^{(1:M)}_\alpha}
\frac{1}{n}
\sum_{i=1}^{n}
S_i(\boldsymbol{\lambda}).
\label{eq:multi-model-argmax}
\end{equation}
By the same exchangeability-based calibration argument as above, the resulting routing system satisfies
\[
\Pr\!\left(
err_{n+1}(\hat{\boldsymbol{\lambda}})=1
\mid
S_{n+1}(\hat{\boldsymbol{\lambda}})=1
\right)
\le \alpha,
\]
provided that the routing policy deterministically maps each input to at most one accepted model output or to abstention.

Therefore, \texttt{LEC} extends naturally to routing systems with an arbitrary number of models. 
The main algorithmic challenge is computational: the threshold vector $\boldsymbol{\lambda}$ is multi-dimensional, and finding the retention-maximizing feasible vector may require an efficient structured search over the threshold space. 
Nevertheless, this extension does not alter the underlying statistical form of the guarantee, because the system still induces binary selection and accepted-error indicators. 
More broadly, the same linear expectation transformation can apply to other task-specific ratio-form risk metrics, as long as the numerator and denominator can be represented through suitable system-level indicators.

%% file: sections/additionalExperimentalSettings.tex
\section{Additional Experimental Settings}
\label{sec: Additional Experimental Settings}

\definecolor{lightgraybg}{RGB}{254,250,246}

\noindent \textbf{Details of Utilized Datasets and Models.} 
For the closed-ended CommonsenseQA dataset, we employ both the full training split (9,741 samples) and the validation split (1,221 samples)\footnote{\href{https://huggingface.co/datasets/tau/commonsense_qa/tree/main/data}{Source files of the CommonsenseQA dataset.}}. 
We remove a small number of samples containing non-ASCII characters in either the query or answer that cannot be encoded by the tokenizer. 
From the remaining data, we select one QA pair as a fixed one-shot demonstration, which is prepended to the prompt for all other samples. 
After filtering and prompt construction, we select 10,000 QA instances in total. 
An example of the complete prompt is presented as follows: 

\begin{tcolorbox}[
    colback=lightgraybg
]
\small{\texttt{$\#\#\#$ System:}\\
\texttt{Make your best effort and select the correct answer for the following multiple-choice question. 
For each question, only one choice is correct. 
Answer should be one among A, B, C, D, E.}
\\

\texttt{$\#\#\#$ User:}\\
\texttt{What is something I need to avoid while playing ball?}\\
\texttt{A: competition}\\
\texttt{B: losing}\\
\texttt{C: injury}\\
\texttt{D: hitting the ball}\\
\texttt{E: having fun}\\
\texttt{$\#\#\#$ Assistant:}\\
\texttt{C}\\

\texttt{$\#\#\#$ User:}\\
\texttt{The sanctions against the school were a punishing blow, and they seemed to what the efforts the school had made to change?}\\
\texttt{A: ignore}\\
\texttt{B: enforce}\\
\texttt{C: authoritarian}\\
\texttt{D: yell at}\\
\texttt{E: avoid}\\
\texttt{$\#\#\#$ Assistant:}}
\end{tcolorbox}

For the open-ended TriviaQA dataset, we randomly select 8,000 QA pairs from the validation split of the \texttt{rc.nocontext} subset\footnote{\href{https://huggingface.co/datasets/mandarjoshi/trivia_qa/tree/main}{Source files of the TriviaQA dataset.}}. 
We also apply a one-shot prompt for each data point. 
An example of the complete prompt is presented as follows: 

\begin{tcolorbox}[
    colback=lightgraybg
]
\small{\texttt{$\#\#\#$ System:}\\
\texttt{This is a bot that correctly answers questions.}
\\

\texttt{$\#\#\#$ User:}\\
\texttt{In 1968, who did radical feminist Valerie Solanas shoot and wound as he entered his New York studio?}\\
\texttt{$\#\#\#$ Assistant:}\\
\texttt{Andy Warhol}\\

\texttt{$\#\#\#$ User:}\\
\texttt{Who was the man behind The Chipmunks?}\\
\texttt{$\#\#\#$ Assistant:}}
\end{tcolorbox}

For the open-ended MM-Vet v2 dataset, we adopt the total test split (517 VQA samples)
\footnote{\href{https://huggingface.co/datasets/whyu/mm-vet-v2/tree/main/data}{Source files of the MM-Vet v2 dataset.}} for evaluation. 
An example of the complete prompt is presented as follows: 
\begin{tcolorbox}[
    colback=lightgraybg
]
\small{\texttt{<image>}\\
\texttt{What is x in the equation?}\\
\texttt{NOTE: Provide only the final answer. Do not provide unrelated details.}}
\end{tcolorbox}

For the closed-ended ScienceQA dataset, we utilize the test split (4,241 samples)\footnote{\href{https://huggingface.co/datasets/derek-thomas/ScienceQA/tree/main/data}{Source files of the ScienceQA dataset.}}. 
Due to missing visual inputs in a subset of the data, we retain 2,017 VQA samples for evaluation. 
An example of the complete prompt is presented as follows: 

\begin{tcolorbox}[
    colback=lightgraybg
]
\small{\texttt{<image>}\\
\texttt{Which of the following could Gordon's test show?}\\
\texttt{A: if the spacecraft was damaged when using a parachute with a 1 m vent going 200 km per hour}
\\
\texttt{B: how steady a parachute with a 1 m vent was at 200 km per hour}\\
\texttt{C: whether a parachute with a 1 m vent would swing too much at 400 km per hour}\\

\texttt{This is a single choice question, answer only with choice number in A, B, C.}}
\end{tcolorbox}

In QA tasks, we employ four series of open-source LLMs available on Hugging Face: OpenChat, LLaMA, Vicuna, and Qwen, divided by the model size into: (1) 3B: Qwen-2.5-3B-Instruct; (2) 7B: OpenChat-3.5, Vicuna-7B-v1.5, and Qwen-2.5-7B-Instruct; (3) 8B: LLaMA-3.1-8B-Instruct; (4) 13B: Vicuna-13B-v1.5; (5) 14B: Qwen-2.5-14B-Instruct; (6) 70B: LLaMA-3.1-70B-Instruct. 
In VQA tasks, we employ three distinct LVLM groups: LLaVA1.5, LLaVA-NeXT, and InternVL2,  divided by the model size into: (1) 1B: InternVL2-1B; (2) 7B: LLaVA-1.5-7B-HF and LLaVA-V1.6-Mistral-7B-HF; (3) 8B: InternVL2-8B. 
We omit ``-Instruct'' and ``-HF'' when reporting the experimental results. 

Due to imperfect instruction-following behavior in some models, a very small number of predictions may be invalid and thus removed during preprocessing. 
Specifically, in closed-ended tasks, a few model outputs do not strictly follow the prescribed option format, while in open-ended tasks, rare cases may result in empty responses after standard cleaning and post-processing. 
Consequently, the set of valid evaluation samples can differ slightly across models, although the number of excluded samples is negligible. 
In the two-model routing setting, we therefore restrict evaluation to the subset of samples that are shared by both models to ensure a fair and consistent comparison. 
As shown in Table~\ref{tab: routing correctness}, the set of evaluation samples shared between LLaMA-3.1-8B and the two Qwen primary models exhibits a minor discrepancy, stemming from a very small number of invalid predictions that are removed during preprocessing. 
This difference is negligible in scale and does not materially affect the evaluation or the conclusions regarding the effectiveness of the proposed method.

\noindent \textbf{Details of Alignment Criteria.} 
In closed-ended QA or VQA tasks, we can directly determine whether the predicted option is consistent with the ground-truth option. 
In open-ended settings, following previous work~\citep{duan2024shifting,wang2025coin}, we estimate the sentence similarity between two answers (ground truth, the most likely answer, or sampled answer) leveraging SentenceTransformers~\citep{reimers-gurevych-2019-sentence} with DistillRoBERTa~\citep{sanh2019distilbert} as the backbone. 
For bi-entailment~\citep{kuhn2023semantic,farquhar2024detecting,wang2025word}, we employ DeBERTa-v3\footnote{We use \href{https://huggingface.co/MoritzLaurer/DeBERTa-v3-large-mnli-fever-anli-ling-wanli}{DeBERTa-v3-large-mnli-fever-anli-ling-wanli}. Different from DeBERTa-large~\citep{he2021deberta} employed in previous research~\citep{lin2024generating}, the three dimensions of its output logits correspond to entailment, neutral, and contradiction, respectively.} as the Natural Language Inference (NLI) classifier, which outputs logits over three semantic relation classes: entailment, neutral, and contradiction.
Two answers are deemed semantically aligned if the classifier predicts entailment for both directions. 
In addition, we also adopt LLM-as-a-Judge by prompting the Qwen2.5-7B model with the following instruction: 
\begin{tcolorbox}[
    colback=lightgraybg
]
\small{\texttt{You are an expert evaluator for open-ended QA correctness.}\\

\texttt{Given a question, a ground-truth answer, and a model's answer, decide which option best describes the model's answer:}\\
\texttt{A. correct   – semantically equivalent to the ground-truth answer.}\\
\texttt{B. partial   – related and contains some correct information but is incomplete or partially wrong.}\\
\texttt{C. incorrect – not compatible with the ground-truth answer.}\\

\texttt{Respond by selecting exactly one of A, B, or C.}\\

\texttt{Question: <TEXT>}\\
\texttt{Ground truth answer: <TEXT>}\\
\texttt{Model answer: <TEXT>}\\

\texttt{Answer:}}
\end{tcolorbox}
We adopt the partial correctness criterion by default. 
Note that the statistical validity of our \texttt{LEC} framework is not affected by changes in the alignment criterion of admission function $A$. 

\noindent \textbf{Details of Uncertainty Estimators.} 
In the closed-ended CommonsenseQA dataset, we compute the PE as $\sum_o - p_o \log p_o$, where $p_o$ is the probability of the $o$-th option. 
In black-box settings, we sample additional 20 answers (i.e., options) and utilize the normalized frequency score as $p_o$. 
We only compute black-box PE in the closed-ended ScienceQA (VQA) dataset. 
In the open-ended TriviaQA (QA) and MM-Vet v2 (VQA) datasets, we sample additional 10 answers by default to compute black-box SE, EigV, Ecc, and Deg. 
In black-box SE, we perform semantic clustering via the bi-entailment criterion. 
See \citet{farquhar2024detecting} for more details of black-box SE. 
See \citet{lin2024generating} for more details of EigV, Ecc, and Deg. 
For SELF, we compute the length-normalized sentence entropy of the most likely generation. 
See \citet{duan2024shifting} for details. 

\noindent \textbf{Details of Additional Hyperparameters.} 
In black-box settings, we set the sampling temperature to 1.0 and top-p to 0.9. 
For both the CommonsenseQA and ScienceQA datasets, limit the model’s output to a single token, since only the option letter is required. For the TriviaQA dataset, we set the maximum output length to 36 tokens. 
For the MM-Vet v2 dataset, we set the maximum output length to 32 tokens. 
In UCB-based calibration, we set the significance level $\delta$ to 0.05. 

\noindent \textbf{Details of Baselines.} 
As presented in Algorithms~\ref{alg: algorithm of UCB-Based PAC-style FDR Control} and~\ref{alg:lec_single}, we detail the threshold calibration procedures in single-model selective prediction for two UCB-based approaches and the proposed \texttt{LEC}. 
Algorithm~\ref{alg:lec routing} details \texttt{LEC-Routing} that follows naturally by reformulating the selection and joint indicators at the system level, without altering the overall calibration procedure. 
Moreover, in the two-model routing setting, rather than selecting the largest feasible threshold pair, we adopt the threshold pair that yields the maximum number of accepted samples. 
See \citet{jung2025trust} for UCB-based routing.

\begin{algorithm}[!h]
  \caption{UCB-based threshold calibration for single-model selective prediction with PAC-style accepted-error control}
  \label{alg: algorithm of UCB-Based PAC-style FDR Control}
  \begin{algorithmic}[1]
    \STATE {\bfseries Input:} Primary model $\mathcal{G}^{(a)}$, calibration set $\{ ( x_i, y_i^*, \hat{y}_i^{(a)} ) \}_{i=1}^{N}$, uncertainty estimator $\mathcal{U}$, admission function $A$, risk level $\alpha$, significance level $\delta$, confidence interval type (\texttt{HFD} or \texttt{CLP})
    \STATE \textbf{Output:} Calibrated threshold $\hat{\lambda}^{(a)}$

    \STATE Compute uncertainty scores for all $i \in [N]$: $u_i^{(a)} \leftarrow \mathcal{U}(x_i; \mathcal{G}^{(a)})$;
    \STATE Compute error indicators for all $i \in [N]$: $err_i^{(a)} \leftarrow \mathbf{1}\{ A(y_i^*, \hat{y}_i^{(a)}) = 0 \}$;
    \STATE Sort uncertainty scores in ascending order $u_{(1)}^{(a)} \le \cdots \le u_{(N)}^{(a)}$, with corresponding error indicators $err_{(1)}^{(a)}, \ldots, err_{(N)}^{(a)}$;
    \STATE Initialize $\hat{\lambda}^{(a)} \leftarrow $ \texttt{NULL};

    \FOR{$i = 1$ \textbf{to} $N$}
        \STATE Let $\lambda \leftarrow u_{(i)}^{(a)}$;
        \STATE Number of accepted samples: $n_\lambda \leftarrow \sum_{j=1}^{N} \mathbf{1} \{ u_{(j)}^{(a)} \leq \lambda \}$;
        \STATE Number of errors among accepted samples: $X_\lambda \leftarrow \sum_{j=1}^{N} \mathbf{1} \{ u_{(j)}^{(a)} \leq \lambda \land err_{(j)}^{(a)} =1 \}$;

        \IF{confidence interval type is \texttt{HFD}}
            \STATE Compute Hoeffding-style $(1-\delta)$ upper confidence bound: $\mathrm{UCB} \leftarrow \frac{X_\lambda}{n_\lambda} 
            + \sqrt{\frac{\log(1/\delta)}{2 n_\lambda}}$;
        \ELSIF{confidence interval type is \texttt{CLP}}
            \IF{$X_\lambda = n_\lambda$}
                \STATE Set $\mathrm{UCB} \leftarrow 1$;
            \ELSE
                \STATE Compute Clopper-Pearson-style $(1-\delta)$ upper confidence bound: $\mathrm{UCB} \leftarrow \mathrm{BetaInv}\!\left(1-\delta;\; X_\lambda+1,\; n_\lambda-X_\lambda\right)$;
            \ENDIF
        \ENDIF

        \IF{$\mathrm{UCB} \le \alpha$}
            \STATE Update $\hat{\lambda}^{(a)} \leftarrow \lambda$;
        \ENDIF
    \ENDFOR

    \IF{$\hat{\lambda}^{(a)} = $ \texttt{NULL}}
        \STATE \textbf{Return} ``No feasible threshold for risk level $\alpha$''
    \ELSE
        \STATE \textbf{Return} $\hat{\lambda}^{(a)}$
    \ENDIF 
  \end{algorithmic}
\end{algorithm}

\begin{algorithm}[h]
\caption{Threshold calibration via \texttt{LEC} for single-model selective prediction with selection-conditioned error control}
\label{alg:lec_single}
\begin{algorithmic}[1]
\STATE \textbf{Input:} $\mathcal{G}^{(a)}$, $\{ ( x_i, y_i^*, \hat{y}_i^{(a)} ) \}_{i=1}^{N}$, $\mathcal{U}$, $A$, $\alpha$ 
\STATE \textbf{Output:} Calibrated threshold $\hat{\lambda}^{(a)}$

\STATE Compute uncertainty scores and error indicators for all $i\in[N]$: $u_i^{(a)} \leftarrow \mathcal{U}(x_i;\mathcal{G}^{(a)}), err_i^{(a)} \leftarrow \mathbf{1}\{A(y_i^*,\hat{y}_i^{(a)})=0\};$
\STATE Sort uncertainty scores in ascending order $u_{(1)}^{(a)} \le \cdots \le u_{(N)}^{(a)}$, with corresponding error indicators $err_{(1)}^{(a)},\ldots,err_{(N)}^{(a)}$;
\STATE Define candidate set $\mathcal{T}^{(a)} \leftarrow \{u_{(1)}^{(a)}, \ldots, u_{(N)}^{(a)}\}$;
\STATE Initialize $\hat{\lambda}^{(a)} \leftarrow$ \texttt{NULL};

\FOR{each $\lambda \in \mathcal{T}^{(a)}$}
    \STATE Obtain selection indicators for all $i\in[N]$: $S_i^{(a)}(\lambda) \leftarrow \mathbf{1}\{u_i^{(a)} \le \lambda\}$;
    \STATE Obtain joint indicators for all $i\in[N]$: $ Z_i^{(a)}(\lambda) \leftarrow S_i^{(a)}(\lambda)\cdot err_i^{(a)}$;
    \STATE Compute the empirical linear constraint: $L(\lambda) \leftarrow \sum_{i=1}^{N}
    \left(
    Z_i^{(a)}(\lambda)-\alpha S_i^{(a)}(\lambda)
    \right)$;
    \IF{$L(\lambda)\le -1$}
        \STATE Update $\hat{\lambda}^{(a)} \leftarrow \lambda$;
    \ENDIF
\ENDFOR

\IF{$\hat{\lambda}^{(a)}=$ \texttt{NULL}}
    \STATE \textbf{Return} ``No feasible threshold for risk level $\alpha$'';
\ELSE
    \STATE \textbf{Return} $\hat{\lambda}^{(a)}$;
\ENDIF
\end{algorithmic}
\end{algorithm}

\begin{algorithm}[!h]
\caption{Joint threshold calibration via \texttt{LEC-Routing} for two-model routing systems}
\label{alg:lec routing}
\begin{algorithmic}[1]
\STATE \textbf{Input:} Models $(\mathcal{G}^{(a)}, \mathcal{G}^{(b)})$, calibration set $\{ ( u_i^{(a)}, u_i^{(b)}, err_i^{(a)}, err_i^{(b)} ) \}_{i=1}^{N}$, risk level $\alpha$ 
\STATE \textbf{Output:} Calibrated threshold pair $(\hat{\lambda}^{(a)}, \hat{\lambda}^{(b)})$

\STATE Sort $\{u_i^{(a)}\}_{i=1}^{N}$ in ascending order and define candidate set $\mathcal{T}^{(a)} \leftarrow \{u_{(1)}^{(a)}, \ldots, u_{(N)}^{(a)}\}$; 
\STATE Sort $\{u_i^{(b)}\}_{i=1}^{N}$ in ascending order and define candidate set $\mathcal{T}^{(b)} \leftarrow \{u_{(1)}^{(b)}, \ldots, u_{(N)}^{(b)}\}$;
\STATE Initialize feasible set $\Lambda_{\alpha}^{(a,b)} \leftarrow \emptyset$;

\FOR{each $\lambda^{(a)} \in \mathcal{T}^{(a)}$}
    \FOR{each $\lambda^{(b)} \in \mathcal{T}^{(b)}$}
        \STATE Obtain system-level selection indicators: $S_i(\lambda^{(a)},\lambda^{(b)})
        \leftarrow
        \mathbf{1}\{u_i^{(a)} \le \lambda^{(a)}\}
        +
        \mathbf{1}\{u_i^{(a)}>\lambda^{(a)} \land u_i^{(b)} \le \lambda^{(b)}\}$;
        \STATE Obtain system-level joint indicators: $Z_i(\lambda^{(a)},\lambda^{(b)})
        \leftarrow\;
        \mathbf{1}\{u_i^{(a)} \le \lambda^{(a)} \land err_i^{(a)}=1\}
        +
        \mathbf{1}\{u_i^{(a)}>\lambda^{(a)} \land u_i^{(b)} \le \lambda^{(b)} \land err_i^{(b)}=1\}$; 
        \STATE Compute the system-level empirical linear constraint: $L(\lambda^{(a)},\lambda^{(b)})
        \leftarrow
        \sum_{i=1}^{N}
        \left(
        Z_i(\lambda^{(a)},\lambda^{(b)})
        -
        \alpha S_i(\lambda^{(a)},\lambda^{(b)})
        \right)$;
        \IF{$L(\lambda^{(a)},\lambda^{(b)})\le -1$}
            \STATE Add $(\lambda^{(a)},\lambda^{(b)})$ to $\Lambda_{\alpha}^{(a,b)}$;
        \ENDIF
    \ENDFOR
\ENDFOR

\IF{$\Lambda_{\alpha}^{(a,b)}=\emptyset$}
    \STATE \textbf{Return} ``No feasible threshold pair for risk level $\alpha$'';
\ELSE
    \STATE Select the retention-maximizing feasible threshold pair: $(\hat{\lambda}^{(a)},\hat{\lambda}^{(b)})
    \leftarrow
    \operatorname*{argmax}_{(\lambda^{(a)},\lambda^{(b)})\in \Lambda_{\alpha}^{(a,b)}}
    \sum_{i=1}^{N} S_i(\lambda^{(a)},\lambda^{(b)})$;
    \STATE \textbf{Return} $(\hat{\lambda}^{(a)},\hat{\lambda}^{(b)})$.
\ENDIF
\end{algorithmic}
\end{algorithm}

When iterating over candidate thresholds, one can adopt binary search or other accelerated strategies. 
The search space can be adjusted according to the specific UQ method, and finer threshold granularity can be achieved by decreasing the search step size. All such procedures are performed offline during calibration; the test-time deployment remains fully real-time. 
\newpage

%% file: sections/additionResults.tex
\section{Additional Experimental Results}
\label{sec: Additional Experimental Results}
\begin{figure*}[!t]
  \centering
  \begin{subfigure}[b]{0.495\textwidth}
    \centering
    \includegraphics[width=\textwidth]{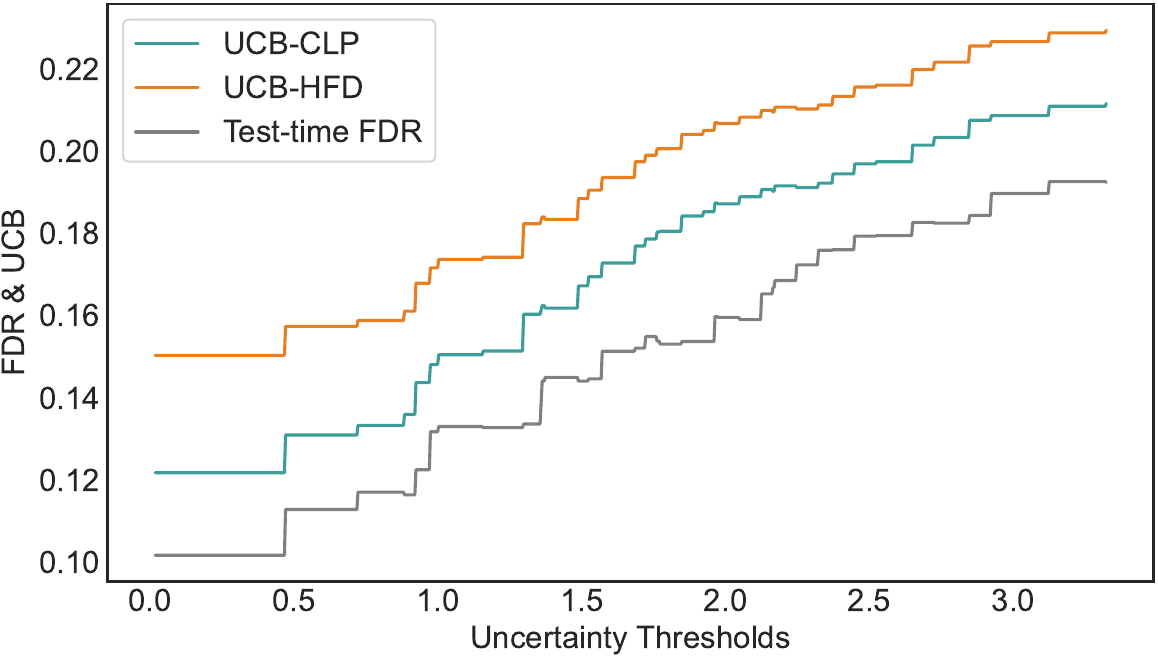}
    \caption{TriviaQA.}\label{fig: Upper confidence bound vs. FDR (a)}
  \end{subfigure}
  \hfill
  \begin{subfigure}[b]{0.495\textwidth}
    \centering
    \includegraphics[width=\textwidth]{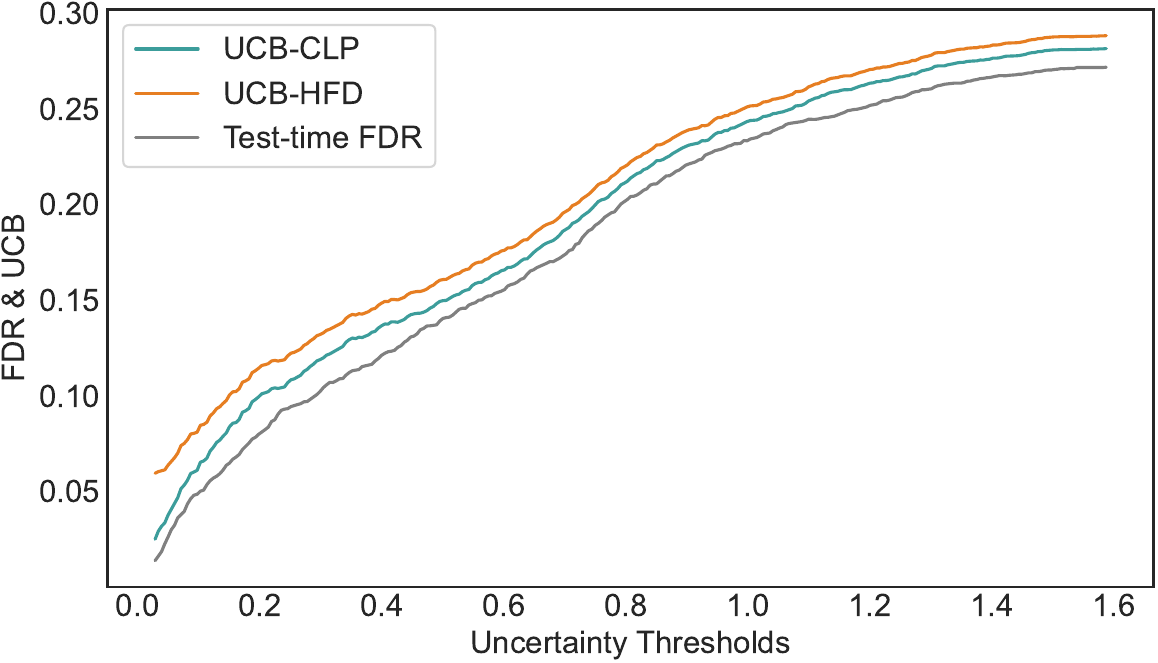}
    \caption{CommonsenseQA.}\label{fig: Upper confidence bound vs. FDR (b)}
  \end{subfigure}
  
    \caption{Two styles of UCBs versus the test-time empirical selection-conditioned error rate at various uncertainty thresholds. 
In (a), we use the LLaMA-3.1-8B model with white-box PE as the uncertainty estimator; in (b), we use Qwen2.5-14B with SE as the uncertainty estimator. 
The y-axis label ``FDR'' denotes the observed fraction of erroneous predictions among accepted predictions.}
  \label{fig: Upper confidence bound vs. FDR}
\end{figure*}

\noindent \textbf{Inherent Conservation of Confidence Intervals.} 
Figure~\ref{fig: Upper confidence bound vs. FDR} illustrates the gap between the test-time selection-conditioned error rate and the UCBs computed on the calibration set across a range of uncertainty thresholds. Specifically, Figure~\ref{fig: Upper confidence bound vs. FDR} (a) reports results on TriviaQA using LLaMA-3.1-8B with white-box PE, while Figure~\ref{fig: Upper confidence bound vs. FDR} (b) shows results on CommonsenseQA using Qwen2.5-14B with SE. 
A key empirical observation is that, across almost all uncertainty thresholds, both Hoeffding-style UCBs and exact Clopper–Pearson UCBs systematically overestimate the corresponding test-time selection-conditioned error rate, indicating that the conservativeness is not merely due to loose concentration inequalities, but rather reflects a more fundamental limitation of UCB-based calibration. 
This behavior can be attributed to the objective of UCB-based methods, which aim to control worst-case tail events with high probability. 
To satisfy this requirement, the UCB must remain valid even under rare but adversarial realizations of the calibration data, such as observing an unusually low empirical error rate despite a relatively high underlying risk. 

\begin{figure*}[!t]
  \centering
  \begin{subfigure}[b]{0.246\textwidth}
    \centering
    \includegraphics[width=\textwidth]{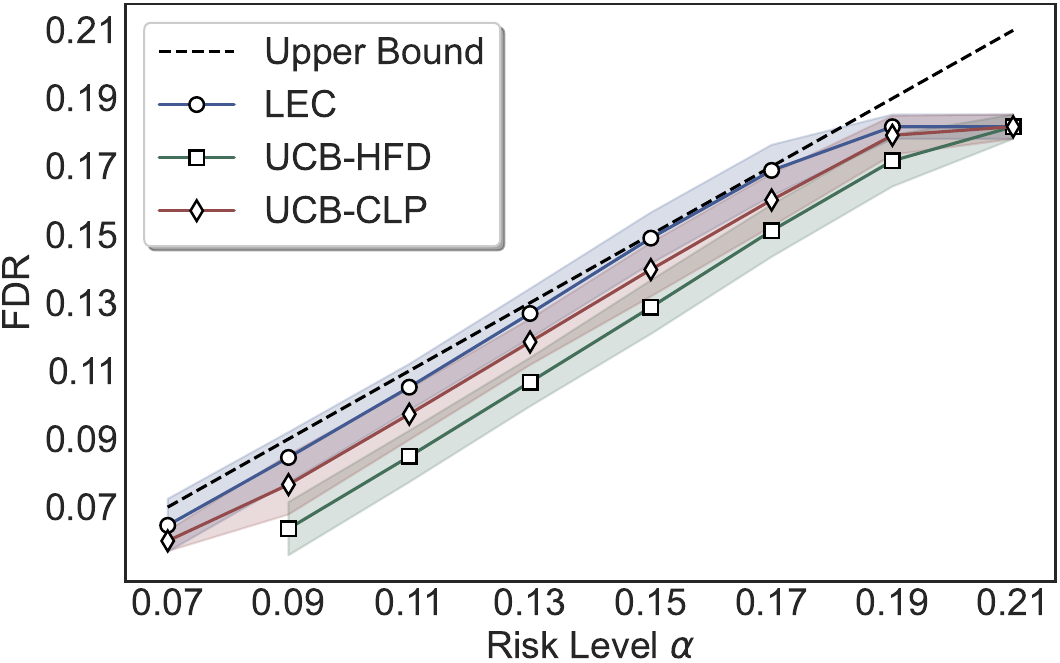}
    \caption{OpenChat-3.5 (FDR).}
  \end{subfigure}
  \hfill
  \begin{subfigure}[b]{0.246\textwidth}
    \centering
    \includegraphics[width=\textwidth]{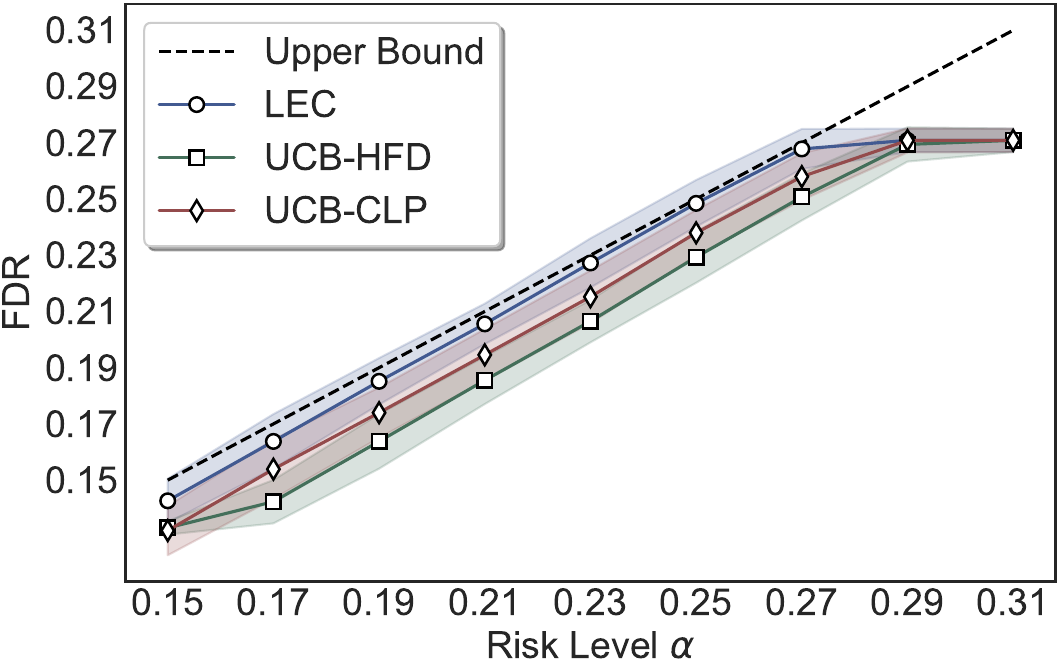}
    \caption{LLaMA-3.1-8B (FDR).}
  \end{subfigure}
  \hfill
  \begin{subfigure}[b]{0.246\textwidth}
    \centering
    \includegraphics[width=\textwidth]{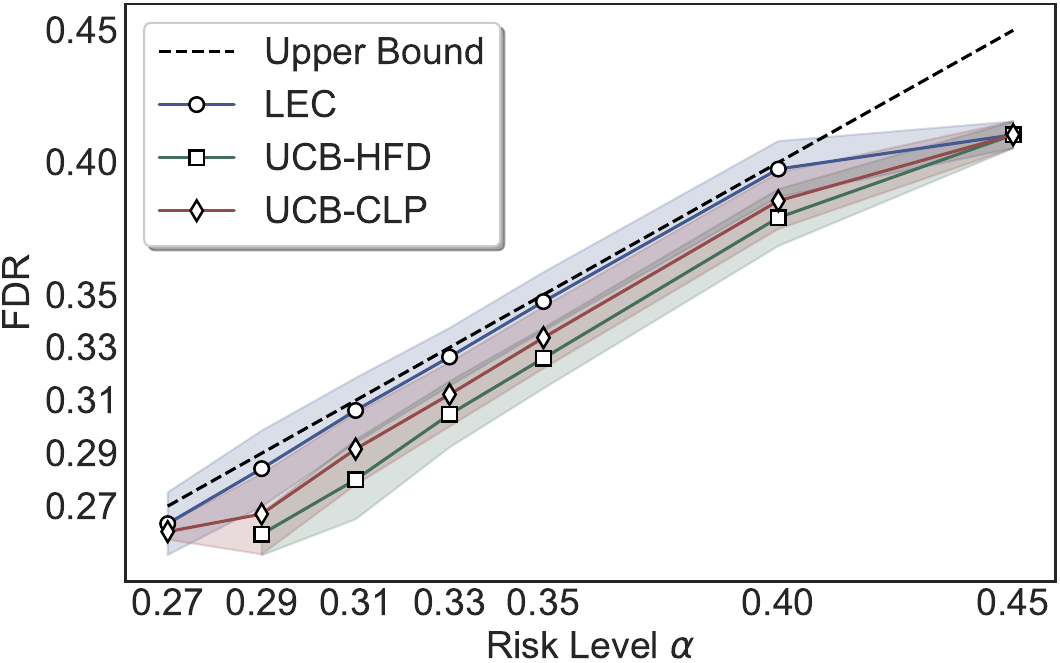}
    \caption{Vicuna-7B-V1.5 (FDR).}
  \end{subfigure}
  \hfill
  \begin{subfigure}[b]{0.246\textwidth}
    \centering
    \includegraphics[width=\textwidth]{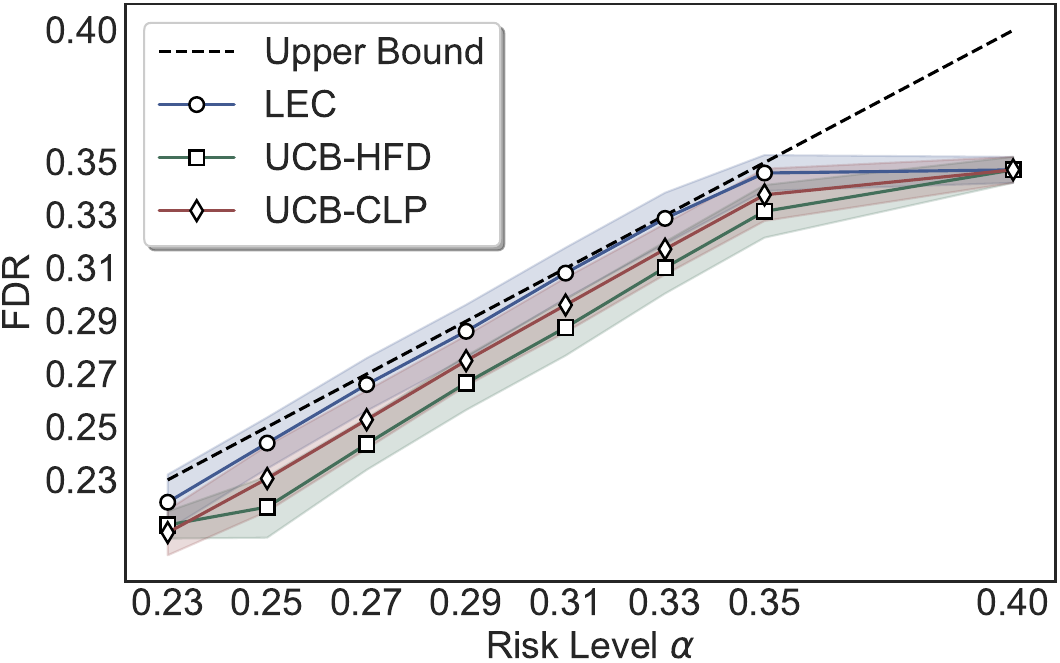}
    \caption{Vicuna-13B-V1.5 (FDR).}
  \end{subfigure}

  \begin{subfigure}[b]{0.246\textwidth}
    \centering
    \includegraphics[width=\textwidth]{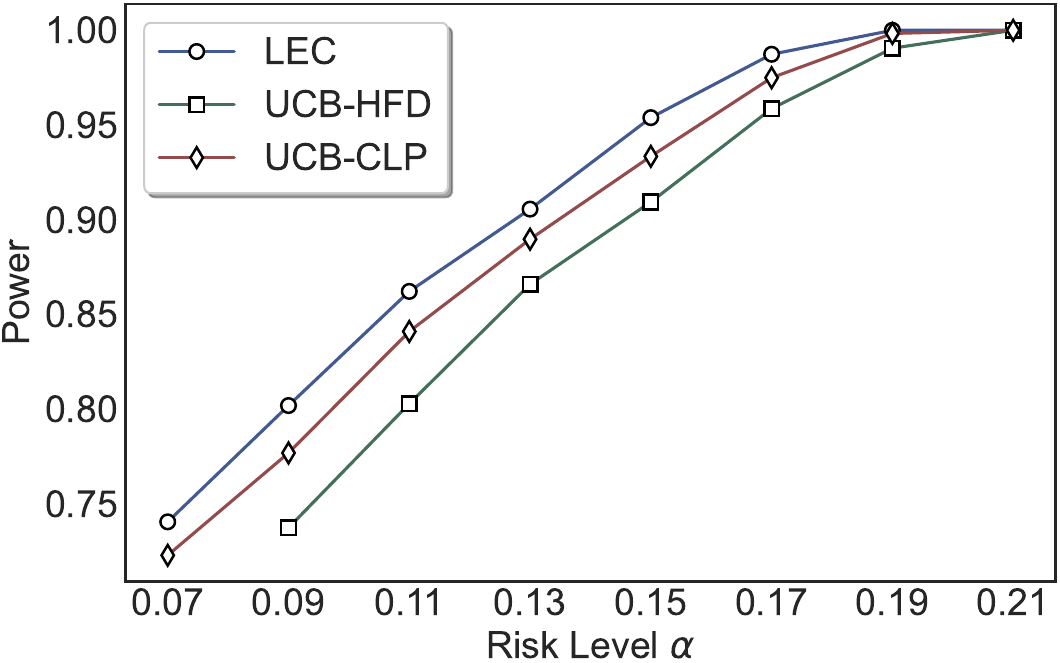}
    \caption{OpenChat-3.5 (Power).}
  \end{subfigure}
  \hfill
  \begin{subfigure}[b]{0.246\textwidth}
    \centering
    \includegraphics[width=\textwidth]{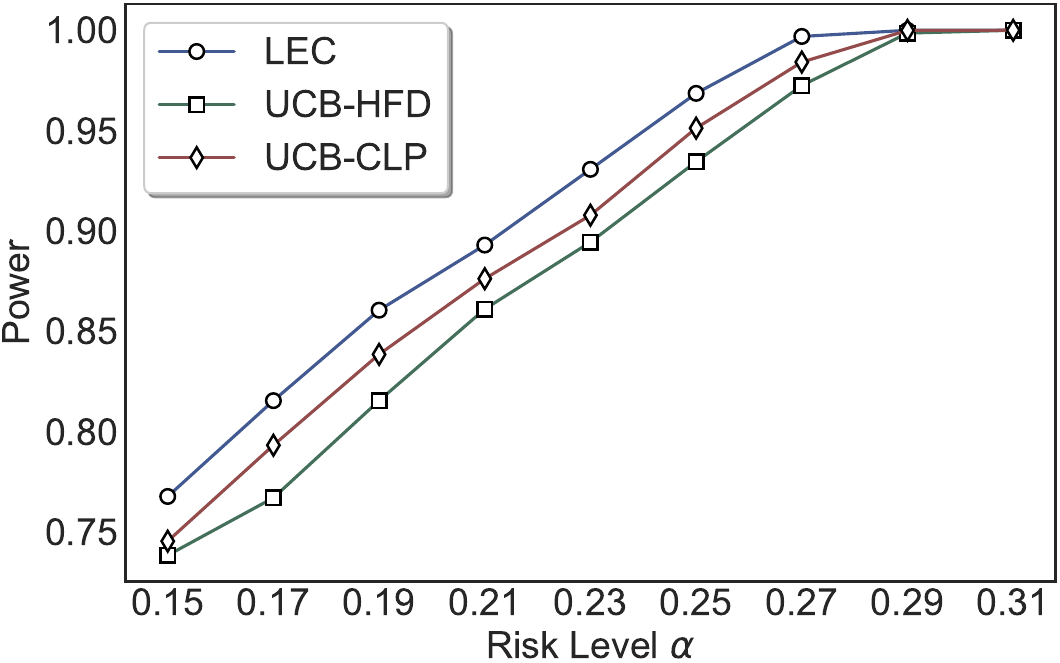}
    \caption{LLaMA-3.1-8B (Power).}
  \end{subfigure}
  \hfill
  \begin{subfigure}[b]{0.246\textwidth}
    \centering
    \includegraphics[width=\textwidth]{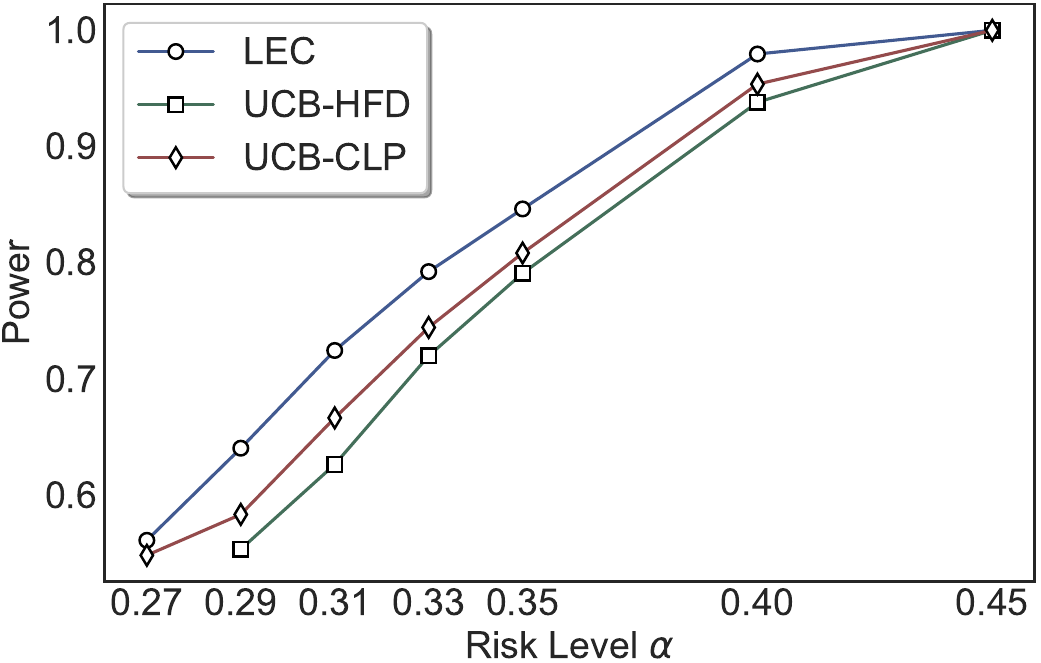}
    \caption{Vicuna-7B-V1.5 (Power).}
  \end{subfigure}
  \hfill
  \begin{subfigure}[b]{0.246\textwidth}
    \centering
    \includegraphics[width=\textwidth]{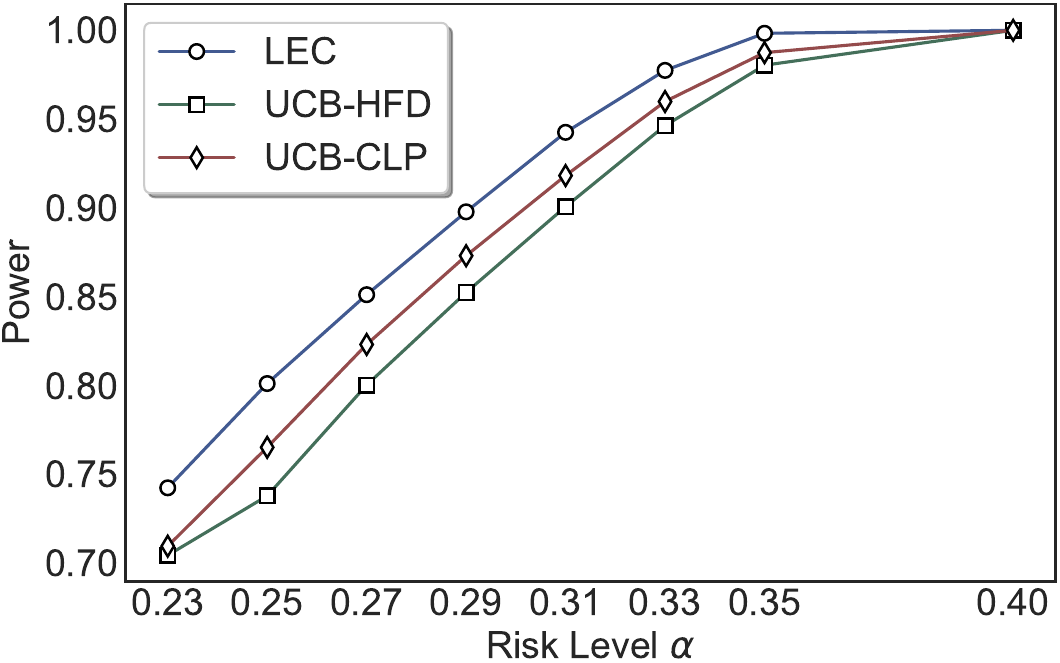}
    \caption{Vicuna-13B-V1.5 (Power).}
  \end{subfigure}

      \caption{Test-time empirical selection-conditioned error rate (mean$\pm$std) and power (mean) on CommonsenseQA using black-box PE.}
  \label{fig: Test-time FDR on the CommonsenseQA dataset using black-box PE (mean±std).}
\end{figure*}

\noindent \textbf{Evaluation of Statistical Validity and Power on CommonsenseQA at Black-Box Settings.} 
Figure~\ref{fig: Test-time FDR on the CommonsenseQA dataset using black-box PE (mean±std).} reports the test-time selection-conditioned error rate and power on the CommonsenseQA dataset using black-box PE across multiple LLMs. 
As presented in Figures~\ref{fig: Test-time FDR on the CommonsenseQA dataset using black-box PE (mean±std).} (a)--(d), \texttt{LEC} consistently achieves valid selection-conditioned risk control at test time, with the realized selection-conditioned error rate remaining below the target risk level across all models and risk settings. Despite the increased noise and reduced resolution typically associated with black-box uncertainty estimates, the proposed framework preserves its finite-sample guarantees, indicating that \texttt{LEC} does not rely on privileged model information. 
In terms of power, Figures~\ref{fig: Test-time FDR on the CommonsenseQA dataset using black-box PE (mean±std).} (e)–(h) demonstrate that \texttt{LEC} consistently retains more admissible samples than UCB-based baselines under the same risk constraints. 
These results confirm that the advantages of \texttt{LEC} extend naturally to black-box settings, where only limited uncertainty information is available.

\noindent \textbf{Power Analysis on CommonsenseQA.} 
Table~\ref{tab: power comparison (CommonsenseQA) white se} reports a comprehensive comparison of power on CommonsenseQA across eight LLMs and various risk levels. 
Consistent with the main results, \texttt{LEC} uniformly achieves higher power than both UCB-based baselines under the same target risk constraints. 
This trend holds across all evaluated models and becomes particularly pronounced at low to moderate risk levels, where conservative calibration has the largest impact on sample retention. 
Notably, for several models, including Vicuna-13B-V1.5, \texttt{LEC} remains feasible at substantially lower risk levels where UCB-based methods either yield significantly lower power or fail to identify valid thresholds. This further highlights the advantage of expectation-level risk control in retaining admissible samples under strict reliability requirements. 

\input{tables/powerCommonsenseQA}

\begin{figure*}[!t]
  \centering
  \begin{subfigure}[b]{0.45\textwidth}
    \centering
    \includegraphics[width=\textwidth]{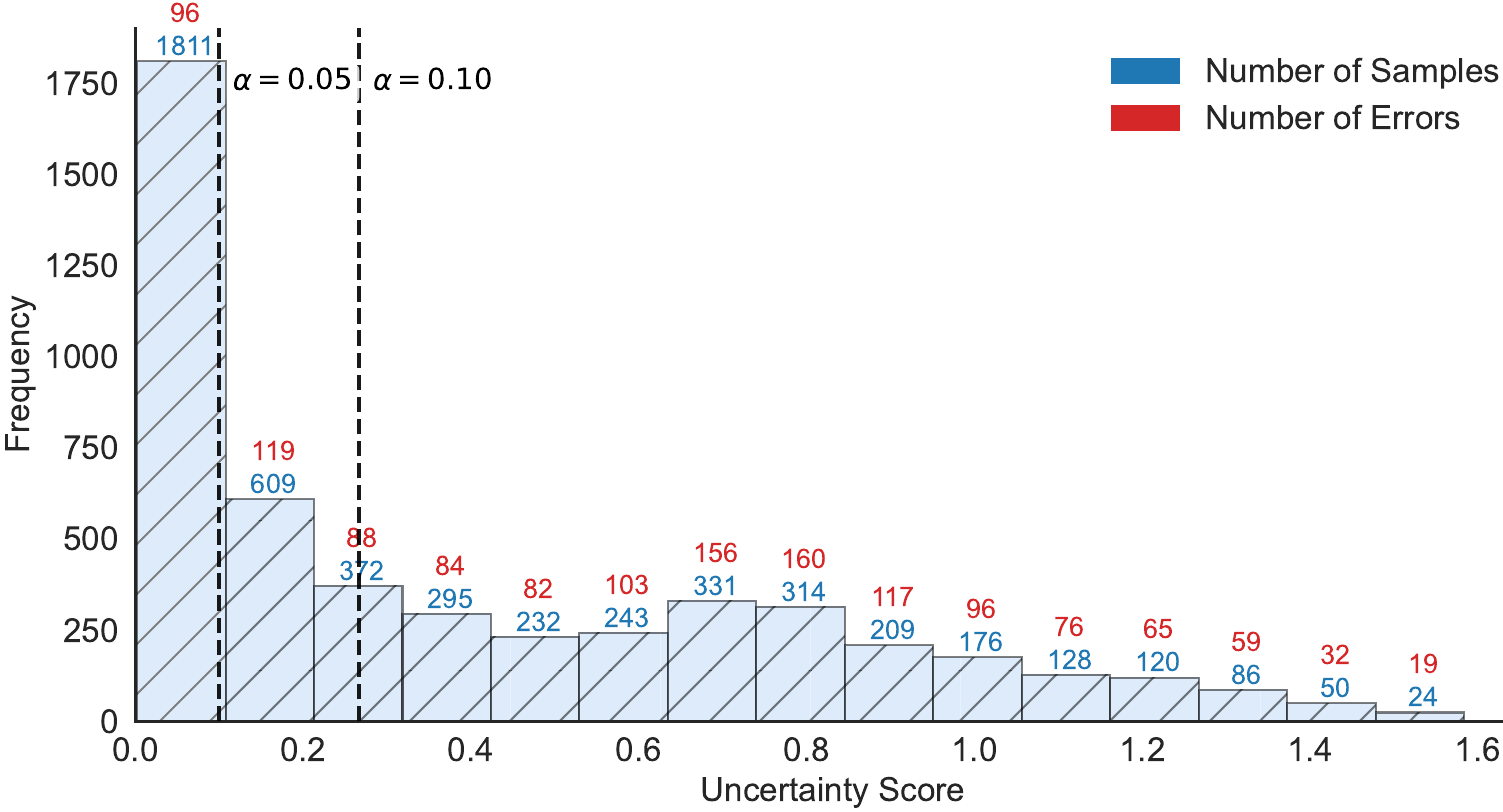}
    \caption{LLaMA-3.1-8B.}
  \end{subfigure}
  \hspace{3mm}
  \begin{subfigure}[b]{0.45\textwidth}
    \centering
    \includegraphics[width=\textwidth]{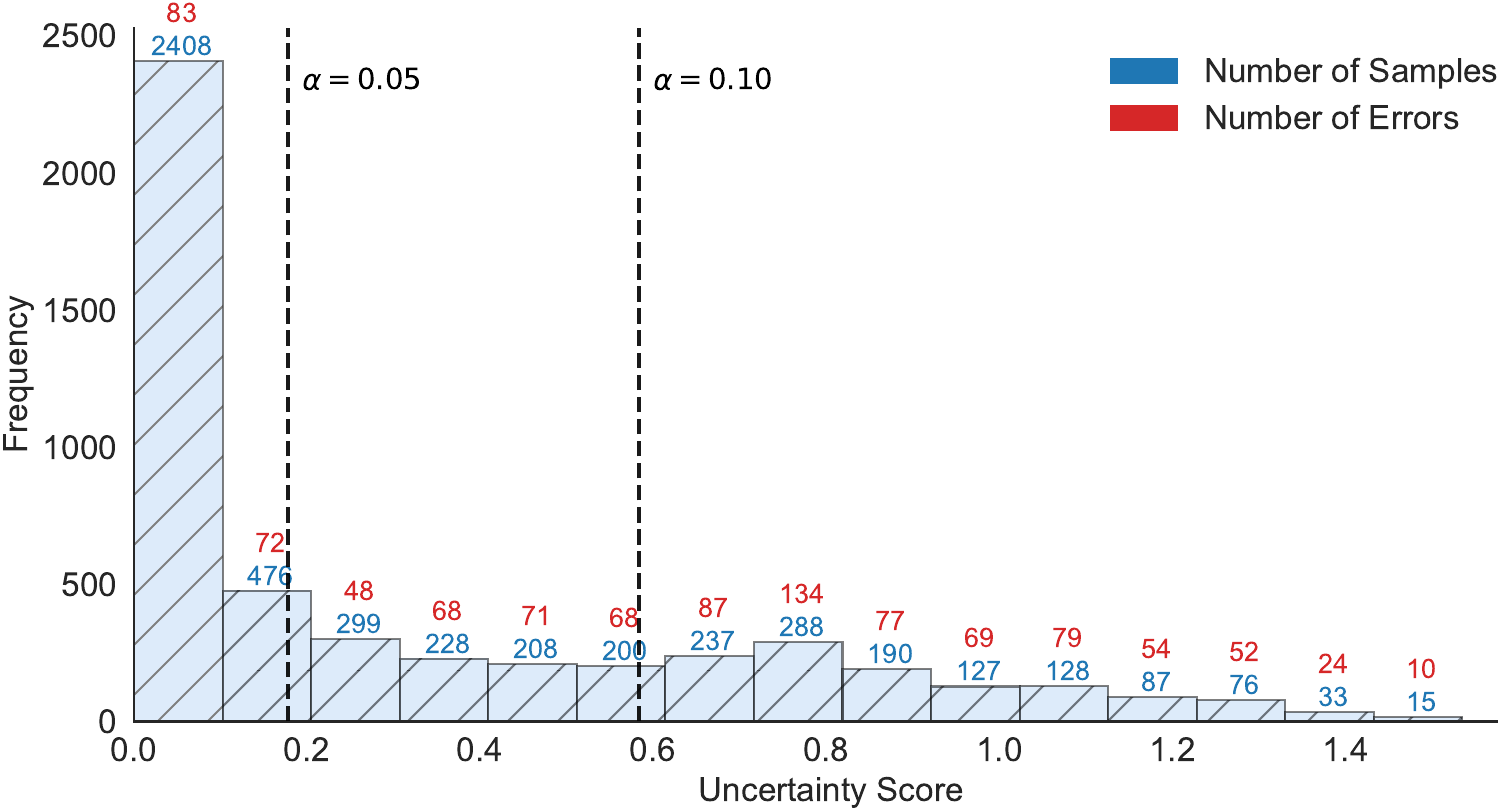}
    \caption{LLaMA-3.1-70B.}
  \end{subfigure}
  
    \caption{Uncertainty and correctness distribution on the CommonsenseQA dataset.}
  \label{fig: uncertainty distribution}
\end{figure*}

\noindent \textbf{Uncertainty-Correctness Distribution.} 
Figure~\ref{fig: uncertainty distribution} visualizes the joint distribution of uncertainty scores and correctness on the test set for LLaMA-3.1-8B and LLaMA-3.1-70B. 
Each histogram bin reports the number of test samples within a given uncertainty range, along with the corresponding number of incorrect predictions. 
A key observation is that incorrect predictions are present across nearly all uncertainty intervals, rather than being confined to a small high-uncertainty region. This indicates that selective prediction inherently involves a trade-off between rejecting uncertain samples and retaining correct ones, and that no single uncertainty threshold can perfectly separate correct and incorrect predictions.

Importantly, the two models exhibit markedly different uncertainty-correctness profiles. Compared to LLaMA-3.1-8B, LLaMA-3.1-70B assigns lower uncertainty scores to a larger fraction of correct predictions, while maintaining a comparable or lower error density in the low-uncertainty region. As a result, for the same target risk level, the larger model can accept a greater number of correct samples before violating the risk constraint, leading to consistently higher power. 
These results reinforce that the performance gains of \texttt{LEC} are driven not only by tighter calibration, but also by its ability to adapt to model-specific uncertainty–correctness characteristics. By directly constraining expected system-level risk, \texttt{LEC} effectively leverages favorable uncertainty distributions to retain more admissible samples, while preserving rigorous statistical validity.

\begin{figure*}[!t]
  \centering
  \begin{subfigure}[b]{0.246\textwidth}
    \centering
    \includegraphics[width=\textwidth]{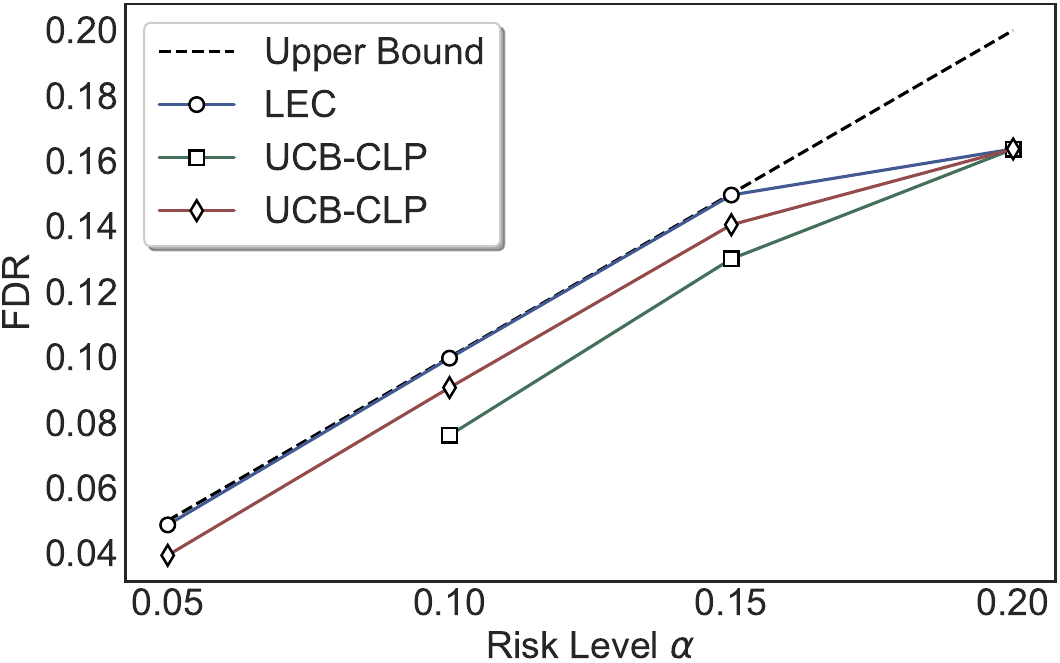}
    \caption{Deg (FDR).}
  \end{subfigure}
  \hfill
  \begin{subfigure}[b]{0.246\textwidth}
    \centering
    \includegraphics[width=\textwidth]{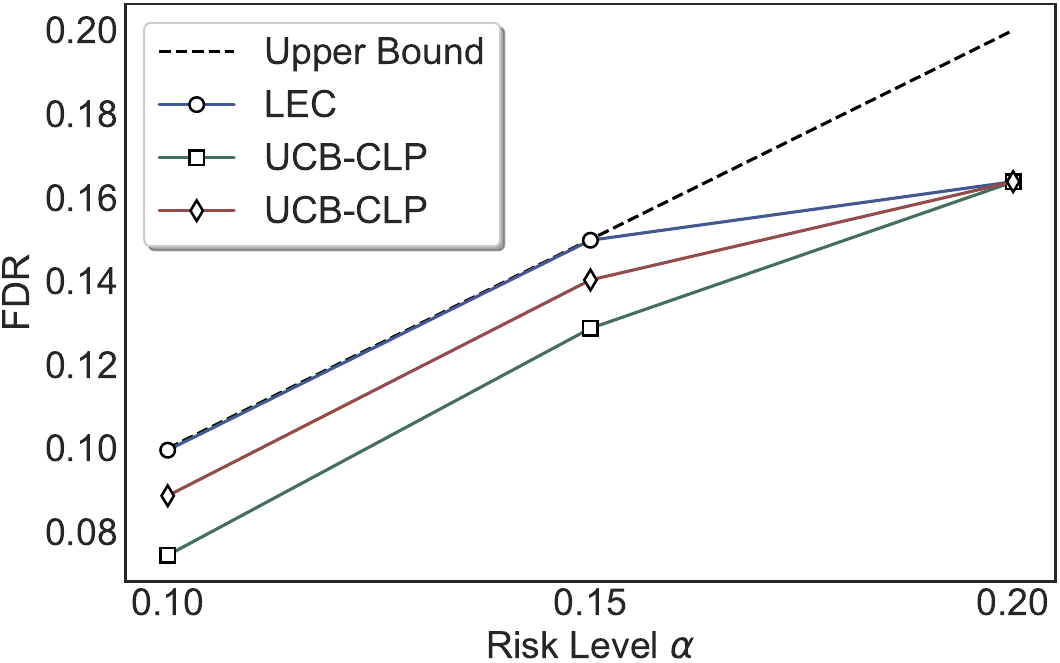}
    \caption{Ecc (FDR).}
  \end{subfigure}
  \hfill
  \begin{subfigure}[b]{0.246\textwidth}
    \centering
    \includegraphics[width=\textwidth]{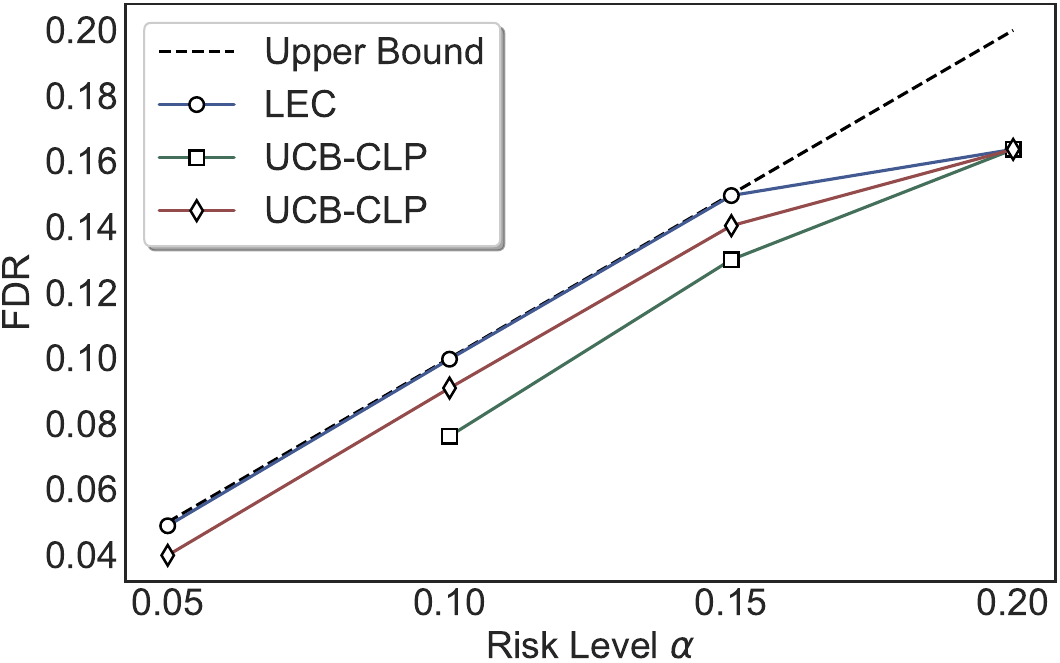}
    \caption{EigV (FDR).}
  \end{subfigure}
  \hfill
  \begin{subfigure}[b]{0.246\textwidth}
    \centering
    \includegraphics[width=\textwidth]{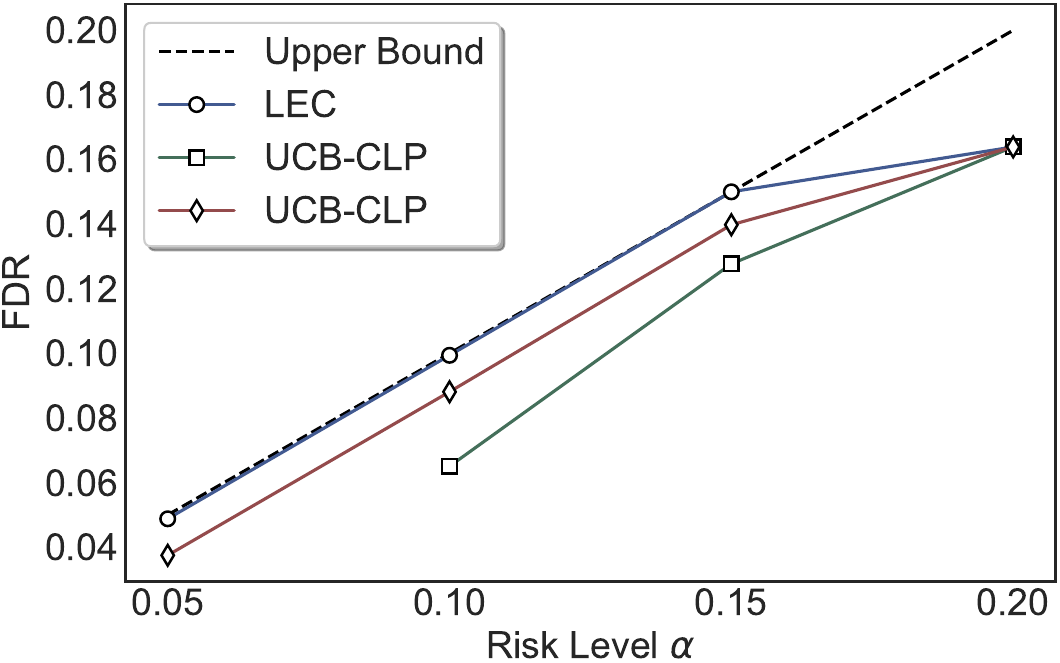}
    \caption{SELF (FDR).}
  \end{subfigure}

  \begin{subfigure}[b]{0.246\textwidth}
    \centering
    \includegraphics[width=\textwidth]{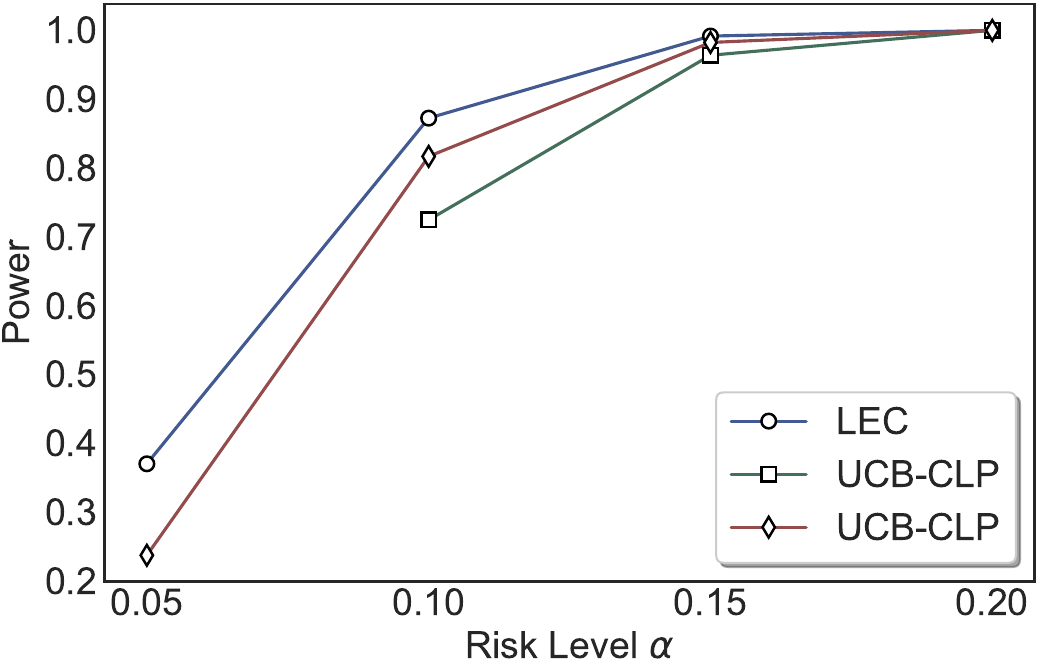}
    \caption{Deg (Power).}
  \end{subfigure}
  \hfill
  \begin{subfigure}[b]{0.246\textwidth}
    \centering
    \includegraphics[width=\textwidth]{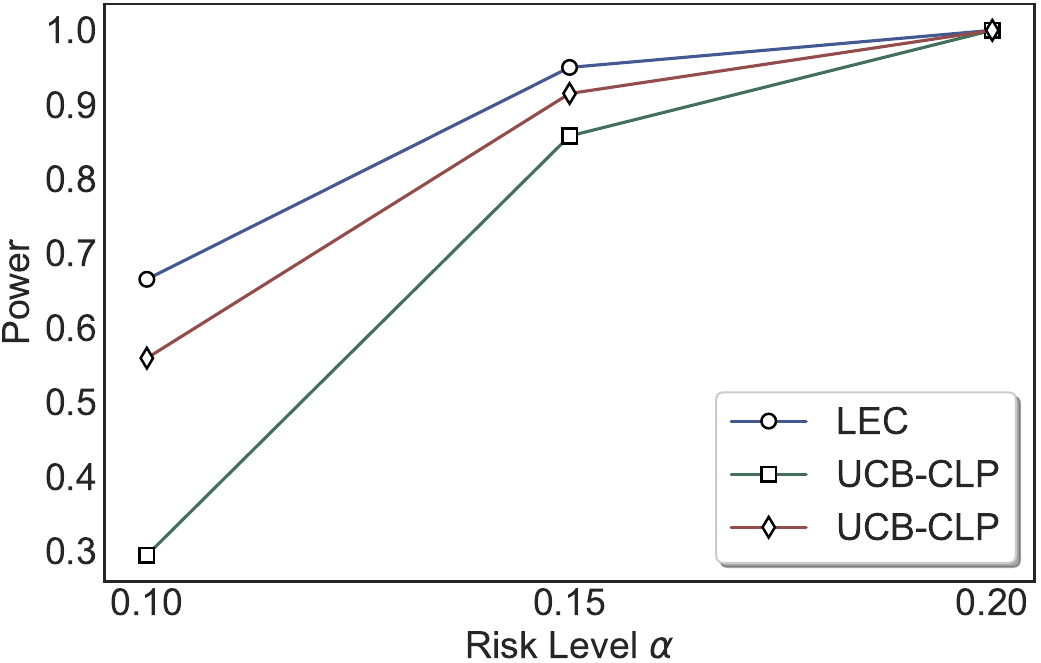}
    \caption{Ecc (Power).}
  \end{subfigure}
  \hfill
  \begin{subfigure}[b]{0.246\textwidth}
    \centering
    \includegraphics[width=\textwidth]{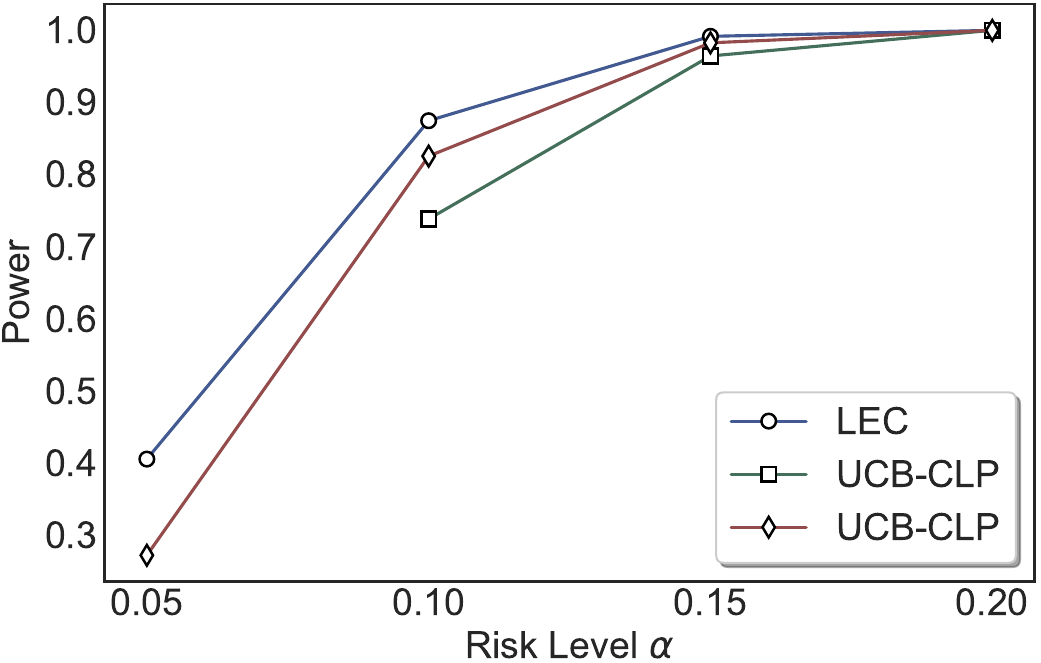}
    \caption{EigV (Power).}
  \end{subfigure}
  \hfill
  \begin{subfigure}[b]{0.246\textwidth}
    \centering
    \includegraphics[width=\textwidth]{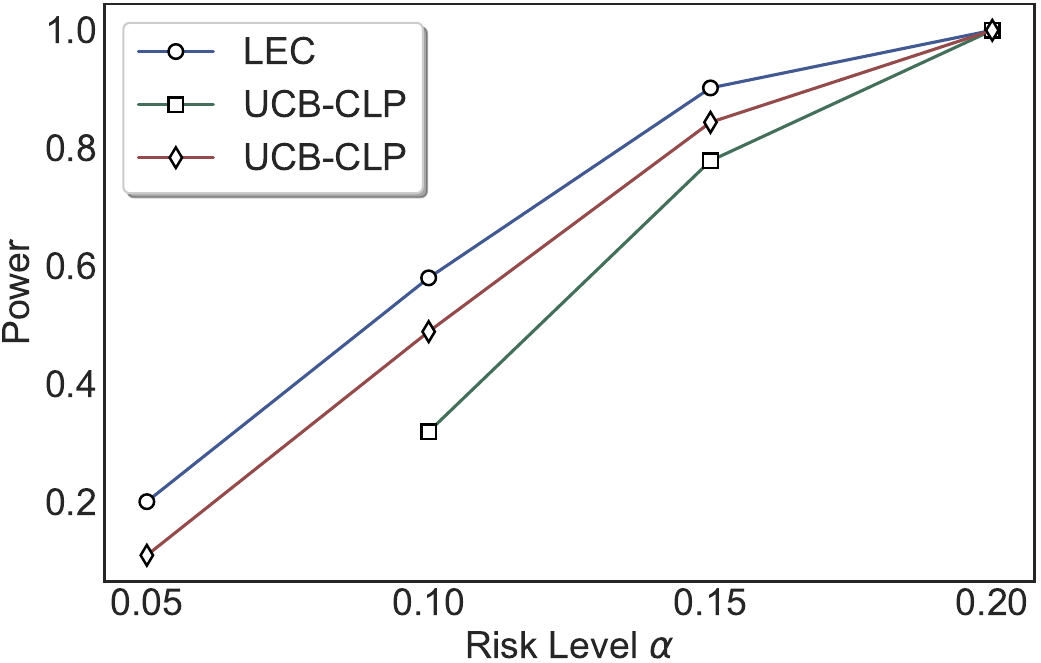}
    \caption{SELF (Power).}
  \end{subfigure}

  \caption{Test-time empirical selection-conditioned error rate and power on the TriviaQA dataset with the Qwen2.5-3B model using different UQ methods (mean). 
\texttt{LEC} provides tighter risk control while retaining more correct samples.}
  \label{fig: Test-time FDR and Power on the TriviaQA dataset using different UQ (mean).}
\end{figure*}

\noindent \textbf{Robustness and Efficiency across UQ Methods and Sampling Sizes on TriviaQA.} 
Figure~\ref{fig: Test-time FDR and Power on the TriviaQA dataset using different UQ (mean).} evaluates \texttt{LEC} on TriviaQA using four different uncertainty estimators (Deg, Ecc, EigV, and SELF). 
Across all uncertainty methods, \texttt{LEC} consistently achieves tighter risk control and higher power than UCB-based baselines, demonstrating that the benefits of \texttt{LEC} are orthogonal to the specific choice of uncertainty estimator. 
Table~\ref{tab:csp_sampling_combined_mean} further examines the effect of sampling size for SE and EigV uncertainty estimators on LLaMA-3.1-70B.
Even with as few as 5 samples, \texttt{LEC} maintains valid risk control, highlighting its strong finite-sample efficiency. As the sampling size increases, both AUROC and power improve monotonically, while selection-conditioned error rate remains tightly controlled. This trend confirms that \texttt{LEC} is able to effectively translate improvements in uncertainty quality into tangible gains in selective prediction performance, without requiring large calibration budgets.

\input{tables/samplingSize}

\begin{figure*}[!t]
  \centering
 \begin{subfigure}[b]{0.3\textwidth}
    \centering
    \includegraphics[width=\textwidth]{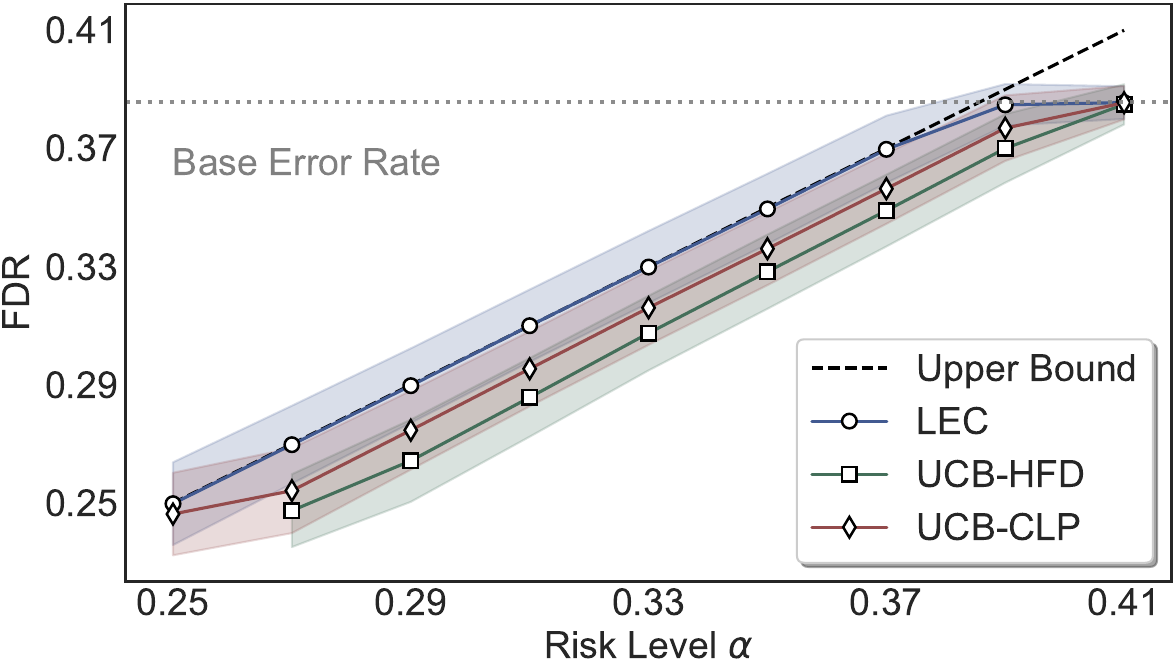}
    \caption{Qwen2.5-3B (FDR).}
  \end{subfigure}
  \hspace{2mm}
  \begin{subfigure}[b]{0.3\textwidth}
    \centering
    \includegraphics[width=\textwidth]{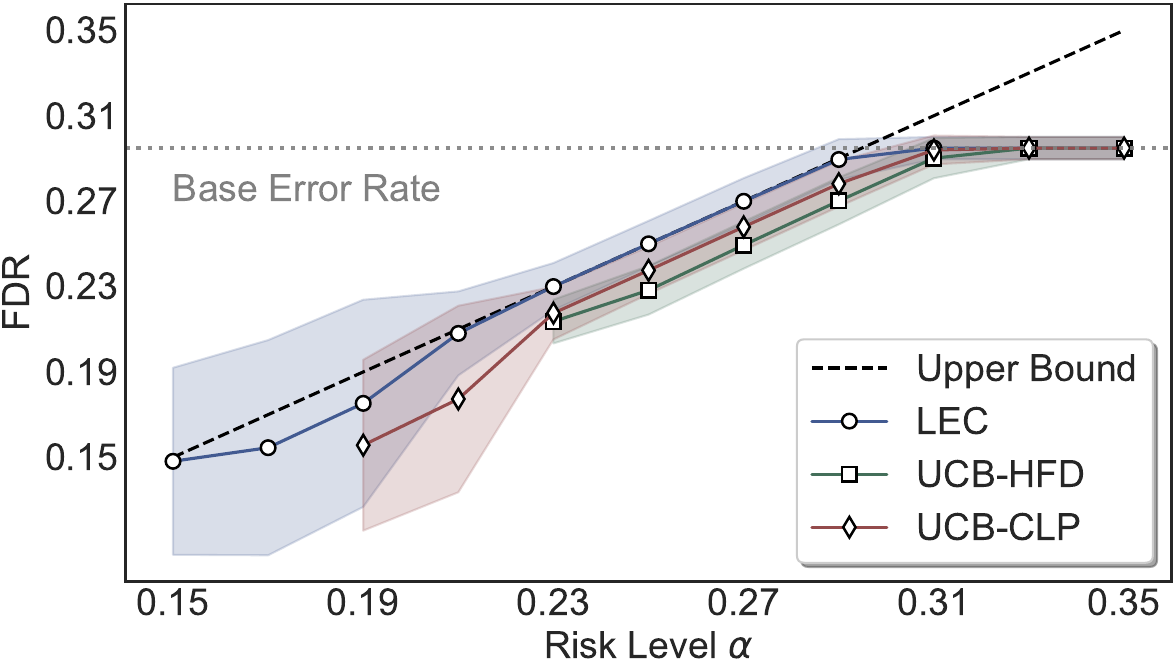}
    \caption{Qwen2.5-7B (FDR).}
  \end{subfigure}
  \hspace{2mm}
  \begin{subfigure}[b]{0.3\textwidth}
    \centering
    \includegraphics[width=\textwidth]{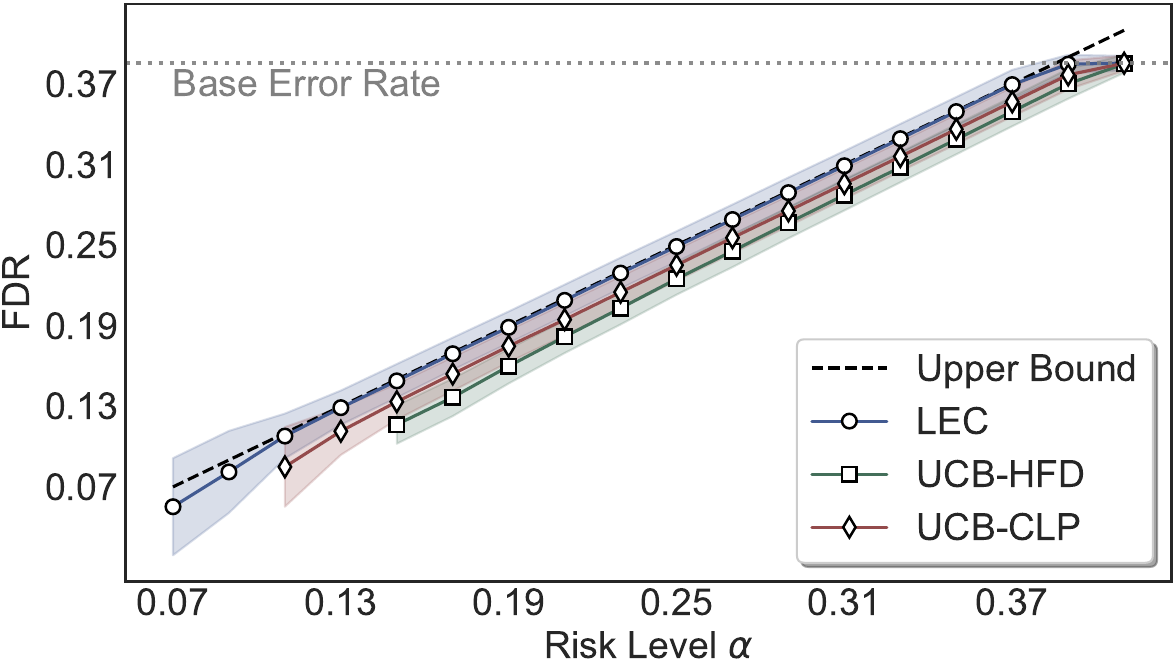}
    \caption{Vicuna-7B-V1.5 (FDR).}
  \end{subfigure}
  
 \begin{subfigure}[b]{0.3\textwidth}
    \centering
    \includegraphics[width=\textwidth]{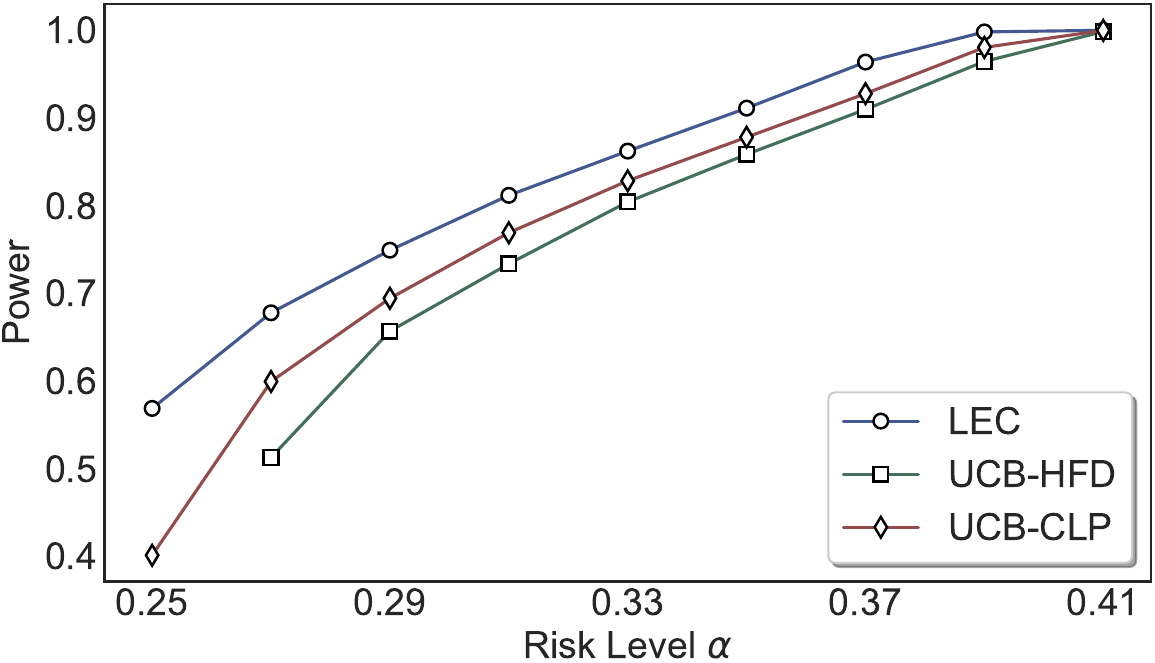}
    \caption{Qwen2.5-3B (Power).}
  \end{subfigure}
  \hspace{2mm}
  \begin{subfigure}[b]{0.3\textwidth}
    \centering
    \includegraphics[width=\textwidth]{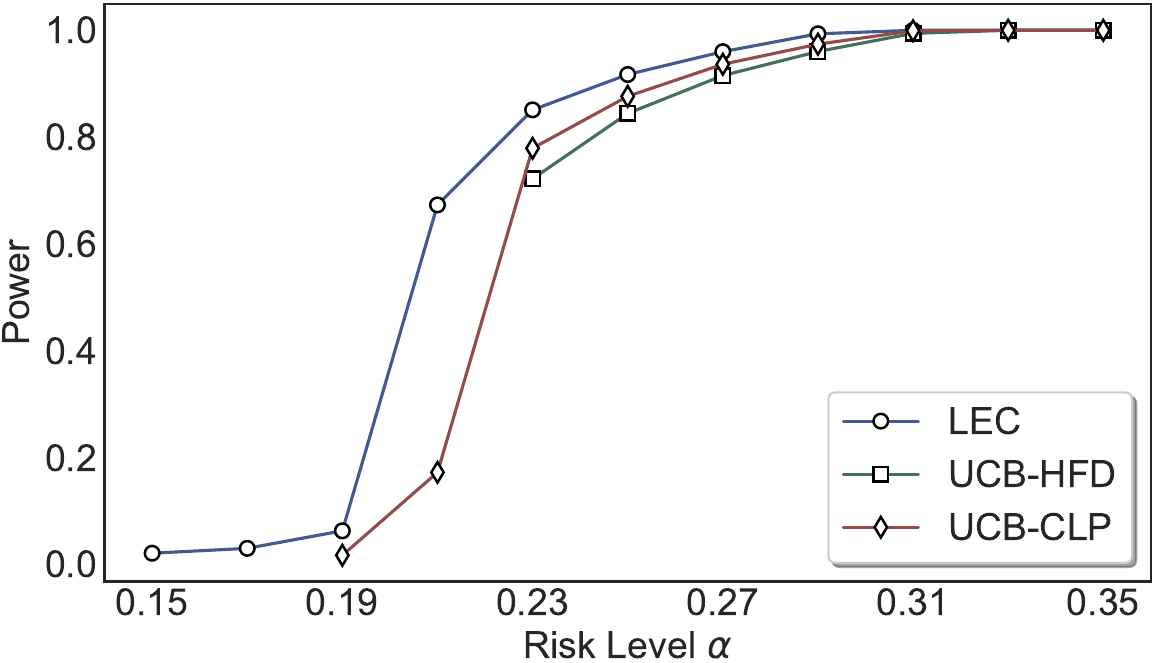}
    \caption{Qwen2.5-7B (Power).}
  \end{subfigure}
  \hspace{2mm}
  \begin{subfigure}[b]{0.3\textwidth}
    \centering
    \includegraphics[width=\textwidth]{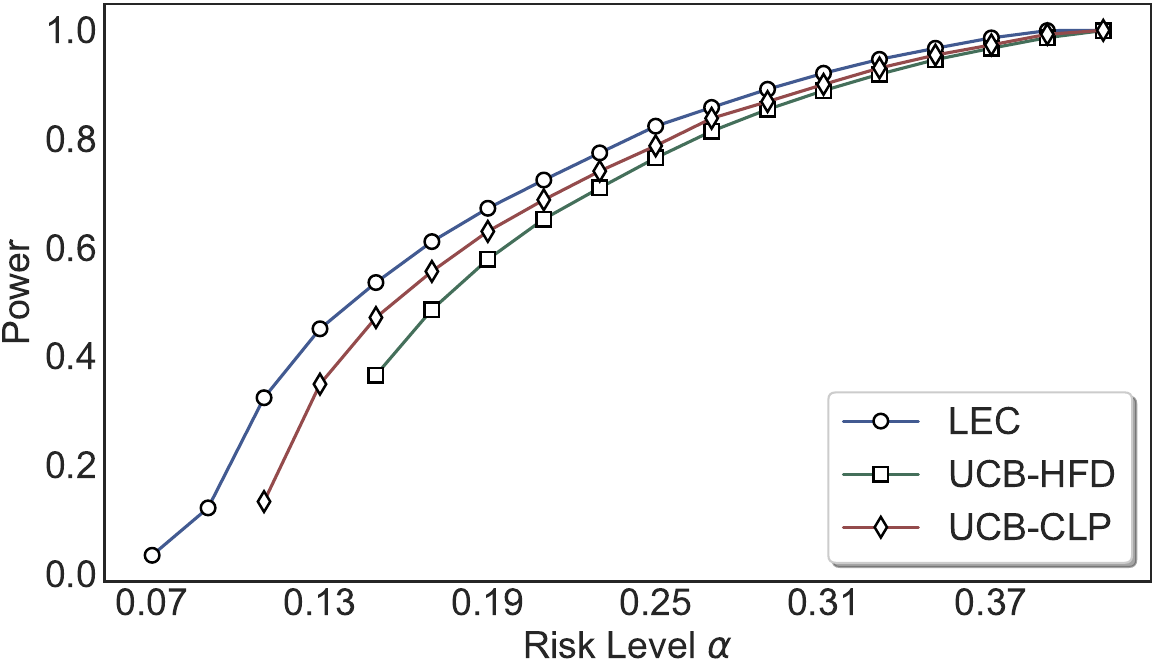}
    \caption{Vicuna-7B-V1.5 (Power).}
  \end{subfigure}

      \caption{Test-time empirical selection-conditioned error rate (mean$\pm$std) and power (mean) on the TriviaQA dataset with LLM-as-a-Judge for correctness evaluation. }
  \label{fig: Test-time Power (mean) on the TriviaQA dataset with LLM-as-a-Judge for correctness evaluation.}
\end{figure*}

\begin{figure*}[!t]
    \centering
    \includegraphics[width=0.82\linewidth]{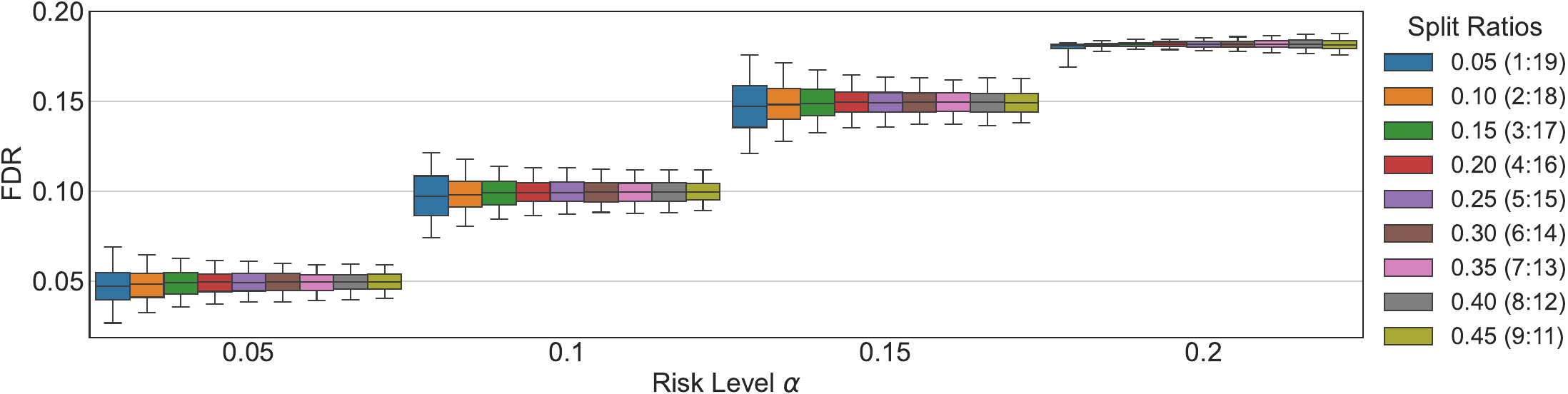}
    \caption{Empirical selection-conditioned error control across various calibration-test split ratios on CommonsenseQA with the OpenChat-3.5 model. }
    \label{fig: FDR control across various calibration-test split ratios on the CommonsenseQA dataset with the OpenChat-3.5 model.}
\end{figure*}

\begin{figure*}[!t]
    \centering
    \includegraphics[width=0.82\linewidth]{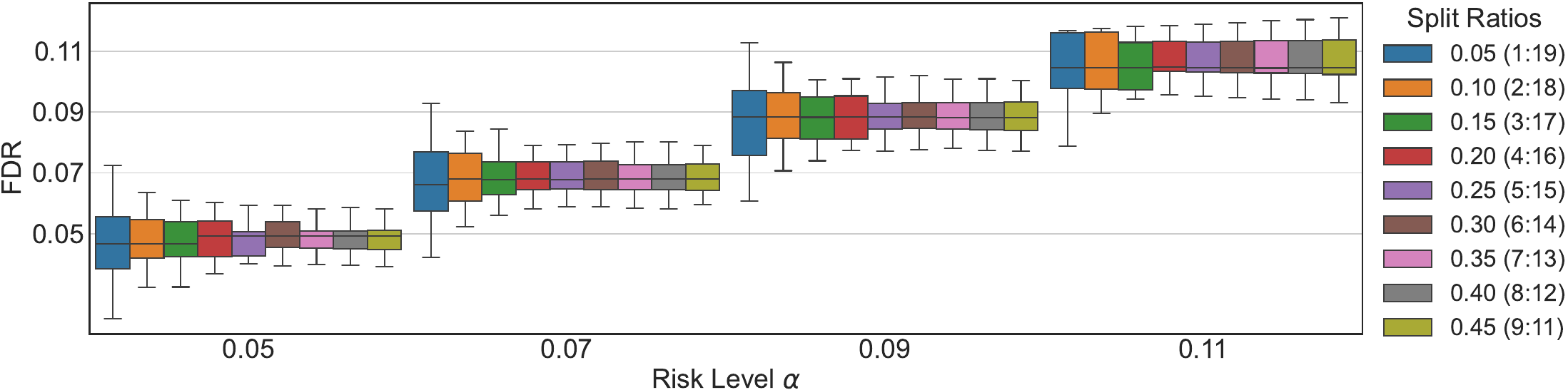}
    \caption{Empirical selection-conditioned error control across various calibration-test split ratios on TriviaQA with the LLaMA-3.1-8B model. }
    \label{fig: FDR control across various calibration-test split ratios on the TriviaQA dataset with the LLaMA-3.1-8B model.}
\end{figure*}

\noindent \textbf{Evaluation with LLM-as-a-Judge for Correctness Assessment.} 
Figure~\ref{fig: Test-time Power (mean) on the TriviaQA dataset with LLM-as-a-Judge for correctness evaluation.} reports the test-time selection-conditioned error rate and power on TriviaQA when correctness is evaluated using an LLM-as-a-Judge instead of exact string matching or similarity-based metrics. This setting introduces an additional layer of uncertainty, as the correctness labels themselves are noisy and may vary across prompts or judging criteria. 
Despite this increased label noise, \texttt{LEC} consistently maintains valid risk control across all evaluated models, with the empirical test-time selection-conditioned error rate closely tracking the target risk level and remaining below the theoretical upper bound. 
More importantly, \texttt{LEC} continues to achieve strictly higher power than both \texttt{UCB-HFD} and \texttt{UCB-CLP} across all models. 
The gap is especially pronounced for medium-sized models such as Qwen2.5-7B, where UCB-based methods suffer a sharp drop in power at low risk levels, while \texttt{LEC} is able to retain a substantial fraction of correct predictions. 
This behavior is consistent with the design of \texttt{LEC}: by enforcing a finite-sample linear constraint on the aggregate system behavior, \texttt{LEC} avoids overreacting to spurious or judge-induced errors that disproportionately inflate confidence bounds in UCB-style calibration. 
Overall, these results demonstrate that \texttt{LEC} remains robust under noisy and subjective correctness evaluation schemes. This robustness is particularly important for open-ended question answering tasks, where exact correctness is often ill-defined and LLM-as-a-Judge–style evaluation is increasingly adopted in practice.

\noindent \textbf{Effect of Calibration–Test Split Ratios.} 
Figures~\ref{fig: FDR control across various calibration-test split ratios on the CommonsenseQA dataset with the OpenChat-3.5 model.} and~\ref{fig: FDR control across various calibration-test split ratios on the TriviaQA dataset with the LLaMA-3.1-8B model.} study the effect of different calibration–test split ratios. 
Across both CommonsenseQA and TriviaQA, LEC maintains valid risk control even when the calibration set is extremely small (e.g., split ratio 0.05, corresponding to only 500 calibration samples and 9500 test samples on CommonsenseQA). 
At the same time, these results clearly reflect the marginal nature of the theoretical guarantee: larger calibration sets lead to tighter concentration and reduced variance in test-time selection-conditioned error rate. 
As expected, increasing the calibration proportion reduces the standard deviation of the realized selection-conditioned error rate, reinforcing the practical benefit of allocating more data to calibration when possible. 

\begin{figure*}[!h]
  \centering
  \begin{subfigure}[b]{0.246\textwidth}
    \centering
    \includegraphics[width=\textwidth]{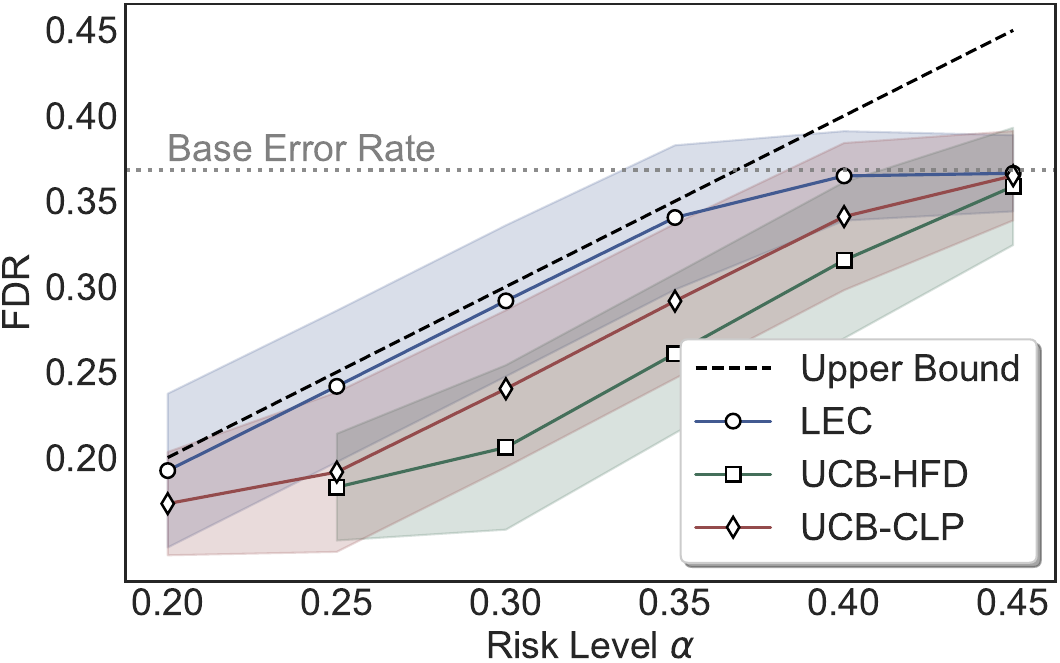}
    \caption{InternVL2-1B.}
  \end{subfigure}
  \hfill
  \begin{subfigure}[b]{0.246\textwidth}
    \centering
    \includegraphics[width=\textwidth]{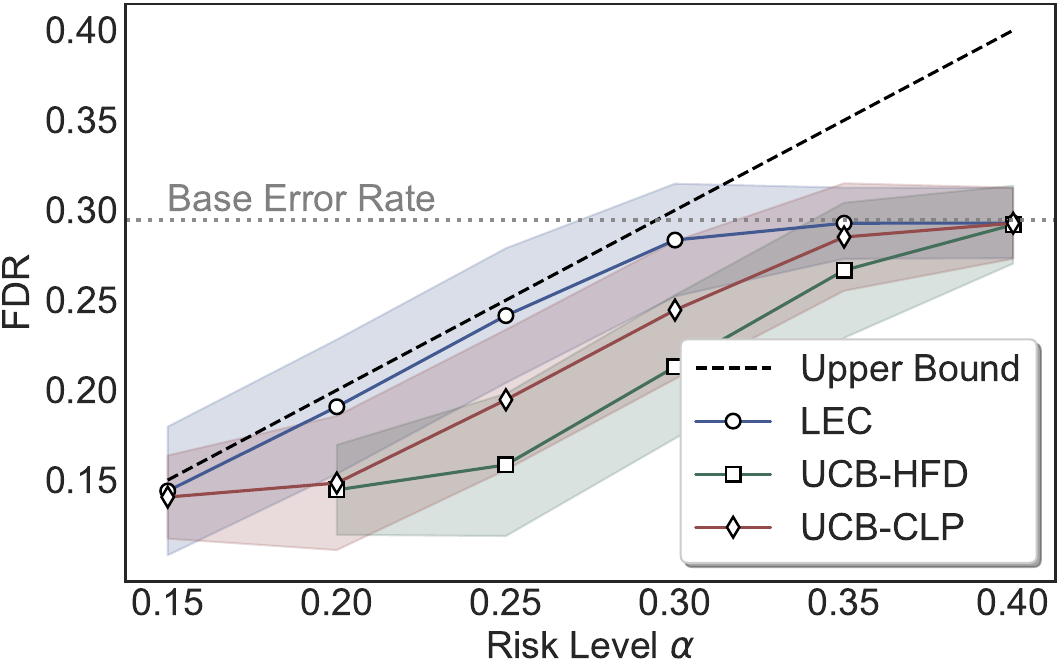}
    \caption{InternVL2-8B.}
  \end{subfigure}
  \hfill
  \begin{subfigure}[b]{0.246\textwidth}
    \centering
    \includegraphics[width=\textwidth]{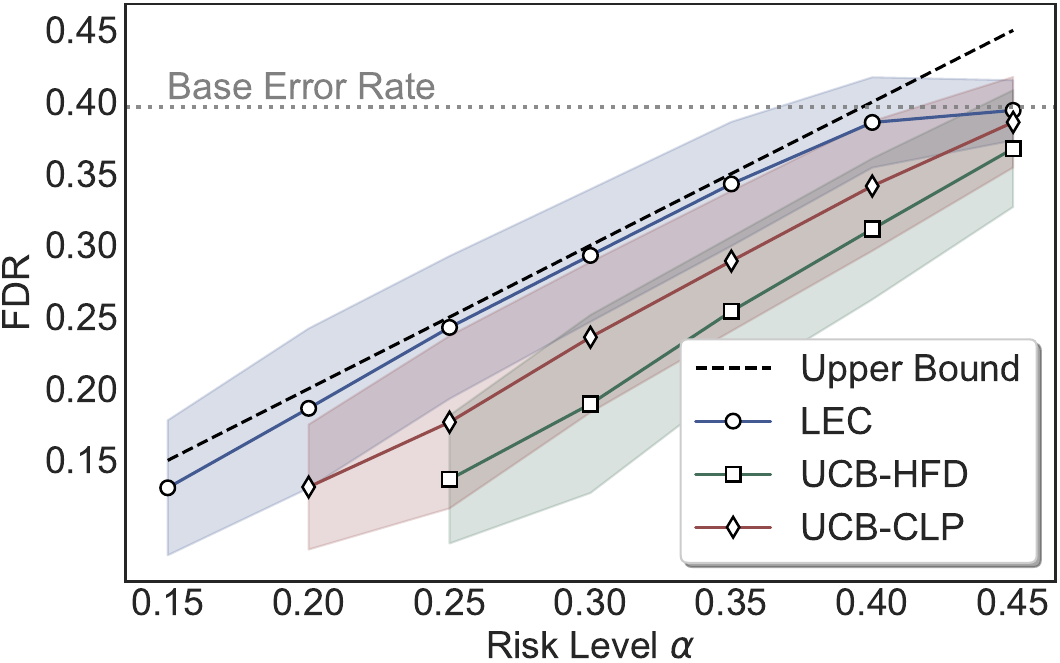}
    \caption{LLaVA-1.5-7B.}
  \end{subfigure}
  \hfill
  \begin{subfigure}[b]{0.246\textwidth}
    \centering
    \includegraphics[width=\textwidth]{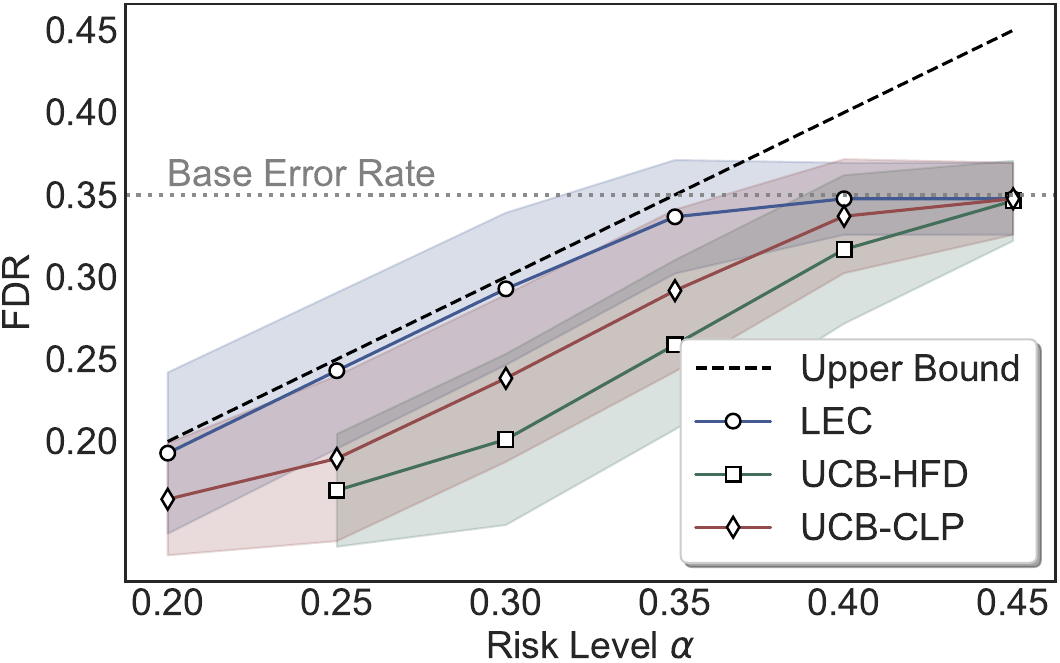}
    \caption{LLaVA-V1.6-Mistral-7B.}
  \end{subfigure}

  \begin{subfigure}[b]{0.246\textwidth}
    \centering
    \includegraphics[width=\textwidth]{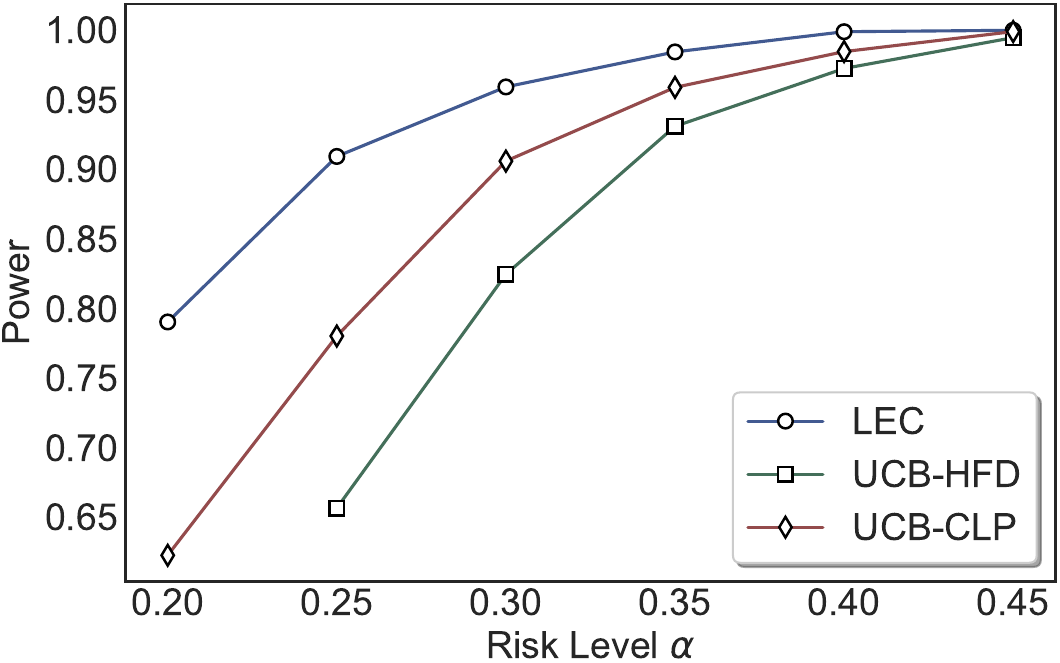}
    \caption{InternVL2-1B.}
  \end{subfigure}
  \hfill
  \begin{subfigure}[b]{0.246\textwidth}
    \centering
    \includegraphics[width=\textwidth]{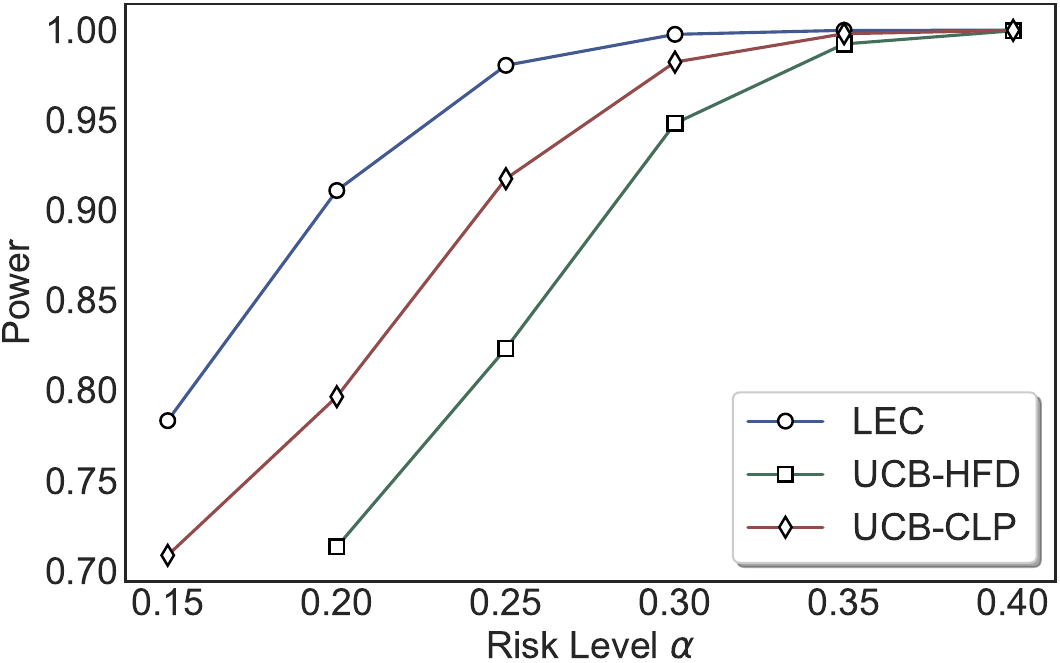}
    \caption{InternVL2-8B.}
  \end{subfigure}
  \hfill
  \begin{subfigure}[b]{0.246\textwidth}
    \centering
    \includegraphics[width=\textwidth]{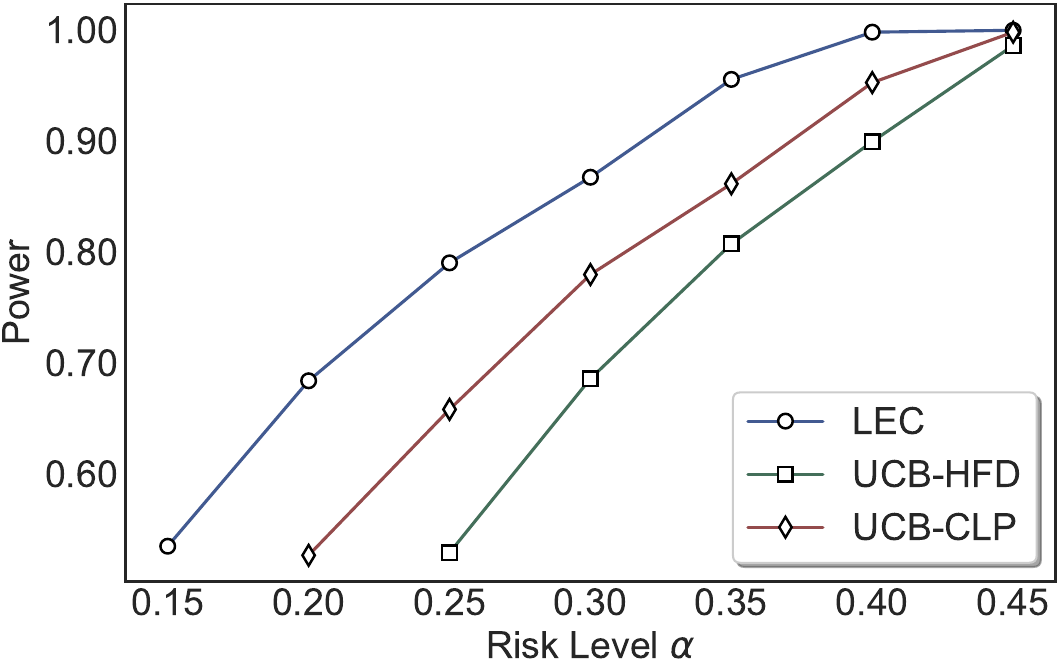}
    \caption{LLaVA-1.5-7B.}
  \end{subfigure}
  \hfill
  \begin{subfigure}[b]{0.246\textwidth}
    \centering
    \includegraphics[width=\textwidth]{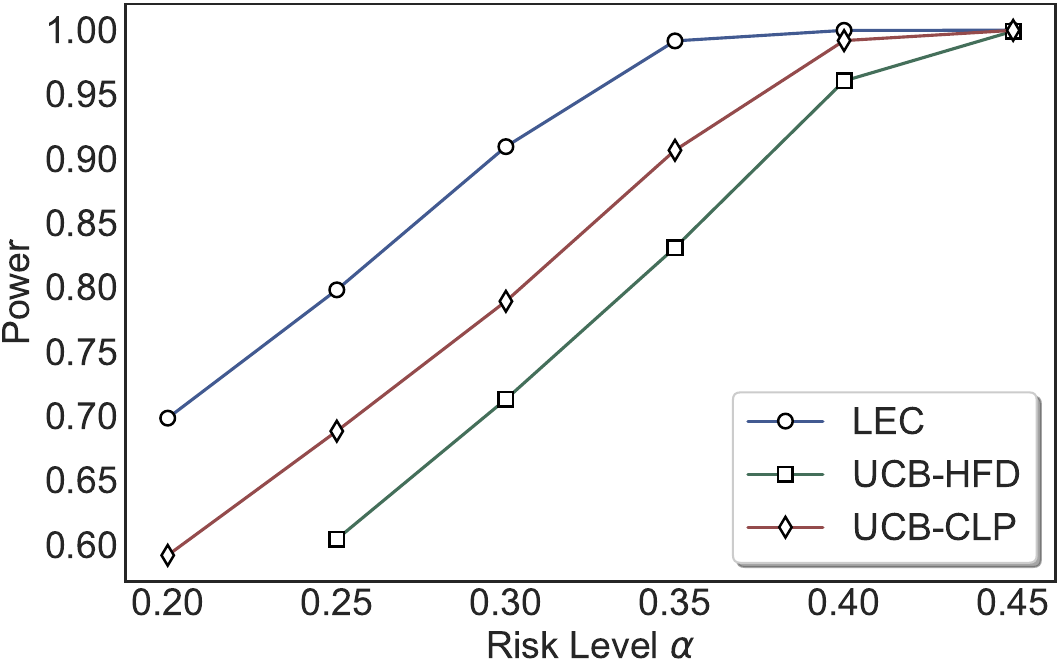}
    \caption{LLaVA-V1.6-Mistral-7B.}
  \end{subfigure}

   \caption{Test-time empirical selection-conditioned error rate (mean$\pm$std) and power (mean) on the MM-Vet v2 dataset. }
  \label{fig: Test-time FDR (mean±std) and Power (mean) on the MM-Vet v2 datasets.}
\end{figure*}
\begin{figure*}[!h]
  \centering
  \begin{subfigure}[b]{0.246\textwidth}
    \centering
    \includegraphics[width=\textwidth]{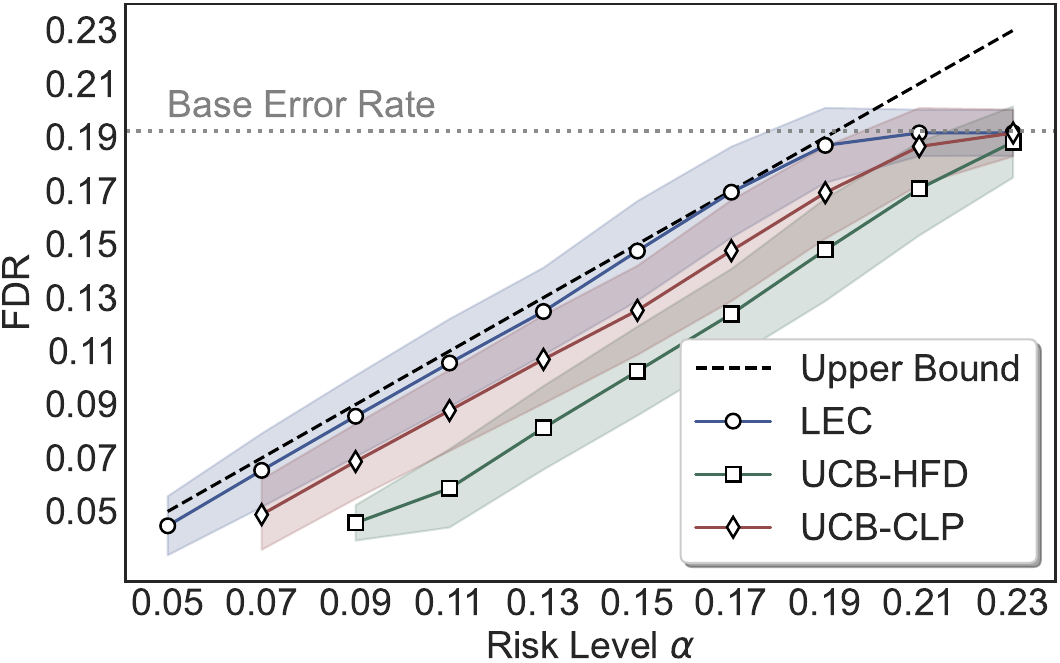}
    \caption{InternVL2-1B.}
  \end{subfigure}
  \hfill
  \begin{subfigure}[b]{0.246\textwidth}
    \centering
    \includegraphics[width=\textwidth]{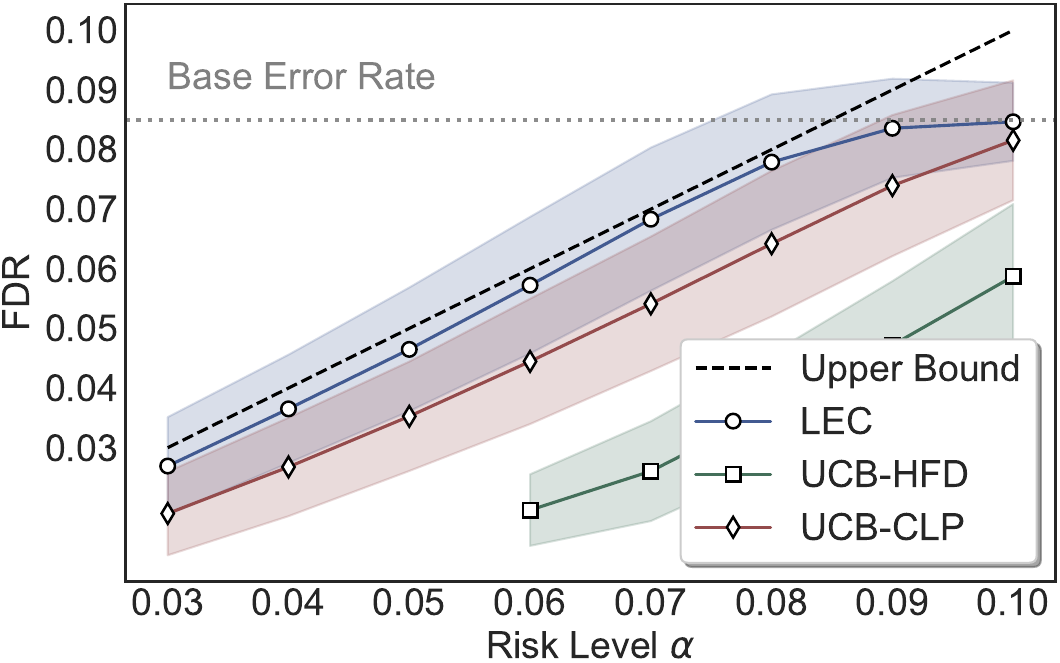}
    \caption{InternVL2-8B.}
  \end{subfigure}
  \hfill
  \begin{subfigure}[b]{0.246\textwidth}
    \centering
    \includegraphics[width=\textwidth]{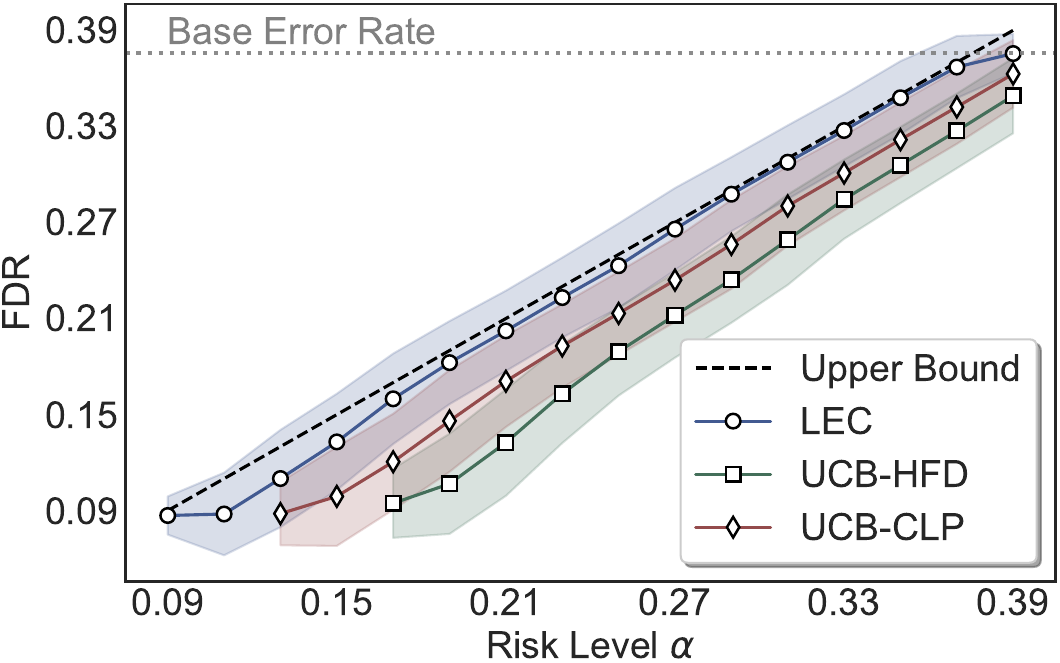}
    \caption{LLaVA-1.5-7B.}
  \end{subfigure}
  \hfill
  \begin{subfigure}[b]{0.246\textwidth}
    \centering
    \includegraphics[width=\textwidth]{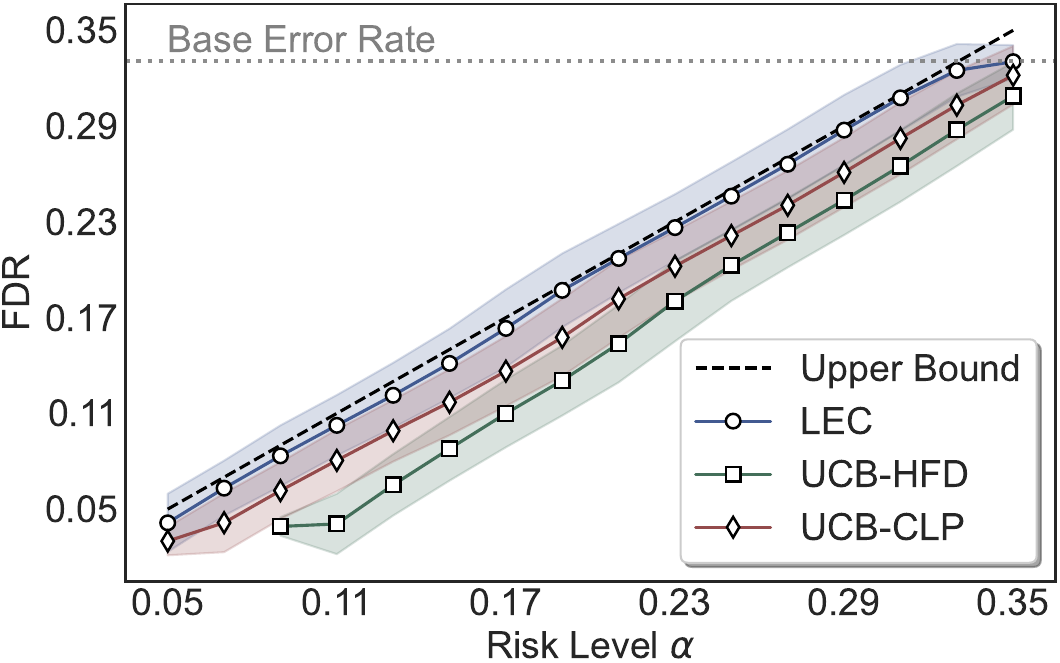}
    \caption{LLaVA-V1.6-Mistral-7B.}
  \end{subfigure}

  \begin{subfigure}[b]{0.246\textwidth}
    \centering
    \includegraphics[width=\textwidth]{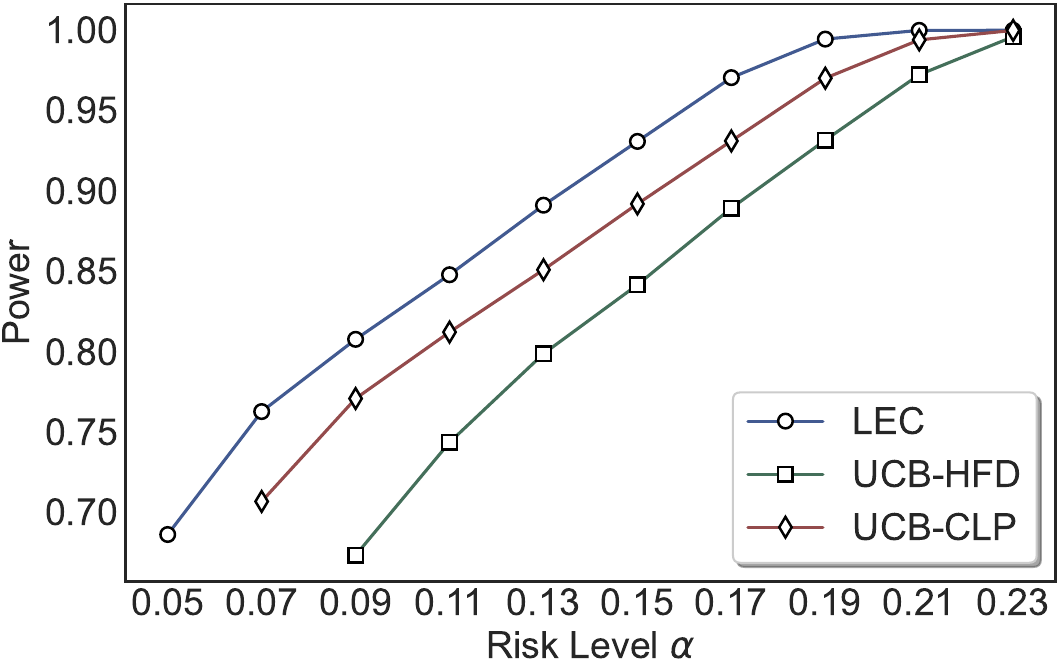}
    \caption{InternVL2-1B.}
  \end{subfigure}
  \hfill
  \begin{subfigure}[b]{0.246\textwidth}
    \centering
    \includegraphics[width=\textwidth]{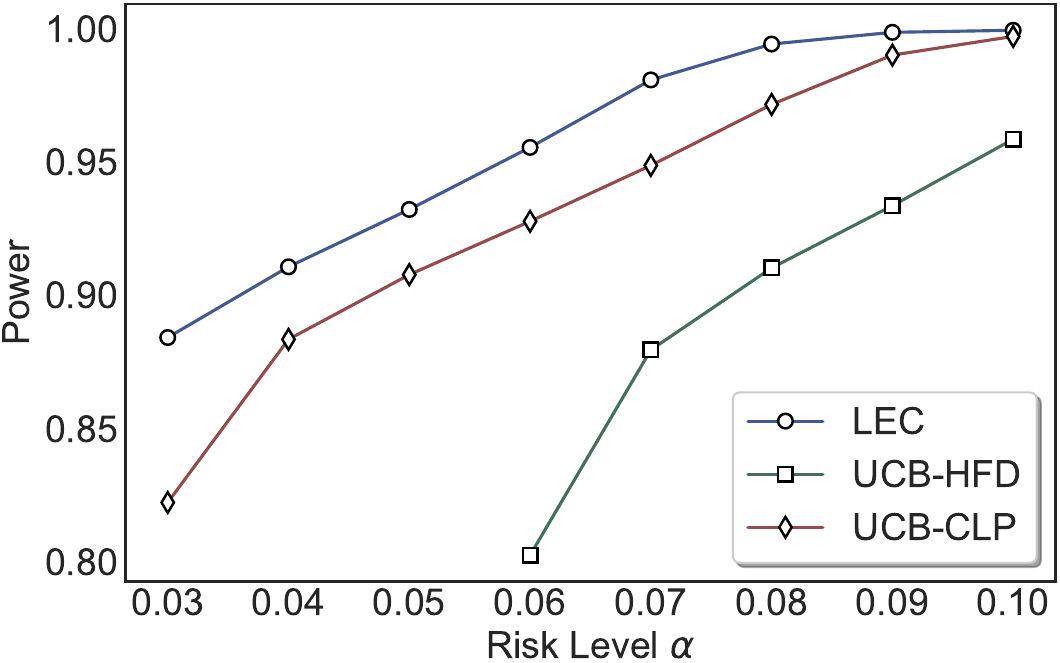}
    \caption{InternVL2-8B.}
  \end{subfigure}
  \hfill
  \begin{subfigure}[b]{0.246\textwidth}
    \centering
    \includegraphics[width=\textwidth]{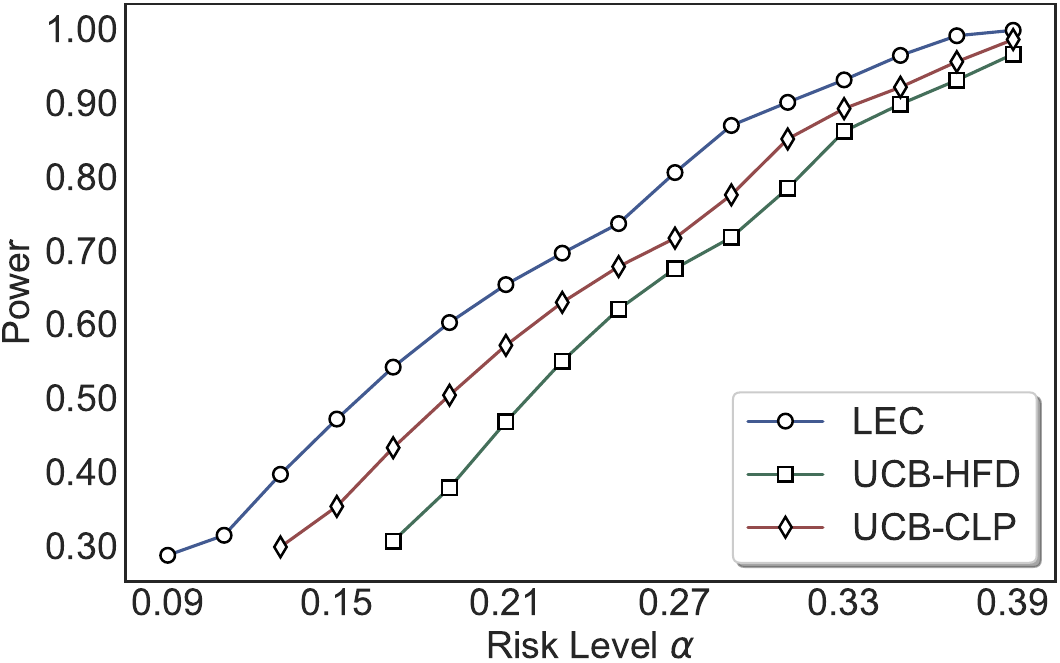}
    \caption{LLaVA-1.5-7B.}
  \end{subfigure}
  \hfill
  \begin{subfigure}[b]{0.246\textwidth}
    \centering
    \includegraphics[width=\textwidth]{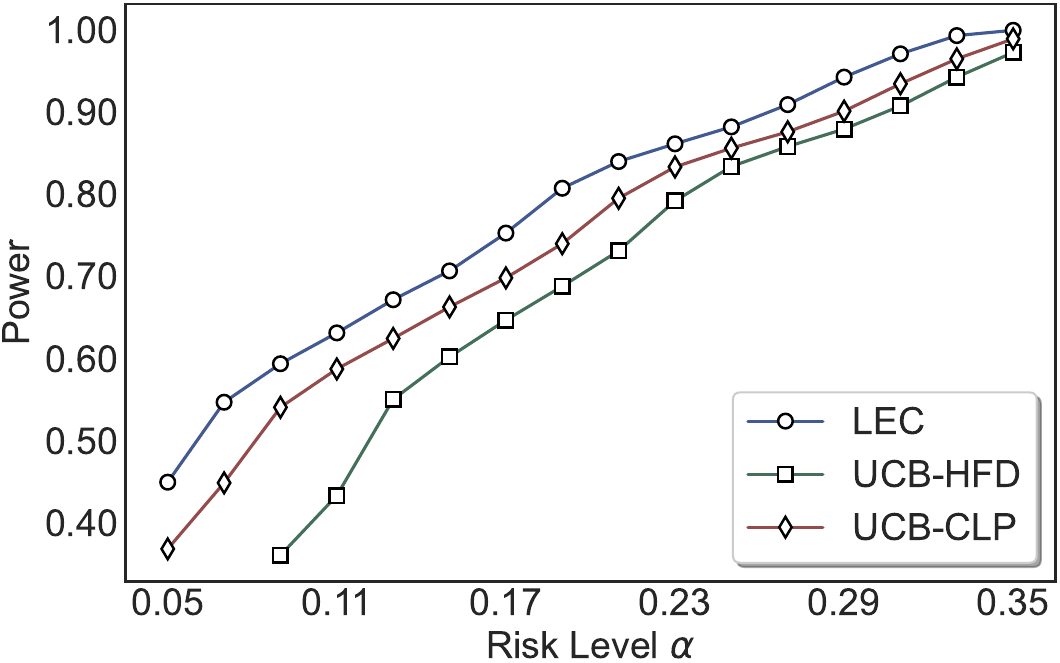}
    \caption{LLaVA-V1.6-Mistral-7B.}
  \end{subfigure}

   \caption{Test-time empirical selection-conditioned error rate (mean$\pm$std) and power (mean) on the ScienceQA dataset. }
  \label{fig: Test-time FDR (mean±std) and Power (mean) on the ScienceQA datasets.}
\end{figure*}
\begin{figure*}[!h]
    \centering
    \includegraphics[width=0.85\linewidth]{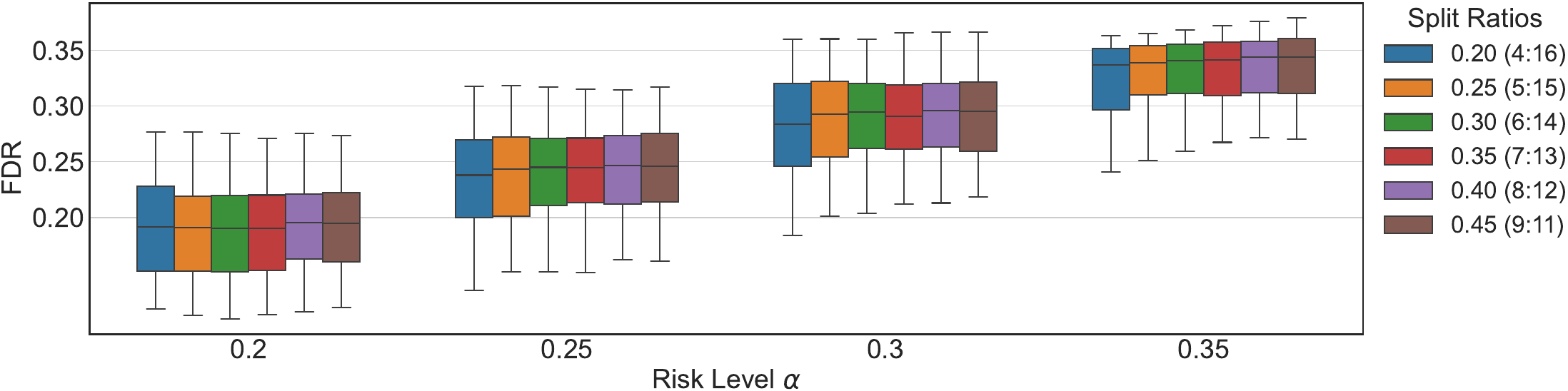}
    \caption{Empirical selection-conditioned error control across various split ratios on MM-Vet v2 with LLaVA-V1.6-Mistral-7B.}
    \label{fig: FDR control across various calibration-test split ratios on the MM-Vet v2 dataset with the LLaVA-V1.6-Mistral-7B model.}
\end{figure*}
\begin{figure*}[!t]
    \centering
    \includegraphics[width=0.85\linewidth]{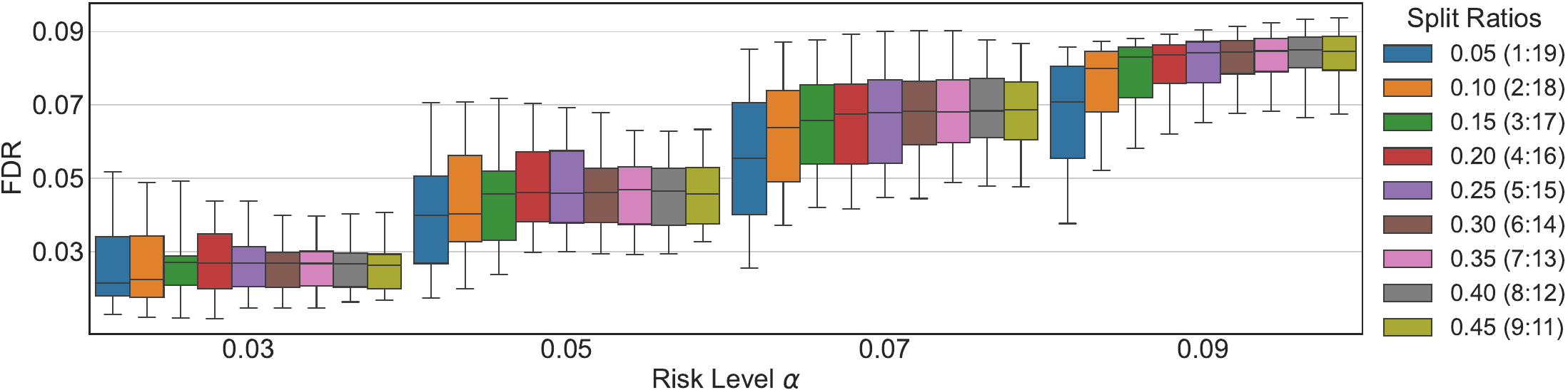}
    \caption{Empirical selection-conditioned error control across various split ratios on ScienceQA with InternVL2-8B.}
    \label{fig: FDR control across various calibration-test split ratios on the ScienceQA dataset with the InternVL2-8B model.}
\end{figure*}

\noindent \textbf{Evaluations on VQA Benchmarks.} 
We further evaluate the proposed \texttt{LEC} framework on two multimodal question answering benchmarks, open-ended MM-Vet v2 and closed-ended ScienceQA, using four representative LVLMs. 
As shown in Figures~\ref{fig: Test-time FDR (mean±std) and Power (mean) on the MM-Vet v2 datasets.} and~\ref{fig: Test-time FDR (mean±std) and Power (mean) on the ScienceQA datasets.}, \texttt{LEC} consistently achieves valid test-time risk control across all LVLMs and VQA datasets, while providing noticeably higher power compared to both UCB-based baselines. 
These results demonstrate that the benefits of \texttt{LEC} are not limited to language-only settings, but extend naturally to multimodal QA. 
Additionally, as demonstrated in Figures~\ref{fig: FDR control across various calibration-test split ratios on the MM-Vet v2 dataset with the LLaVA-V1.6-Mistral-7B model.} and~\ref{fig: FDR control across various calibration-test split ratios on the ScienceQA dataset with the InternVL2-8B model.}, \texttt{LEC} consistently maintains marginal guarantees across a wide range of calibration-test split ratios.

\begin{figure*}[h]
  \centering
  \begin{subfigure}[b]{0.47\textwidth}
    \centering
    \includegraphics[width=\textwidth]{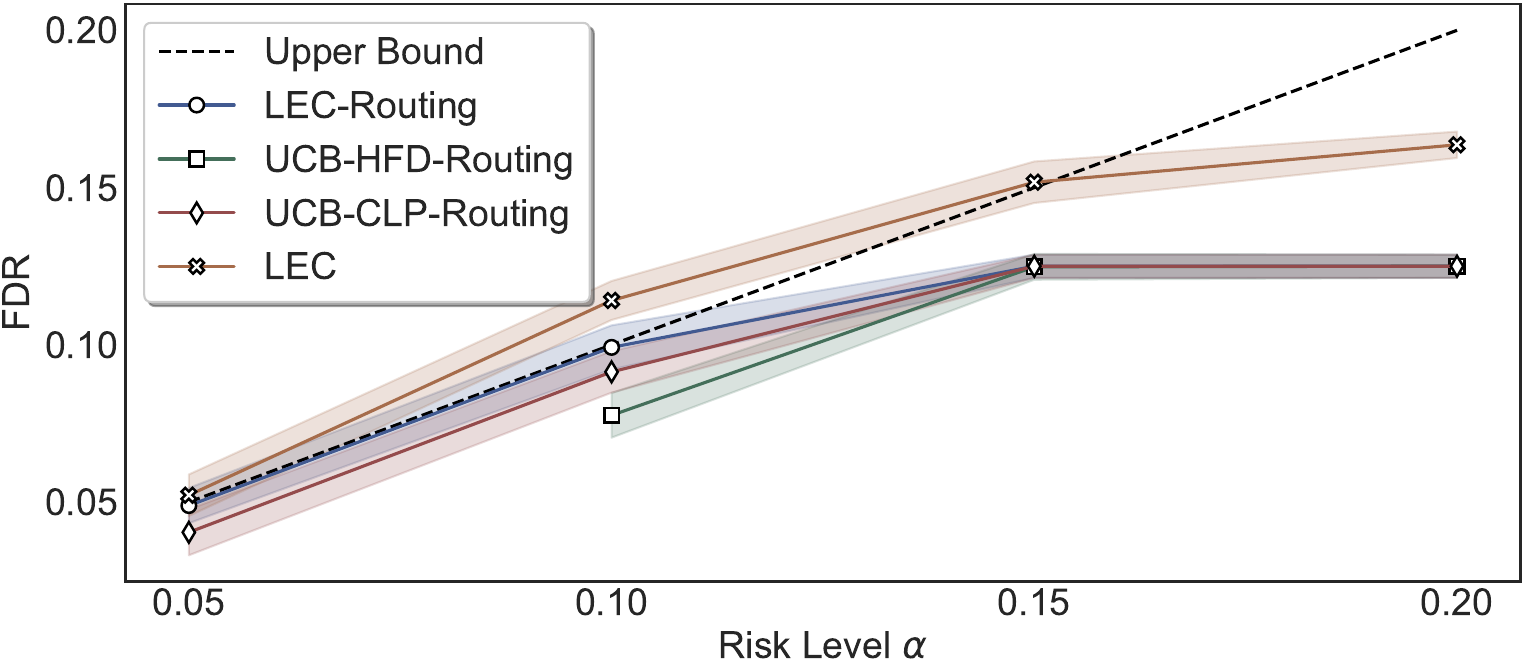}
    \caption{Qwen2.5-3B $\&$ OpenChat-3.5.}
  \end{subfigure}
  \hfill
  \begin{subfigure}[b]{0.47\textwidth}
    \centering
    \includegraphics[width=\textwidth]{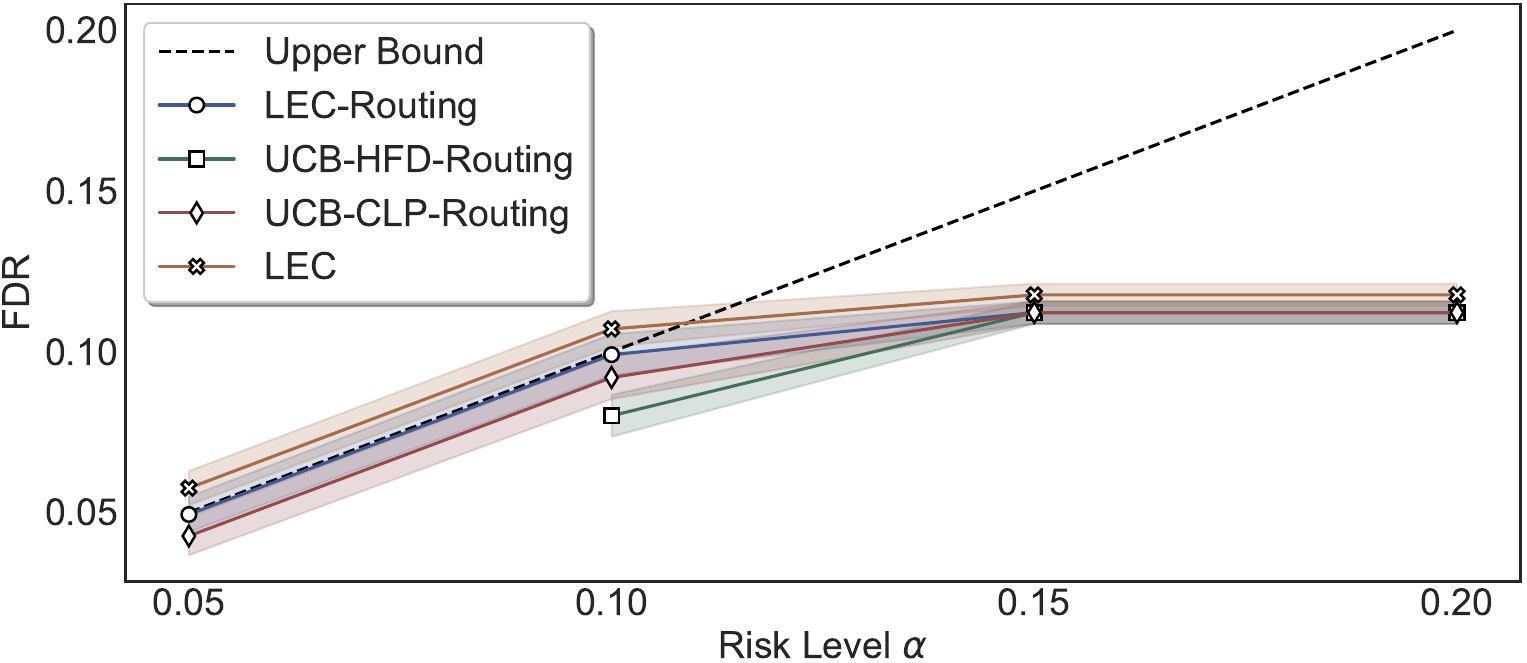}
    \caption{OpenChat-3.5 $\&$ LLaMA-3.1-8B.}
  \end{subfigure}
  
  \begin{subfigure}[b]{0.495\textwidth}
    \centering
    \includegraphics[width=\textwidth]{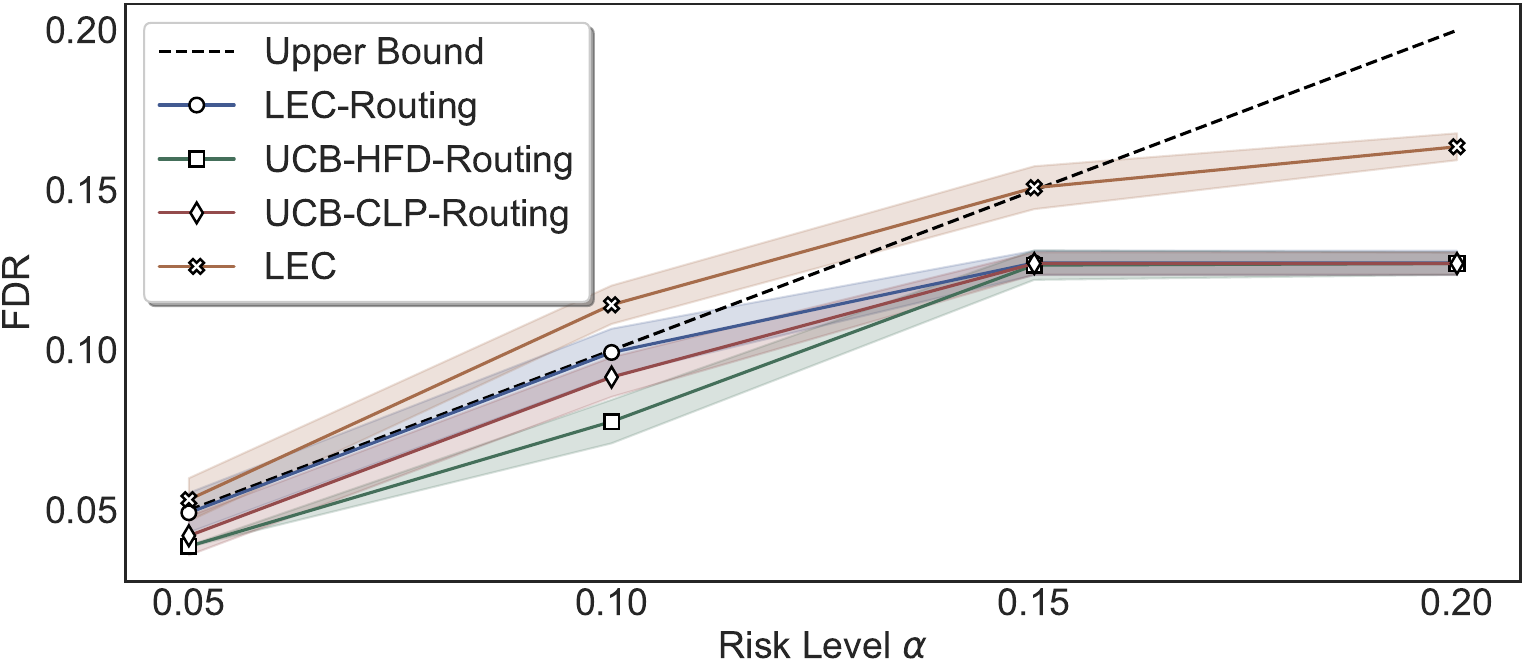}
    \caption{Qwen2.5-3B $\&$ LLaMA-3.1-8B.}
  \end{subfigure}
  \hfill
  \begin{subfigure}[b]{0.495\textwidth}
    \centering
    \includegraphics[width=\textwidth]{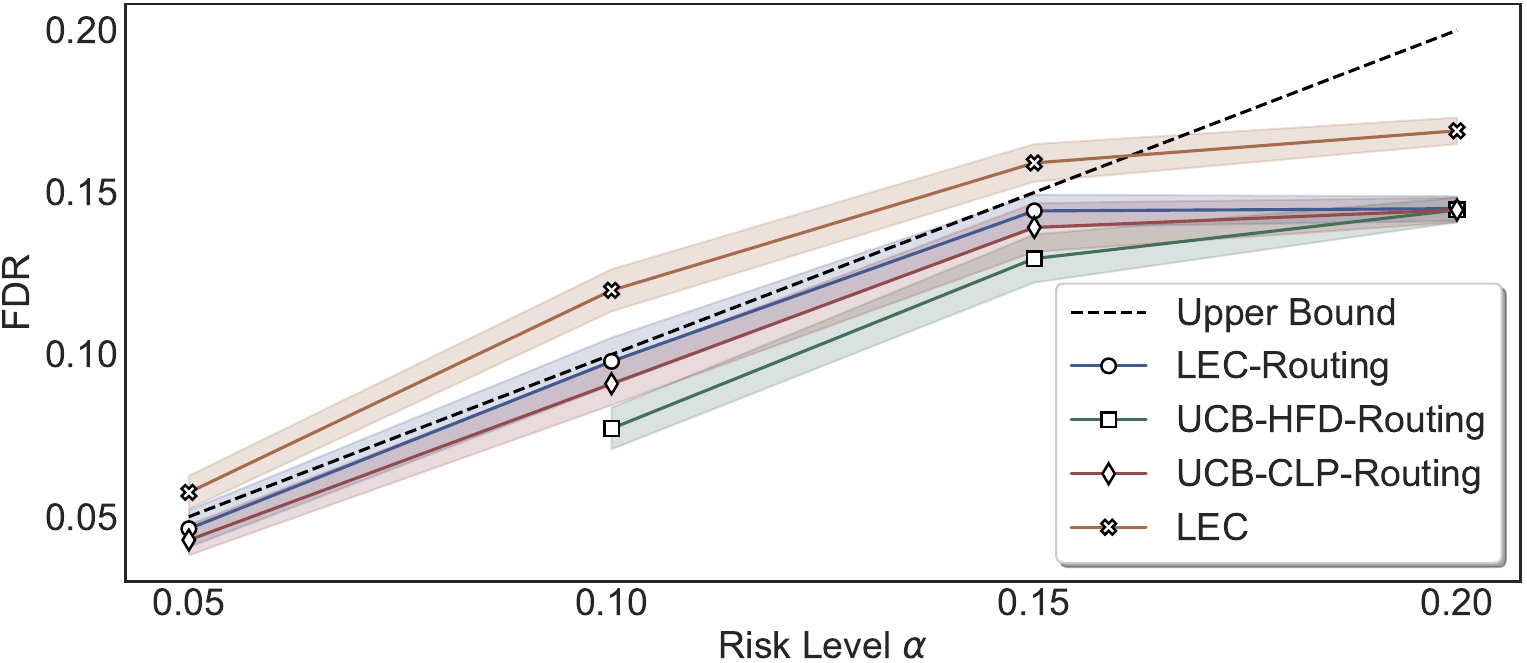}
    \caption{Vicuna-7B-V1.5 $\&$ LLaMA-3.1-8B.}
  \end{subfigure}

      \caption{Test-time empirical system-level selection-conditioned error rate of two-model routing systems on TriviaQA (mean$\pm$std). }
  \label{fig: routing FDR control triviaqa.}
\end{figure*}

\noindent \textbf{Two-Model Routing on TriviaQA.} 
Figure~\ref{fig: routing FDR control triviaqa.} reports the test-time selection-conditioned error rate of two-model routing systems on TriviaQA under different target risk levels. 
Across all model pairs, \texttt{LEC-Routing} consistently achieves tight risk control, operating close to the target risk level while remaining below the theoretical upper bound. 
In contrast, both \texttt{UCB-HFD-Routing} and \texttt{UCB-CLP-Routing} exhibit more conservative behavior, with realized selection-conditioned error rates staying noticeably below the target across most risk levels. 
Crucially, \texttt{LEC} calibrating independently for each model, without joint threshold calibration, tends to violate the target risk level or behave inconsistently across model pairs, since the resulting selection and error indicators no longer satisfy the finite-sample sufficient condition at the system level.

\begin{figure*}[!t]
  \centering
  \begin{subfigure}[b]{0.495\textwidth}
    \centering
    \includegraphics[width=\textwidth]{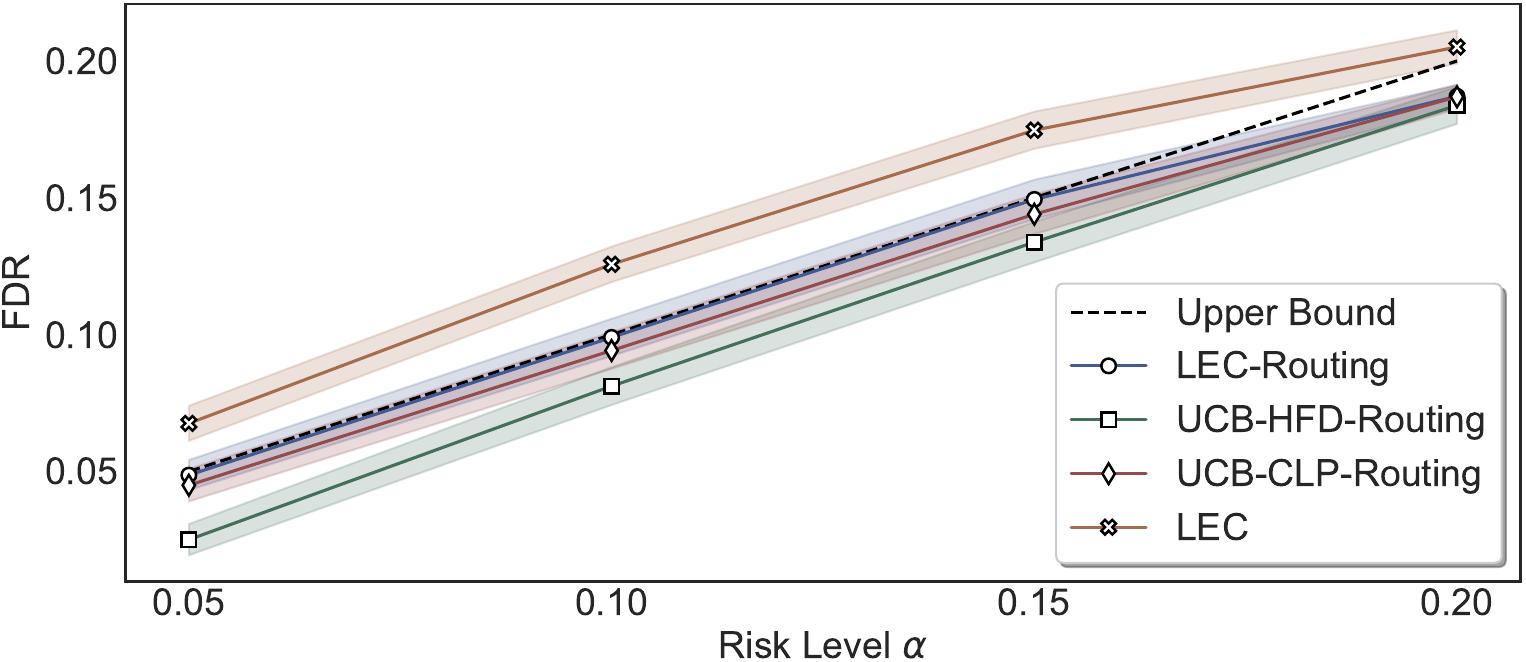}
    \caption{Qwen2.5-7B $\&$ LLaMA-3.1-70B}
  \end{subfigure}
  \hfill
  \begin{subfigure}[b]{0.495\textwidth}
    \centering
    \includegraphics[width=\textwidth]{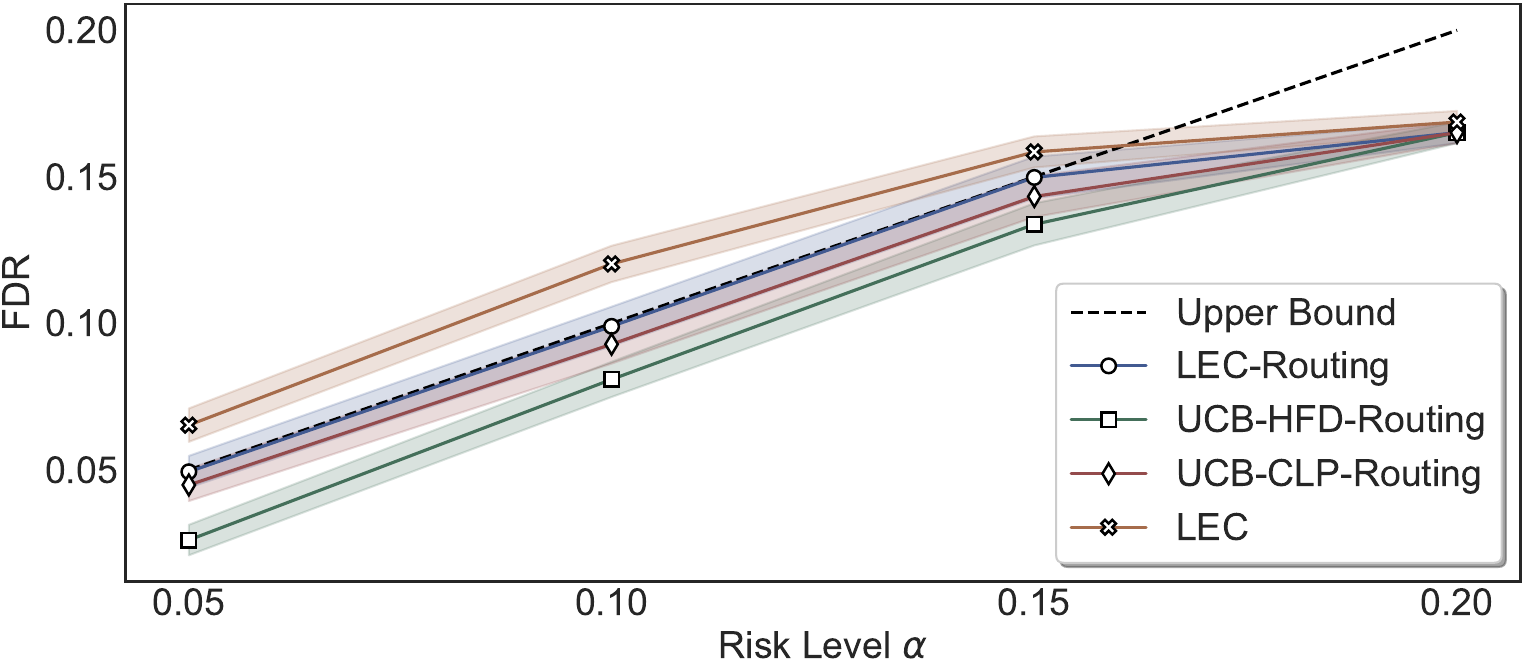}
    \caption{Qwen2.5-14B $\&$ LLaMA-3.1-70B.}
  \end{subfigure}

      \caption{Test-time system-level selection-conditioned error rate of two-model routing systems on CommonsenseQA (mean$\pm$std).}
  \label{fig: routing FDR control commonsenseqa 70B.}
\end{figure*}

\noindent \textbf{Accepted Correct Samples under Two-Model Routing.} 
Table~\ref{tab:commonsense-routing-comparison} further reports the allocation of accepted samples and the number of accepted correct predictions under two-model routing on the CommonsenseQA dataset. 
Compared to using either model alone, \texttt{LEC-Routing} consistently increases the number of accepted correct samples across model pairs and risk levels, while maintaining valid system-level risk control, as demonstrated in Figure~\ref{fig: routing FDR control commonsenseqa 70B.}. 
At $\alpha=0.05$, for example, routing Qwen2.5-7B with LLaMA-3.1-70B under \texttt{LEC-Routing} increases the total acceptance rate from 50.69$\%$ (Qwen2.5-7B alone) and 55.57$\%$ (LLaMA-3.1-70B alone) to 57.09$\%$, resulting in a higher number of accepted correct samples. Similar trends are observed at $\alpha=0.10$ and for the Qwen2.5-14B pairing, where routing yields both higher coverage and more correct acceptances than either individual model. 
Compared to UCB-based routing methods, \texttt{LEC-Routing} achieves a more favorable balance between allocation efficiency and correctness, reflecting its tighter feasible region under finite-sample guarantees. Overall, these results suggest that joint calibration under the \texttt{LEC} framework can effectively leverage complementary strengths of multiple models to retain more correct predictions than single-model deployment, without sacrificing statistical reliability. 

Beyond the total number of accepted correct samples, Table~\ref{tab:commonsense-routing-comparison} highlights an important practical advantage of \texttt{LEC-Routing}: it tends to allocate a larger fraction of accepted samples to the cheaper primary model and invokes the more expensive secondary model only when necessary, while still improving overall correctness under the same risk budget. 
For instance, under the pair {Qwen2.5-7B} (primary) $\rightarrow$ {LLaMA-3.1-70B} (secondary) at $\alpha=0.05$, \texttt{LEC-Routing} accepts 57.09$\%$ of test samples in total, where 43.77$\%$ are handled by the primary model and only 13.32$\%$ are delegated to the 70B model, yielding 2700 accepted correct samples. In contrast, \texttt{UCB-HFD-Routing} accepts substantially fewer samples overall (41.62$\%$) and produces fewer correct acceptances (2016), despite still routing 18.44$\%$ of samples to the expensive model. \texttt{UCB-CLP-Routing} attains a similar total acceptance rate to \texttt{LEC-Routing} (56.95$\%$ vs.\ 57.09$\%$) and comparable correct acceptances (2703 vs.\ 2700), but it requires routing a larger fraction to the 70B model (17.14$\%$ vs.\ 13.32$\%$), implying higher inference cost for essentially the same utility.

Overall, these results suggest that \texttt{LEC-Routing} is not only statistically reliable but also operationally appealing: under finite-sample risk control, it can exploit the cheap model for the majority of accepted predictions and reserve the expensive model for genuinely uncertain cases, leading to a more favorable accuracy-cost trade-off in real deployments. 

\input{tables/routingCommonsenseQA}

%% file: tables/powerCommonsenseQA.tex
\begin{table*}[!t]
\centering
\caption{Power comparison on the CommonsenseQA dataset (mean).} 
\label{tab: power comparison (CommonsenseQA) white se}
\adjustbox{max width=0.95\linewidth}{
    \begin{tabular}{ccccccccccc}
        \toprule
        
        \textbf{LLMs} & \textbf{Methods /} $\mathbf{\alpha}$ & \textbf{0.05} & \textbf{0.1} & \textbf{0.15} & \textbf{0.2} & \textbf{0.25} & \textbf{0.3} & \textbf{0.35} & \textbf{0.4} & \textbf{0.45}\\

        \midrule

        \multirow{3}{*}{OpenChat-3.5} & \texttt{UCB-CLP} & 0.6556 & 0.8397 & 0.9401 & 0.9999$_{(1)}$ & 0.9999$_{(4)}$ & 0.9999$_{(4)}$ & 0.9999$_{(4)}$ & 0.9999$_{(4)}$ & 0.9999$_{(4)}$\\
        {} & \texttt{UCB-HFD} & 0.4887 & 0.8024 & 0.9214 & 0.9989 & 0.9999$_{(4)}$ & 0.9999$_{(4)}$ & 0.9999$_{(4)}$ & 0.9999$_{(4)}$ & 0.9999$_{(4)}$\\
        {} & \cellcolor{gray!20}  \texttt{LEC} & \textbf{0.6850} & \textbf{0.8559} & \textbf{0.9567} & \textbf{0.9999}$_{(4)}$ & \textbf{0.9999}$_{(4)}$ & \textbf{0.9999}$_{(4)}$ & \textbf{0.9999}$_{(4)}$ & \textbf{0.9999}$_{(4)}$ & \textbf{0.9999}$_{(4)}$\\
        \midrule
        \multirow{3}{*}{Qwen2.5-3B} & \texttt{UCB-CLP} & 0.1803 & 0.6242 & 0.8084 & 0.9114 & 0.9964 & 0.9999$_{(9)}$ & 0.9999$_{(9)}$ & 0.9999$_{(9)}$ & 0.9999$_{(9)}$\\
        {} & \texttt{UCB-HFD} & - & 0.4976 & 0.7758 & 0.8968 & 0.9851 & 0.9999$_{(9)}$ & 0.9999$_{(9)}$ & 0.9999$_{(9)}$ & 0.9999$_{(9)}$\\
        {} & \cellcolor{gray!10}  \texttt{LEC} & \textbf{0.2553} & \textbf{0.6779} & \textbf{0.8366} & \textbf{0.9315} & \textbf{0.9999}$_{(8)}$ & \textbf{0.9999}$_{(9)}$ & \textbf{0.9999}$_{(9)}$ & \textbf{0.9999}$_{(9)}$ & \textbf{0.9999}$_{(9)}$\\
        \midrule
        \multirow{3}{*}{Qwen2.5-7B} & \texttt{UCB-CLP} & 0.5704 & 0.7647 & 0.8810 & 0.9685 & 0.9999$_{(8)}$ & 0.9999$_{(8)}$ & 0.9999$_{(8)}$ & 0.9999$_{(8)}$ & 0.9999$_{(8)}$\\
        {} & \texttt{UCB-HFD} & 0.3546 & 0.7186 & 0.8573 & 0.9522 & 0.9999$_{(8)}$ & 0.9999$_{(8)}$ & 0.9999$_{(8)}$ & 0.9999$_{(8)}$ & 0.9999$_{(8)}$\\
        {} & \cellcolor{gray!10}  \texttt{LEC} & \textbf{0.6108} & \textbf{0.7850} & \textbf{0.8982} & \textbf{0.9839} & \textbf{0.9999}$_{(8)}$ & \textbf{0.9999}$_{(8)}$ & \textbf{0.9999}$_{(8)}$ & \textbf{0.9999}$_{(8)}$ & \textbf{0.9999}$_{(8)}$\\
        \midrule
       \multirow{3}{*}{Qwen2.5-14B} & \texttt{UCB-CLP} & 0.6583 & 0.8644 & 0.9595 & 0.9999$_{(7)}$ & 0.9999$_{(7)}$ & 0.9999$_{(7)}$ & 0.9999$_{(7)}$ & 0.9999$_{(7)}$ & 0.9999$_{(7)}$\\
        {} & \texttt{UCB-HFD} & 0.4402 & 0.8320 & 0.9426 & 0.9999$_{(7)}$ & 0.9999$_{(7)}$ & 0.9999$_{(7)}$ & 0.9999$_{(7)}$ & 0.9999$_{(7)}$ & 0.9999$_{(7)}$\\
        {} & \cellcolor{gray!10}  \texttt{LEC} & \textbf{0.6997} & \textbf{0.8833} & \textbf{0.9746} & \textbf{0.9999}$_{(7)}$ & \textbf{0.9999}$_{(7)}$ & \textbf{0.9999}$_{(7)}$ & \textbf{0.9999}$_{(7)}$ & \textbf{0.9999}$_{(7)}$ & \textbf{0.9999}$_{(7)}$\\
        \midrule
        \multirow{3}{*}{Vicuna-7B-V1.5} & \texttt{UCB-CLP} & - & 0.0346 & 0.1039 & 0.3437 & 0.5053 & 0.6684 & 0.8283 & 0.9641 & 0.9999$_{(9)}$\\
        {} & \texttt{UCB-HFD} & - & - & 0.0583 & 0.2717 & 0.4674 & 0.6312 & 0.7976 & 0.9517 & 0.9999$_{(9)}$\\
        {} & \cellcolor{gray!10}  \texttt{LEC} & - & \textbf{0.0575} & \textbf{0.2263} & \textbf{0.3963} & \textbf{0.5477} & \textbf{0.7099} & \textbf{0.8716} & \textbf{0.9845} & \textbf{0.9999}$_{(9)}$\\
        \midrule
       \multirow{3}{*}{Vicuna-13B-V1.5} & \texttt{UCB-CLP} & - & 0.1365 & 0.4695 & 0.6468 & 0.7945 & 0.9138 & 0.9901 & 0.9999$_{(9)}$ & 0.9999$_{(9)}$\\
        {} & \texttt{UCB-HFD} & - & - & 0.3829 & 0.6188 & 0.7643 & 0.8989 & 0.9818 & 0.9999$_{(9)}$ & 0.9999$_{(9)}$\\
        {} & \cellcolor{gray!10}  \texttt{LEC} & \textbf{0.0543} & \textbf{0.2785} & \textbf{0.5278} & \textbf{0.6736} & \textbf{0.8311} & \textbf{0.9342} & \textbf{0.9989} & \textbf{0.9999}$_{(9)}$ & \textbf{0.9999}$_{(9)}$\\
        \midrule
        \multirow{3}{*}{LLaMA-3.1-8B} & \texttt{UCB-CLP} & 0.3903 & 0.6078 & 0.7702 & 0.8739 & 0.9556 & 0.9999$_{(7)}$ & 0.9999$_{(7)}$ & 0.9999$_{(7)}$ & 0.9999$_{(7)}$\\
        {} & \texttt{UCB-HFD} & - & 0.5502 & 0.7398 & 0.8596 & 0.9417 & 0.9999$_{(6)}$ & 0.9999$_{(7)}$ & 0.9999$_{(7)}$ & 0.9999$_{(7)}$\\
        {} & \cellcolor{gray!10}  \texttt{LEC} & \textbf{0.4341} & \textbf{0.6468} & \textbf{0.7939} & \textbf{0.8893} & \textbf{0.9745} & \textbf{0.9999}$_{(7)}$ & \textbf{0.9999}$_{(7)}$ & \textbf{0.9999}$_{(7)}$ & \textbf{0.9999}$_{(7)}$\\
        \midrule
        \multirow{3}{*}{LLaMA-3.1-70B} & \texttt{UCB-CLP} & 0.6328 & 0.8099 & 0.9193 & 0.9929 & 0.9999$_{(7)}$ & 0.9999$_{(7)}$ & 0.9999$_{(7)}$ & 0.9999$_{(7)}$ & 0.9999$_{(7)}$\\
        {} & \texttt{UCB-HFD} & 0.5109 & 0.7792 & 0.9012 & 0.9856 & 0.9999$_{(7)}$ & 0.9999$_{(7)}$ & 0.9999$_{(7)}$ & 0.9999$_{(7)}$ & 0.9999$_{(7)}$\\
        {} &\cellcolor{gray!10}  \texttt{LEC} & \cellcolor{gray!10} \textbf{0.6570} & \cellcolor{gray!10} \textbf{0.8323} & \cellcolor{gray!10} \textbf{0.9383} & \cellcolor{gray!10} \textbf{0.9990} & \cellcolor{gray!10} \textbf{0.9999}$_{(7)}$ & \cellcolor{gray!10} \textbf{0.9999}$_{(7)}$ & \cellcolor{gray!10} \textbf{0.9999}$_{(7)}$ & \cellcolor{gray!10} \textbf{0.9999}$_{(7)}$ & \cellcolor{gray!10} \textbf{0.9999}$_{(7)}$\\

        \bottomrule
        
    \end{tabular}
}    
\end{table*}

%% file: tables/samplingSize.tex
\begin{table*}[!h]
\centering
\caption{Test-time empirical selection-conditioned error rate and power on TriviaQA with the LLaMA-3.1-70B model across various user-specified risk levels under different uncertainty methods and sampling sizes (mean). 
We also report AUROC to reflect the quality of UQ methods.}
\label{tab:csp_sampling_combined_mean}
\adjustbox{max width=0.8\textwidth}{
\begin{tabular}{cccccc cccc c}
\toprule
\textbf{Uncertainty} & \textbf{Sampling} &
\multicolumn{4}{c}{\textbf{Test-time FDR}} &
\multicolumn{4}{c}{\textbf{Power}} &
\multirow{2}{*}{\textbf{AUROC}} \\
\cmidrule(lr){3-6}
\cmidrule(lr){7-10}
\textbf{Estimators} & \textbf{Sizes} &
\textbf{0.02} & \textbf{0.03} &\textbf{ 0.04} & \textbf{0.05} &
\textbf{0.02} & \textbf{0.03} & \textbf{0.04} & \textbf{0.05} & \\
\midrule

\multirow{4}{*}{SE} 
& 5  & 0.0157 & 0.0266 & 0.0362 & 0.0470 & 0.7573 & 0.9326 & 0.9730 & 0.9985 & 0.8007 \\
& 10 & 0.0193 & 0.0296 & 0.0396 & 0.0474 & 0.8667 & 0.9491 & 0.9831 & 0.9995 & 0.8247 \\
& 15 & 0.0197 & 0.0299 & 0.0398 & 0.0474 & 0.8903 & 0.9545 & 0.9851 & 0.9995 & 0.8379 \\
& 20 & 0.0198 & 0.0300 & 0.0399 & 0.0474 & 0.8930 & 0.9551 & 0.9867 & 0.9995 & 0.8400 \\
\midrule

\multirow{4}{*}{EigV} 
& 5  & 0.0198 & 0.0299 & 0.0398 & 0.0475 & 0.9364 & 0.9794 & 0.9949 & 0.9996 & 0.8668 \\
& 10 & 0.0199 & 0.0299 & 0.0397 & 0.0475 & 0.9414 & 0.9804 & 0.9960 & 0.9996 & 0.8821 \\
& 15 & 0.0198 & 0.0299 & 0.0398 & 0.0475 & 0.9430 & 0.9817 & 0.9965 & 0.9996 & 0.8853 \\
& 20 & 0.0199 & 0.0299 & 0.0397 & 0.0474 & 0.9436 & 0.9829 & 0.9968 & 0.9997 & 0.8874 \\
\bottomrule
\end{tabular}}
\end{table*}

%% file: tables/routingCommonsenseQA.tex
\begin{table*}[!h]
\centering
\caption{Allocation of accepted samples ($\%$) and accepted correct samples of two-model routing on the CommonsenseQA (mean).}
\label{tab:commonsense-routing-comparison}
\adjustbox{max width=\linewidth}{
\begin{tabular}{cccccccccc}
\toprule
\multirow{2}{*}{\large{\textbf{LLMs}}} & \multirow{2}{*}{\large{\textbf{Methods}}} 
& \multicolumn{4}{c}{$\boldsymbol{\alpha=0.05}$}
& \multicolumn{4}{c}{$\boldsymbol{\alpha=0.10}$}\\
\cmidrule(lr){3-6}
\cmidrule(lr){7-10}
{} & {} 
& \textbf{Prop.~1} & \textbf{Prop.~2} & \textbf{Total} & \textbf{Corr.}
& \textbf{Prop.~1} & \textbf{Prop.~2} & \textbf{Total} & \textbf{Corr.}\\
\midrule
Qwen2.5-7B & \texttt{LEC} 
& 50.69 & - & 50.69 & 2393
& 68.81 & - & 68.81 & 3078\\
LLaMA-3.1-70B & \texttt{LEC} 
& - & 55.57 & 55.57 & 2624
& - & 74.39 & 74.39 & 3329\\
\rowcolor{gray!20}  Qwen2.5-7B $\&$ LLaMA-3.1-70B &  \texttt{LEC-Routing} 
&  43.77 & 13.32 & \textbf{57.09} & \textbf{2700}
&  59.22 &  18.24 & \textbf{77.46} &  \textbf{3469} \\
Qwen2.5-7B $\&$ LLaMA-3.1-70B & \texttt{UCB-HFD-Routing} 
& 23.18 & 18.44 & 41.62 & 2016
& 53.37 & 18.96 & 72.33 & 3303\\
Qwen2.5-7B $\&$ LLaMA-3.1-70B & \texttt{UCB-CLP-Routing} 
& 39.81 & 17.14 & 56.95 & 2703
& 57.97 & 18.04 & 76.01 & 3422\\
\midrule
\midrule
Qwen2.5-14B & \texttt{LEC} 
& 61.19 & - & 61.19 & 2789
& 81.60 & - & 81.60 & 3524\\
LLaMA-3.1-70B & \texttt{LEC} 
& - & 57.34 & 57.34 & 2614
& - & 76.46 & 76.46 & 3302\\
\rowcolor{gray!20} Qwen2.5-14B $\&$ LLaMA-3.1-70B & \texttt{LEC-Routing} 
& 55.91 & 7.58 & \textbf{63.49} & \textbf{2895}
& 77.77 & 6.19 & \textbf{83.96} & \textbf{3629}\\
Qwen2.5-14B $\&$ LLaMA-3.1-70B & \texttt{UCB-HFD-Routing} 
& 32.80 & 15.53 & 48.33 & 2257
& 73.19 & 4.57 & 77.76 & 3428\\
Qwen2.5-14B $\&$ LLaMA-3.1-70B & \texttt{UCB-CLP-Routing} 
& 53.66 & 7.31 & 60.97 & 2793
& 75.83 & 6.48 & 82.31 & 3581\\
\bottomrule
\end{tabular}
}
\end{table*}